\pdfoutput=1

\documentclass[11pt]{article}
\usepackage{graphicx}

\usepackage{acl}
\usepackage{times}
\usepackage{latexsym}
\usepackage{inconsolata}
\usepackage{xparse}
\NewDocumentCommand\emojibat{}{
    \includegraphics[scale=0.05]{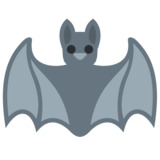}
}
\NewDocumentCommand\emojibaseballbat{}{
    \includegraphics[scale=0.05]{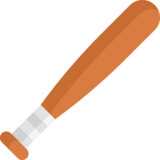}
}
\usepackage[T1]{fontenc}
\usepackage[utf8]{inputenc}

\usepackage{amsmath}

\usepackage{booktabs}
\usepackage{tabularx,ragged2e}
\newcolumntype{L}{>{\RaggedRight}X}

\usepackage{xcolor}
\definecolor{sense1}{HTML}{6495ED}
\definecolor{sense2}{HTML}{BA274A}
\definecolor{amb}{HTML}{8F5E9C}

\newcommand{\vb}[1]{\mathbf{#1}}

\newcommand{\colamb}[1]{{\color{amb} #1}}
\newcommand{\colone}[1]{{\color{sense1} #1}}
\newcommand{\coltwo}[1]{{\color{sense2} #1}}

\newcommand{\mone}{\colone{m_1}}
\newcommand{\mtwo}{\coltwo{m_2}}

\newcommand{\samb}[1]{\vb{s}_{#1}^{\colamb{\text{amb}}}}
\newcommand{\sone}[1]{\vb{s}_{#1}^{\colone{(1)}}}
\newcommand{\stwo}[1]{\vb{s}_{#1}^{\coltwo{(2)}}}

\newcommand{\ambone}{\colamb{\text{amb}}\rightarrow\colone{1}}
\newcommand{\ambtwo}{\colamb{\text{amb}}\rightarrow\coltwo{2}}

\newcommand{\vone}{\vb{v}_{\colone{1}}}
\newcommand{\vtwo}{\vb{v}_{\coltwo{2}}}

\newcommand{\vx}{\vb{x}}
\newcommand{\vy}{\vb{y}}
\newcommand{\vz}{\vb{z}}

\newcommand{\prompt}[1]{``\textit{#1}''}

\usepackage{microtype}
\usepackage{amsfonts}
\usepackage{amssymb}
\usepackage{float}
\usepackage{caption}
\usepackage{subcaption}
\usepackage{hyperref}
\usepackage{cleveref}
\crefname{section}{\S}{\S\S}
\Crefname{section}{\S}{\S\S}
\crefformat{section}{\S#2#1#3}
\crefname{figure}{Fig.}{Fig.}
\crefname{alg}{Alg.}{Alg.}
\crefname{thm}{Theorem}{Theorems}
\crefname{line}{line}{lines}
\crefname{appendix}{App.}{}
\crefname{equation}{Eq.}{Eq.}
\crefname{defin}{Def.}{Defs.}
\crefname{tab}{Table}{Tables}
\crefname{prop}{Proposition}{Propositions}

\newcommand{\dalle}{DALL\textperiodcentered E 2}
\DeclareMathOperator{\proj}{proj}
\DeclareMathOperator{\clip}{CLIP}

\usepackage[disable]{todonotes}

\makeatother

\newcommand{\note}[4][]{\todo[author=#2,color=#3,size=\scriptsize,fancyline,caption={},#1]{#4}} 
 
\newcommand{\ryan}[2][]{\note[#1]{ryan}{violet!40}{#2}}

\everypar{\looseness=-1}

\title{Schr\"{o}dinger's Bat: Diffusion Models Sometimes Generate Polysemous Words in Superposition}

\usepackage{tipa}

\newcommand{\ucambridge}{\emojibat}
\newcommand{\ethz}{\emojibaseballbat}

\author{Jennifer C. White$^{\ucambridge}$~\;~~\;~Ryan Cotterell$^{\ethz}$ \\
  ${\ucambridge}$University of Cambridge~\;~${\ethz}$ETH Z\"{u}rich \\
  \texttt{\href{mailto:jw2088@cam.ac.uk}{jw2088@cam.ac.uk}}~\;~\texttt{\href{mailto:ryan.cotterell@inf.ethz.ch}{ryan.cotterell@inf.ethz.ch}}
}

\begin{document}
\maketitle
\begin{abstract}
Recent work has shown that despite their impressive capabilities, text-to-image diffusion models such as \dalle\ \citep{dalle} can display strange behaviours when a prompt contains a word with multiple possible meanings, often generating images containing both senses of the word \citep{rassin-2022-dalle-2}.
In this work we seek to put forward a possible explanation of this phenomenon.
Using the similar Stable Diffusion model \citep{stablediffusion}, we first show that when given an input that is the sum of encodings of two distinct words, the model can produce an image containing both concepts represented in the sum.
We then demonstrate that the CLIP encoder used to encode prompts \citep{clip} encodes polysemous words as a superposition of meanings, and that using linear algebraic techniques we can edit these representations to influence the senses represented in the generated images.
Combining these two findings, we suggest that the homonym duplication phenomenon described by \citet{rassin-2022-dalle-2} is caused by diffusion models producing images representing both of the meanings that are present in superposition in the encoding of a polysemous word.
\newline
\newline
\vspace{1.5em}
\hspace{.5em}\includegraphics[width=1.25em,height=1.25em]{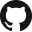}\hspace{.75em}\parbox{\dimexpr\linewidth-10\fboxsep-10\fboxrule}{\url{https://github.com/rycolab/diffusion-polysemy}}
\vspace{-.5em}
\end{abstract}

\section{Introduction}
\begin{figure}
    \centering
    \includegraphics[width=0.47\textwidth]{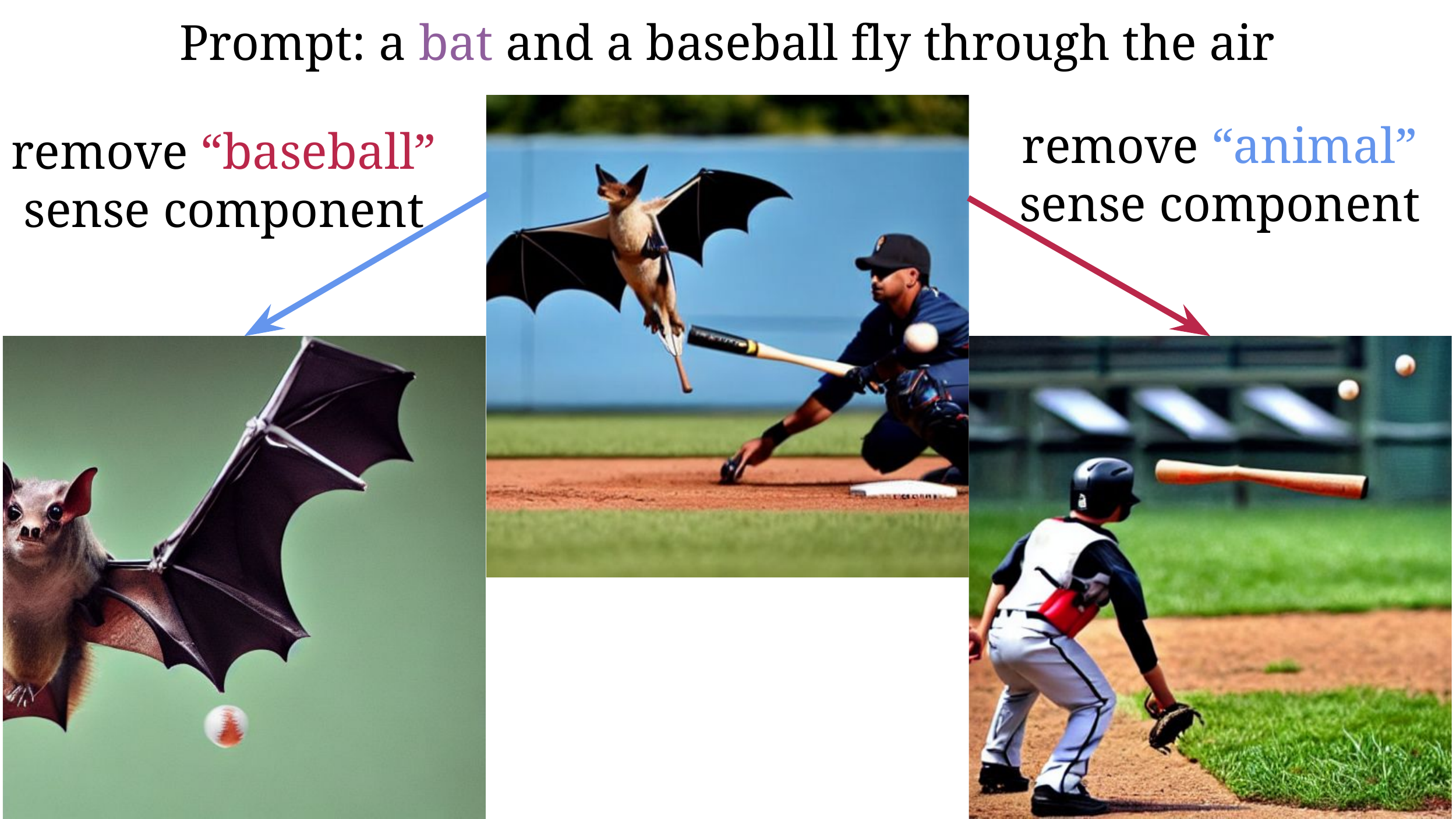}
    \caption{By editing prompt encodings by removing projections of vectors corresponding to each sense, we are able to reduce homonym duplication and produce the desired sense of the word \prompt{bat}.}
    \label{fig:bat}
\end{figure}

When humans encounter a word with more than one possible meaning, they can use information such as situational context and selectional restrictions to infer which meaning is the most likely. 
Models for language-based tasks such as translation struggle with this, often being biased towards the sense that is more commonly attested regardless of context and thus often producing a translation of the incorrect sense \citep{campolungo-etal-2022-dibimt}.
Recent research by \citet{rassin-2022-dalle-2} showed that text-to-image diffusion models behave in a different and unexpected way when confronted with a prompt containing polysemous words.
Specifically, they showed that when \dalle\ 
is prompted with a sentence containing a word with multiple possible meanings, such as \prompt{bat} 
or \prompt{bow}, the generated image often represents \emph{more than one} of these meanings.
For example, in the initial image in \cref{fig:bat} \prompt{bat} is represented both as a baseball bat and an animal.
They call this behaviour homonym duplication.
Diffusion models have attracted much attention, even beyond the reaches of the research communities that spawned them.\footnote{For example, the Twitter account \url{https://twitter.com/weirddalle}, which posts images generated by diffusion models, has over 1 million followers at time of writing.}
Despite this, we are still lacking in our understanding of some of the unusual tendencies they exhibit, such as the phenomenon of homonym duplication.\looseness=-1

In this work, we put forward one possible explanation of this phenomenon.
We suggest that this occurs due to polysemous words in the prompt being encoded as a linear superposition of their possible meanings.
In a text-based decoding task, such as translation, a representation of this form would not pose a problem, as the decoding model ultimately converts its representations into a probability distribution over tokens which it uses to select an output word.
Thus, although both possibilities will influence the distribution obtained, it is forced to select a word representing one meaning or the other.
However, diffusion models use the representation as input to condition the denoising process which predicts noise to remove, and such a choice between meanings is never forced.
This means that both meanings influence the denoising process, and both meanings can be represented in the final image.
To draw an analogy with quantum mechanics, in a famous thought experiment, \citet{schrodinger1935gegenwartige} described a scenario in which a cat in a box with a radioactive material is poisoned once a particle decay occurs, and suggested that since a particle's state exists in a superposition of decayed and undecayed states until an observation is made, that the cat must be considered to be both alive and dead until the box is opened.
In the translation scenario, the meaning of a word such as \prompt{bat} exists as a superposition of possibilities until an ``observation'' is made by sampling an output word.
Since this never happens in diffusion models, the metaphorical box is never opened and the bat continues to be \emph{both} a baseball bat and an animal.

Unfortunately, the internal workings of \dalle\ are not openly accessible, so we test this hypothesis by performing experiments using the implementation of Stable Diffusion \citep{stablediffusion} available on HuggingFace \citep{wolf-etal-2020-transformers}.
This is a smaller diffusion model which encodes prompts using a CLIP \citep{clip} encoder, the same type of encoder used by \dalle.
We begin by showing that diffusion models can represent multiple concepts that are in linear superposition, by demonstrating that when Stable Diffusion takes as input a sum of CLIP encodings for two words, both objects can be represented in the generated image (\cref{sec:sumenc}).
We then demonstrate that CLIP encodes polysemous words as a sum over their possible meanings, and that linear-algebraic manipulation of these representations can be used to generate one sense over another (\cref{sec:superpos}), as demonstrated in \cref{fig:bat}.
Combined, these two facts offer a possible explanation for the homonym duplication described by \citet{rassin-2022-dalle-2}.\looseness=-1

\section{Background: Diffusion Models}

Text-to-image diffusion models such as \dalle\ \citep{dalle} and Stable Diffusion \citep{stablediffusion} take a text prompt as input and generate an image corresponding to the prompt.
During training, an image $\vb{x}_0$ is transformed into random noise through the progressive addition of noise sampled from a Gaussian distribution whose mean depends on the image's current value, i.e. $\vb{x}_t = \vb{x}_{t-1} + \boldsymbol{\epsilon}_t$, $\boldsymbol{\epsilon}_t \sim \mathcal{N}(\sqrt{\alpha_t}\vb{x}_{t-1},(1-\alpha_t)\vb{I})$ where $\alpha_t$ is a value controlling the magnitude of the noise added at each step.
The model is then trained to gradually denoise these images, step by step, conditioned on a descriptive caption of the image.
Once trained, the model will apply this process to random noise to produce an image consistent with a given prompt.

\subsection{CLIP Encoding}

Both \dalle\ and Stable Diffusion encode text prompts using CLIP \citep{clip}.
CLIP is trained by jointly training a transformer-based text encoder and a transformer-based image encoder on pairs of images and captions with a contrastive objective -- given $N$ pairs, it is trained to maximise the cosine similarity between the text and image encodings for these $N$ pairs, but also to minimise cosine similarity between the $N^2-N$ other pairings.

\subsection{Conditioned Denoising Process} 
Given a prompt $\vb{s}$, diffusion models sample $\vb{x}_T \sim \mathcal{N}(\vb{0}, \vb{I})$ as their initial image.
This image is progressively denoised by taking $\vb{x}_{t-1} = \vb{x}_{t} - A_t\widehat{\boldsymbol{\epsilon}}_\theta(\vb{x}_t, t\mid \vb{s}) + B_t$, where $A_t$ and $B_t$ are coefficients calculated by a scheduler.
To allow generation to be guided by the prompt, $\widehat{\boldsymbol{\epsilon}}_\theta(\vb{x}_t, t\mid \vb{s}) =\boldsymbol{\epsilon}_\theta(\vb{x}_t, t\mid \emptyset) + \gamma \cdot (\boldsymbol{\epsilon}_\theta(\vb{x}_t, t\mid \vb{s}) - \boldsymbol{\epsilon}_\theta(\vb{x}_t, t\mid \emptyset))$, where $\gamma\geq 1$ describes the strength of guidance and $\boldsymbol{\epsilon}_\theta(\vb{x}_t, t\mid \vb{s})$ is the model that has been trained to predict the noise to remove at each step.
$\boldsymbol{\epsilon}_\theta(\vb{x}_t, t\mid \vb{s})$ is generally implemented with a version of U-Net \citep{unet}.

\paragraph{\dalle.}
In \dalle, the prompt $\vb{s}$ is encoded using CLIP, a CLIP image encoding is predicted using a transformer-based diffusion prior model conditioned on $\vb{s}$ and $\clip(\vb{s})$.
The main diffusion model is then conditioned on this image encoding by projecting it to the appropriate dimension and then concatenating it to the attention context at each layer.\looseness=-1

\paragraph{Stable Diffusion.}
In Stable Diffusion, the denoising process takes place entirely in the latent space.
The starting point $\vb{z}_T$ is sampled and then de-noised conditioned on $\clip(\vb{s})$.
The conditioning is implemented through cross-attention between U-Net representations and a projection of $\clip(\vb{s})$.
The final output $\vb{z}_0$ is decoded to an image using the decoder of a Variational Auto-encoder \citep{kingmawelling}.\looseness=-1

\subsection{Homonym Duplication in \dalle}

\citet{rassin-2022-dalle-2} demonstrate a number of strange behaviours exhibited by \dalle.
In this work, we are primarily concerned with the phenomenon of homonym duplication -- when multiple senses of a polysemous word are realised in one image.
They use a number of prompts to show that when a prompt contains a polysemous word such as \prompt{bat} or \prompt{bow}, rather than realising the sense that the model deems to be most likely, it often realises both senses in the final image.
They also point out that this is observed less frequently in smaller models such as DALL\textperiodcentered E 
Mini and Stable Diffusion.\looseness=-1

\subsubsection{Stable Diffusion and \dalle}

In this work we consider Stable Diffusion in lieu of \dalle.
We make this choice since our method directly manipulates the representations used in the generation process, which is not possible in publicly available releases of \dalle.
Both are similar in that they both text-to-image generation models based on diffusion and a CLIP encoder.
They differ primarily in the fact that Stable Diffusion performs the denoising process entirely in the latent space.
\dalle\ is also a much larger model in terms of the number of parameters.
Our hypothesis, if true, would be more difficult to show in a smaller model with less representational capacity.
Because of this, we consider it likely that positive conclusions drawn from this work would also apply to \dalle, though we invite replication of our approach using \dalle.

\subsubsection{Homonym Duplication in Stable Diffusion}

\begin{figure}
    \centering
    \begin{subfigure}{0.23\textwidth}
        \centering
        \includegraphics[width=0.9\textwidth]{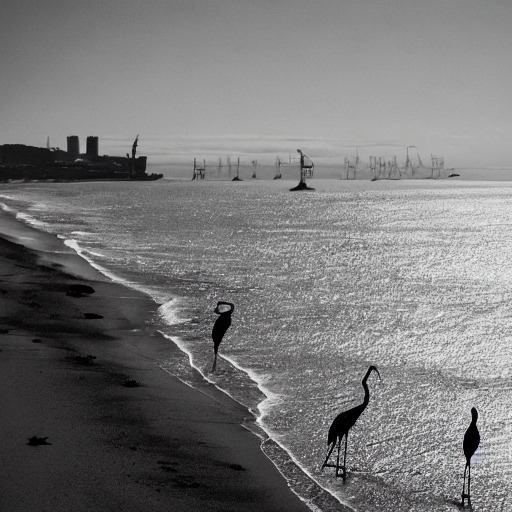}
        \caption{Prompt: \prompt{tall cranes by the sea}}
        \label{fig:crane_dup_body}
    \end{subfigure}
    \hfill
    \begin{subfigure}{0.23\textwidth}
        \centering
        \includegraphics[width=0.9\textwidth]{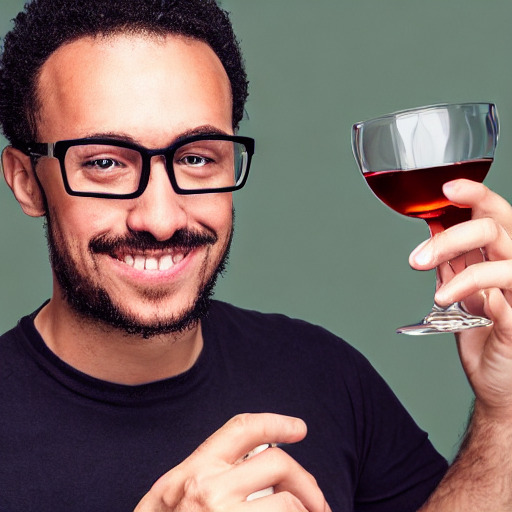}
        \caption{Prompt: \prompt{a man holding glasses}}
        \label{fig:glasses_dup}
    \end{subfigure}
    \begin{subfigure}{0.23\textwidth}
        \centering
        \includegraphics[width=0.9\textwidth]{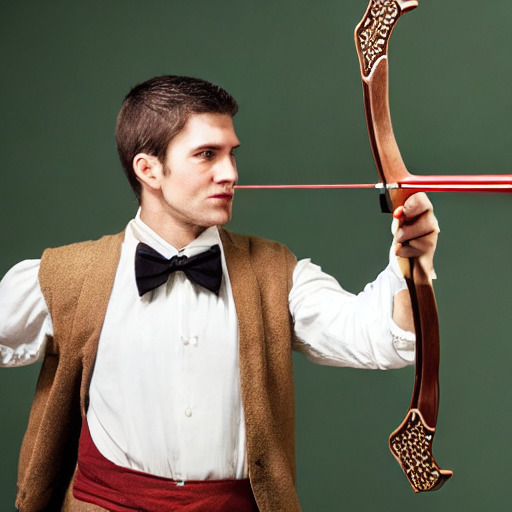}
        \caption{Prompt: \prompt{a gentleman with a bow and arrow}}
        \label{fig:bow_dup_body}
    \end{subfigure}
    \hfill
    \begin{subfigure}{0.23\textwidth}
        \centering
        \includegraphics[width=0.9\textwidth]{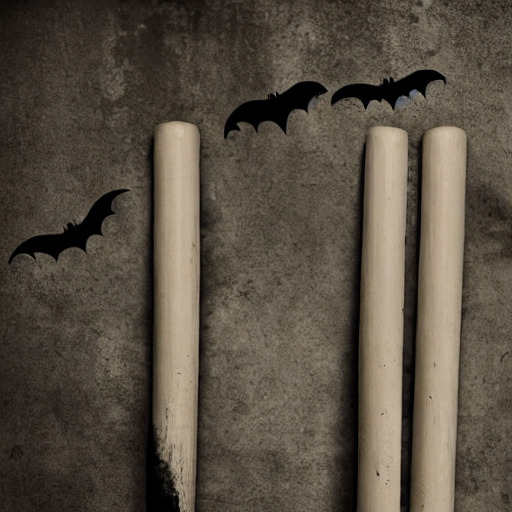}
        \caption{Prompt: \prompt{a baseball bat inside a spooky cave}}
        \label{fig:bat_cave}
    \end{subfigure}
    \caption{Examples of homonym generation observed in images generated using Stable Diffusion}
    \label{fig:hom_dup_rep}
\end{figure}

As \citet{rassin-2022-dalle-2} point out, the homonym duplication phenomenon that they observe in \dalle\ is seen less frequently in Stable Diffusion.
Nonetheless, we were successful in replicating the behaviour for some prompts.
Some of these images are shown in \cref{fig:bat} and \cref{fig:hom_dup_rep}.
More are included in \cref{subsec:moredup}.\looseness=-1
\section{Claim 1: Diffusion can realise multiple concepts in superposition }\label{sec:sumenc}

To construct an explanation for the phenomenon of homonym duplication, we first show that an encoding that is composed of a sum of encodings of multiple words can produce an image in which each word in the sum is represented.
When using the model in the standard way, the prompt $\vb{s}$ is encoded using CLIP, and the result $\clip(\vb{s})$ is then used to condition the denoising process.
In these experiments, we instead take two prompts $\vb{s}_1$ and $\vb{s}_2$, and the denoising process is conditioned on the weighted sum of their CLIP encodings $\alpha_1 \clip(\vb{s}_1) + \alpha_2\clip(\vb{s}_2)$, where $\alpha_1,\alpha_2\in\mathbb{R}$ and $\alpha_1+\alpha_2=1$.\footnote{In practice, we achieved best results with $\alpha_1=\alpha_2=0.5$.}

\subsection{Experimental Results}

\begin{figure}[!t]
    \centering
    \begin{subfigure}{0.23\textwidth}
        \centering
        \includegraphics[width=0.9\textwidth]{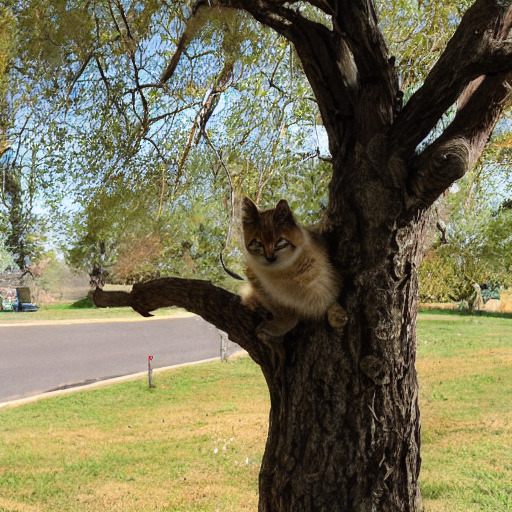}
        \caption{$(0.5\cdot \clip(\text{\prompt{cat}}))$ $+(0.5\cdot\clip(\text{\prompt{tree}}))$}
        \label{fig:treecat}
    \end{subfigure}
    \hfill
    \begin{subfigure}{0.23\textwidth}
        \centering
        \includegraphics[width=0.9\textwidth]{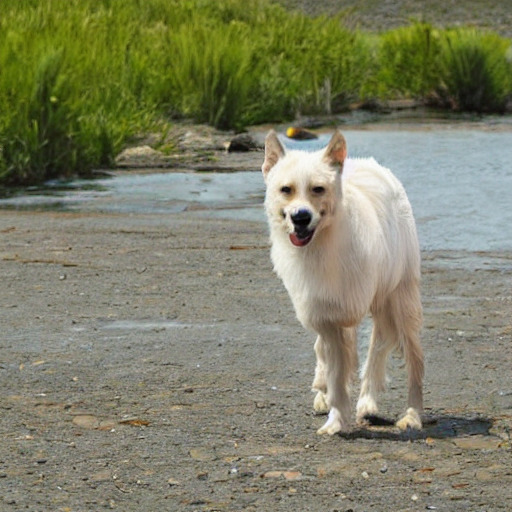}
        \caption{$(0.5\cdot \clip(\text{\prompt{dog}}))$ $+(0.5\cdot\clip(\text{\prompt{lake}}))$}
        \label{fig:lakedog}
    \end{subfigure}
    \begin{subfigure}{0.23\textwidth}
        \centering
        \includegraphics[width=0.9\textwidth]{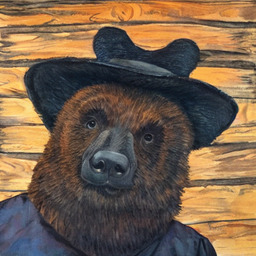}
        \caption{$(0.5\cdot \clip(\text{\prompt{bear}}))$ $+(0.5\cdot\clip(\text{\prompt{hat}}))$}
        \label{fig:bearhatfig}
    \end{subfigure}
    \hfill
    \begin{subfigure}{0.23\textwidth}
        \centering
        \includegraphics[width=0.9\textwidth]{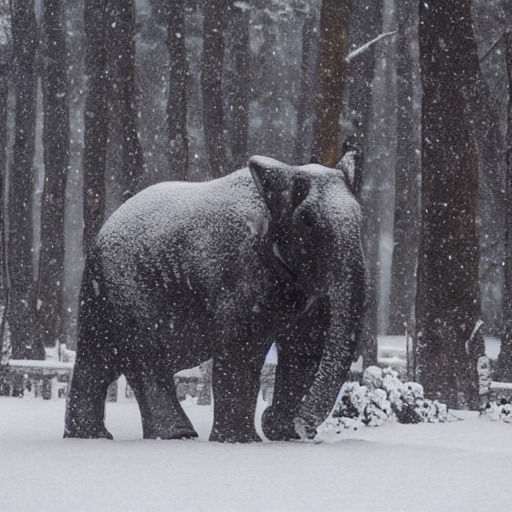}
        \caption{$(0.5\cdot \clip(\text{\prompt{elephant}}))$ $+(0.5\cdot\clip(\text{\prompt{snow}}))$}
        \label{fig:elephantsnowfig}
    \end{subfigure}
    \begin{subfigure}{0.23\textwidth}
        \centering
        \includegraphics[width=0.9\textwidth]{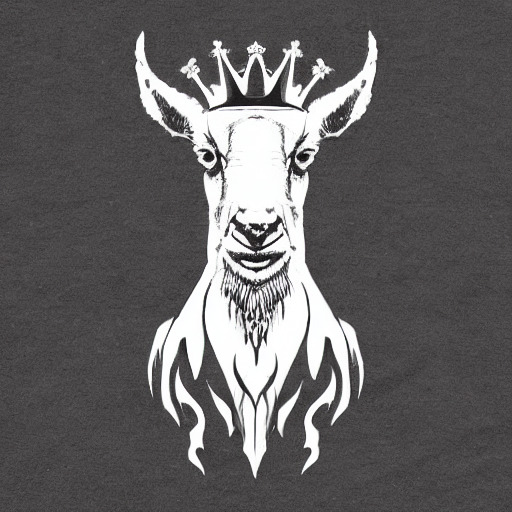}
        \caption{$(0.5\cdot \clip(\text{\prompt{goat}}))$ $+(0.5\cdot\clip(\text{\prompt{crown}}))$}
        \label{fig:goatcrownfig}
    \end{subfigure}
    \hfill
    \begin{subfigure}{0.23\textwidth}
        \centering
        \includegraphics[width=0.9\textwidth]{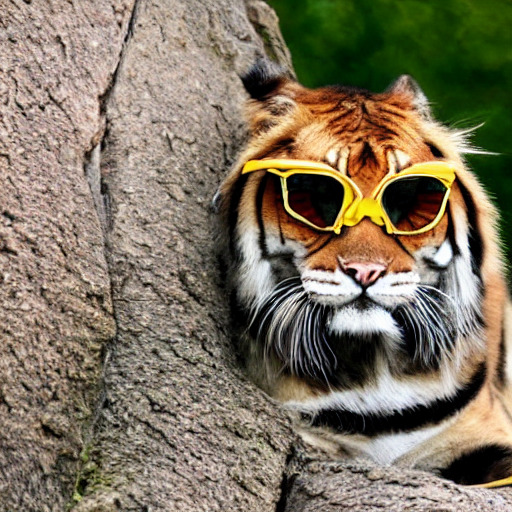}
        \caption{$(0.5\cdot \clip(\text{\prompt{tiger}}))$ $+(0.5\cdot\clip(\text{\prompt{glasses}}))$}
        \label{fig:tigerglassesfig}
    \end{subfigure}
    \caption{Images generated by summing representations of multiple prompts}
    \label{fig:sumgen}
\end{figure}

For each pair of prompts, $\vb{s}_1$ and $\vb{s}_2$, 30 images were generated from $\alpha_1\clip(\vb{s}_1) + \alpha_2\clip(\vb{s}_2)$, 30 images were generated from $\clip(\vb{s}_1)$ and 30 images were generated from $\clip(\vb{s}_2)$.
Results presented here are for pairs that were observed to result in realisation of both prompts.
Some of the images generated using this method are shown in \cref{fig:sumgen}, and \cref{tab:sum_results} shows the percentage of images generated that contained the object described by $\vb{s}_1$, that described by $\vb{s}_2$, both or neither.
We can see that the model can, in fact, generate an image representing both of the words in the sum.
In fact, we see that this occurs in over 40\% of generated images.
More generated images are included in \cref{subsubsec:realise_sum} and a detailed breakdown of the contents of the images generated for each pair of prompts, along with details of statistical significance tests are given in \cref{subsec:sum_stats}.\looseness=-1

There is no particular reason to assume that summing the representations of two words would result in a representation that can be interpreted by the model.
Adding the representations in this way could result in a representation that doesn't correspond to anything at all, and leads to the generation of low quality images.
It's also plausible that it would produce an interpolation between the two concepts described.
However, we can see that this is not the case.
The model sometimes produced both of the objects described in the two prompts, and otherwise largely produced one or the other.
This indicates that the representations are encoded in such a way that the important information corresponding to each prompt does not interfere when the two are summed, potentially because they are stored in directions that are close to orthogonal.

As an additional observation on this point, we found that some pairs, such as $\vb{s}_1=$\prompt{cat}, $\vb{s}_2=$\prompt{dog}, did not behave in the same way, instead producing a cat-dog hybrid creature.
We speculate that this may be because similar concepts such as \prompt{cat} and \prompt{dog} are likely to share similar representation space.
\begin{table}
\footnotesize
\setlength{\tabcolsep}{0.5em}
\setlength{\defaultaddspace}{0.5em}
    \centering
    \begin{tabular}{ccccc}\toprule
        & \multicolumn{4}{c}{Percentage containing:}\\
        Input & $\vb{s}_1$ & $\vb{s}_2$ & Both & Neither \\\midrule
        $\clip(\vb{s}_1)$ & 100.0 & 0 & 0 & 0\\
        \addlinespace
        $\clip(\vb{s}_2)$  & 0 & 100.0 & 0 & 0\\
        \begin{tabular}{c}$0.5\cdot(\clip(\vb{s}_1))+$\\$0.5\cdot(\clip(\vb{s}_2))$\end{tabular}& 35.2 & 20 & 41.5 & 3.3 \\\bottomrule
    \end{tabular}
    \caption{Table showing which prompts were represented in images generated from each prompt in a pair and from their weighted sum}
    \label{tab:sum_results}
\end{table}
\section{Claim 2: Polysemous words are represented as linear superpositions of meanings}\label{sec:superpos}

Next, we seek to show that the CLIP encoder used by Stable Diffusion represents polysemous word as a linear superposition of their possible meanings.
Other methods of encoding words have been previously shown to represent polysemy in this way \citep{arora-etal-2018-linear}.
In this case, since the encoder is transformer-based, there is no static representation of each word, but instead a representation of each token of the word in the context of a sentence. 
Due to this, we instead approximate directions corresponding to each meaning across several sentences.
For a particular word of interest, we take several sets of syntactically similar sentences where each sentence contains either an ambiguous use of the word, or a use where one sense is more likely than the other.
A procedure similar to that used by \citet{bolukbasi} is then used to obtain bases for the spaces corresponding to the difference between a representation of a particular meaning and the representation of the word in the corresponding ambiguous sentence.
We then use projections into these spaces to obtain approximations of the representation of each meaning.
We then demonstrate how these approximations can be used to intervene on a representation and change the sense that appears in generated images.
This involves editing the representation in the subspace spanned by our approximate representations, and removing the contribution from the representation of the undesired meaning.

\subsection{Identifying a Meaning Subspace}
\begin{figure}[t]
    \centering
    \begin{subfigure}{0.45\textwidth}
        \centering
        \includegraphics[width=0.3\textwidth]{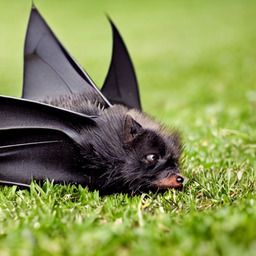}
        \includegraphics[width=0.3\textwidth]{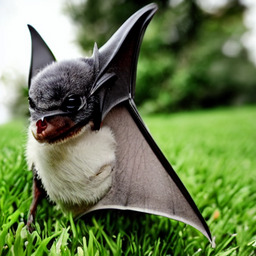}
        \includegraphics[width=0.3\textwidth]{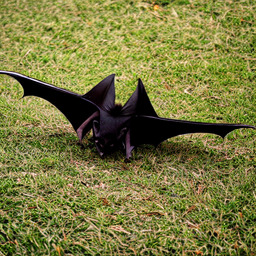}
        \caption{Unedited prompt encoding}
        \label{fig:bat_amb_body}
    \end{subfigure}
    \begin{subfigure}{0.45\textwidth}
        \centering
        \includegraphics[width=0.3\textwidth]{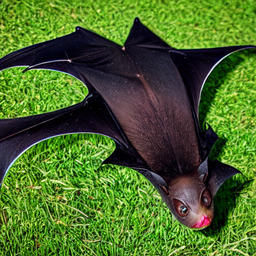}
        \includegraphics[width=0.3\textwidth]{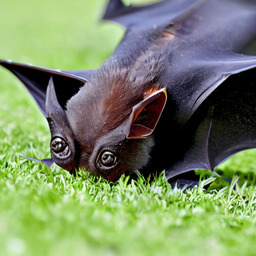}
        \includegraphics[width=0.3\textwidth]{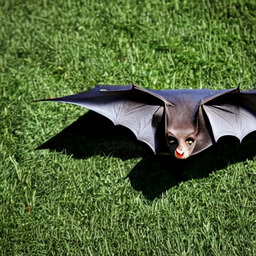}
        \caption{Encoding edited to favour the animal sense}
        \label{fig:bat_2_body}
    \end{subfigure}
    \begin{subfigure}{0.45\textwidth}
        \centering
        \includegraphics[width=0.3\textwidth]{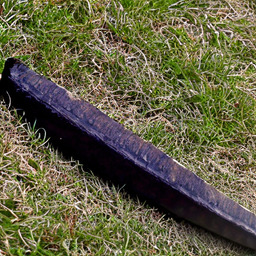}
        \includegraphics[width=0.3\textwidth]{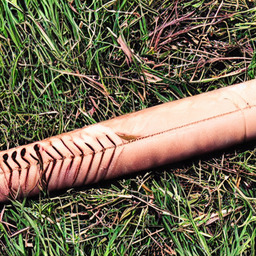}
        \includegraphics[width=0.3\textwidth]{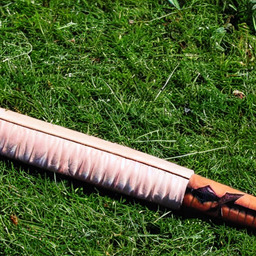}
        \caption{Encoding edited to favour the sports-related sense}
        \label{fig:bat_1_body}
    \end{subfigure}
    \caption{Prompt: \prompt{a bat laying on the grass}}
    \label{fig:bat_edit_body}
\end{figure}

\begin{figure}[t]
    \centering
    \begin{subfigure}{0.45\textwidth}
        \centering
        \includegraphics[width=0.3\textwidth]{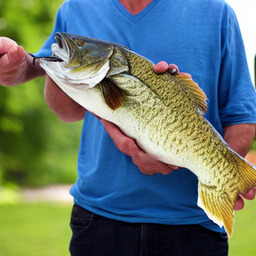}
        \includegraphics[width=0.3\textwidth]{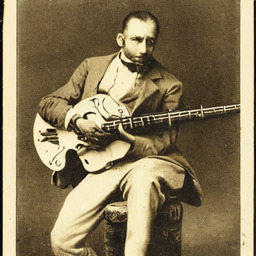}
        \includegraphics[width=0.3\textwidth]{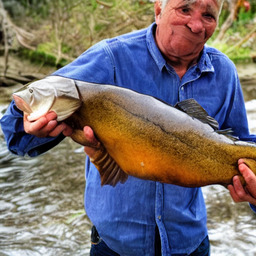}
        \caption{Unedited prompt encoding}
        \label{fig:bass_amb_body}
    \end{subfigure}
    \begin{subfigure}{0.45\textwidth}
        \centering
        \includegraphics[width=0.3\textwidth]{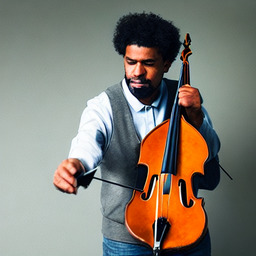}
        \includegraphics[width=0.3\textwidth]{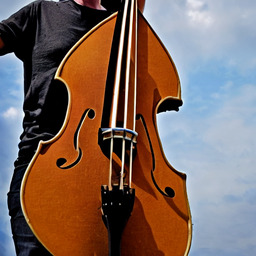}
        \includegraphics[width=0.3\textwidth]{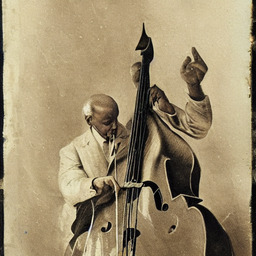}
        \caption{Encoding edited to favour the music-related sense}
        \label{fig:bass_1_body}
    \end{subfigure}
        \begin{subfigure}{0.45\textwidth}
        \centering
        \includegraphics[width=0.3\textwidth]{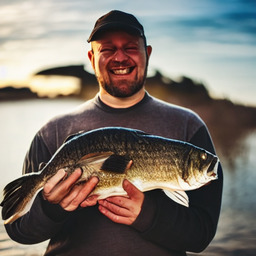}
        \includegraphics[width=0.3\textwidth]{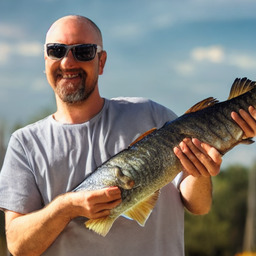}
        \includegraphics[width=0.3\textwidth]{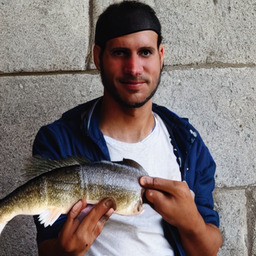}
        \caption{Encoding edited to favour the fish sense}
        \label{fig:bass_2_body}
    \end{subfigure}
    \caption{Prompt: \prompt{a man holding a bass}}
    \label{fig:bass_edit_body}
\end{figure}
In a procedure based on that described by \citet{bolukbasi}, we begin by identifying a subspace in which representations of different meanings differ.
Given a polysemous word $w_p$ with distinct possible meanings, $\mone$ and $\mtwo$, we hand-write a set of $N$ prompts containing $w_p$ in which either of the two candidate meanings is possible.
From each such prompt, $\samb{}$, we manually create $\sone{}$ and $\stwo{}$, which differ from $\samb{}$ by the addition of a disambiguating modifier and changes to the lexical content to more strongly imply meanings $\mone$ and $\mtwo$ respectively, but otherwise have the same syntactic structure.
For example, 
\begin{align}
    \samb{}&=\text{\prompt{a \colamb{bat} in a box}}\nonumber\\
    \sone{}&=\text{\prompt{a vampire \colone{bat} in a cave}}\nonumber\\
    \stwo{}&=\text{\prompt{a baseball \coltwo{bat} in a locker}}\nonumber
\end{align}
Each sentence is encoded using CLIP and the vector representing $w_p$ is extracted from each to give $V_{w_p}^{\vb{s}_n} = \Big\{\vb{v}_{w_p}^{\samb{n}}, \vb{v}_{w_p}^{\sone{n}}, \vb{v}_{w_p}^{\stwo{n}}\Big\}$, where $\vb{v}_{w_p}^{\vb{s}}= \clip(\vb{s})_{w_p}$.\footnote{In practice, all of the words we used in our experiments were tokenised to a single token, so our description assumes that we have one encoding per word. For a word consisting of multiple tokens, this process would be repeated for the encoding corresponding to each token.}

Next we construct $D^{\ambone}_n = \Big\{\vb{v}_{w_p}^{\samb{n}}, \vb{v}_{w_p}^{\sone{n}}\Big\}$, and calculate the mean of each such set:
\begin{equation}
    \boldsymbol{\mu}^{\ambone}_n = \frac{1}{|D^{\ambone}_n|}\sum_{\vb{v}\in D^{\ambone}_n} \vb{v}
\end{equation} 
A matrix is calculated by summing the outer products of each vector's difference with the mean of its set:\looseness=-1
\begin{align}
    C&^{\ambone}=\\
    &\sum_{n=1}^N \sum_{\vb{v}\in D^{\ambone}_n} (\vb{v} - \boldsymbol{\mu}^{\ambone}_n) (\vb{v} - \boldsymbol{\mu}^{\ambone}_n)^\top\nonumber
\end{align}
The singular value decomposition (SVD) 
gives $C^{\ambone}=U_{\colone{1}}\Sigma_{\colone{1}} V_{\colone{1}}^\top$.
The first $k$ columns of the matrix $U_{\colone{1}}$, $\{\vb{u}^{\ambone}_1,\dots,\vb{u}^{\ambone}_k\}$ give an orthonormal basis for the subspace of difference between $\vb{v}_{w_p}^{\samb{}}$ and $\vb{v}_{w_p}^{\vb{s}^{\sone{}}}$.\footnote{We choose $k$ based on a threshold on the corresponding singular values, specifically we take $k$ to be the smallest integer greater than 2 such that $\frac{\sum_{i=1}^k \sigma_i}{\sum_{i=1}^N \sigma_i}>0.95$, where $\sigma_i$ is the $i$th diagonal entry of $\Sigma_{\colone{1}}$. In practice, for all of our examples this results in $k<5$.} 
To find $\{\vb{u}^{\ambtwo}_1,\dots,\vb{u}^{\ambtwo}_k\}$, this procedure is repeated starting with $D^{\ambtwo}_n = \Big\{ \vb{v}_{w_p}^{\samb{n}}, \vb{v}_{w_p}^{\stwo{n}}\Big\}$.

\subsection{Finding Meaning Directions}

We then use the difference subspaces defined by these bases to approximate vectors corresponding to $\mone$ and $\mtwo$, which we term $\vone$ and $\vtwo$.

The procedure for approximating these vectors begins by finding the average projection of $\vb{v}_{w_p}^{\sone{}}$ into the subspace defined by $\{\vb{u}^{\ambone}_1,\dots,\vb{u}^{\ambone}_k\}$:
\begin{equation}
    \vone^{(0)} = \sum_{j=1}^k \left(\frac{1}{N}\sum_{n=1}^{N}\vb{v}_{w_p}^{\sone{n}}\cdot\vb{u}^{\ambone}_j \right) \vb{u}^{\ambone}_j 
\end{equation}
We then make this orthogonal to all $\vb{v}_{w_p}^{\stwo{}}$, by taking $\widehat{\vone}$ to be $\frac{\vone^{(N)}}{||\vone^{(N)}||}$, where
\begin{equation}
    \vone^{(n)} = \vone^{(n-1)} - \proj(\vone^{(n-1)}, \vb{v}_{w_p}^{\stwo{n}})
\end{equation}
where $\proj(\vb{a},\vb{b})$ denotes the projection of $\vb{a}$ onto $\vb{b}$, i.e. $\proj(\vb{a},\vb{b}) = \frac{\vb{a}\cdot\vb{b}}{\vb{b}\cdot\vb{b}}\vb{b}$.
The vector $\widehat{\vtwo}$ is found analogously.

By examination, we find that representations from ambiguous sentences tend to have negative dot products with both $\widehat{\vone}$ and $\widehat{\vtwo}$ (or occasionally positive dot products with both).
Representations from sentences favouring $\mone$ tend to have a a positive dot product with $\widehat{\vone}$ and a near-zero dot product with $\widehat{\vtwo}$ (by construction), and vice versa for sentences favouring $\mtwo$.
This suggests how we may edit representations to target a desired sense. 

The normalised vectors $\widehat{\vone}$ and $\widehat{\vtwo}$ are not used, instead we take $\vone = \max_n (\vb{v}_{w_p}^{\sone{n}}\cdot\widehat{\vone})\widehat{\vone}$ and $\vtwo = \max_n (\vb{v}_{w_p}^{\stwo{n}}\cdot\widehat{\vtwo})\widehat{\vtwo}$. 
We do this so that the vector also represents the magnitude of projection is this direction that is sufficient to have this sense represented.\looseness=-1

\subsection{Sense Editing Procedure}

If we have a sentence $\vb{s}$ containing $w_p$ then its standard, unedited encoding is given by $\vb{v}_{w_p} = \clip(\vb{s})_{w_p}$.
To edit this representation to push its sense towards $\mtwo$, we consider its projection onto the subspace spanned by $\vone$ and $\vtwo$ and aim to make its dot product with $\vone$ in this subspace near-zero while aligning it with $\vtwo$, based on the observations described in the previous section.

Since $\vone$ and $\vtwo$ are, in general, not orthogonal, we first obtain an orthonormal basis of the subspace that they span: $U=\left\{\frac{\vone}{||\vone||}, \frac{\vtwo - \proj(\vtwo, \frac{\vone}{||\vone||})}{||\vtwo - \proj(\vtwo, \frac{\vone}{||\vone||})||}\right\}$, and then compute
\begin{align}
    &\widetilde{\vb{v}}^{\mtwo}_{w_p} = \vb{v}_{w_p} - \proj_U\vb{v}_{w_p} \\
    &+ \frac{||\vtwo||}{||\vtwo- \proj(\vtwo,\vone)||}(\vtwo - \proj(\vtwo,\vone))\nonumber
\end{align}
where $\proj_U\vb{v}$ is the projection onto the subspace defined by $U$, i.e. $\proj_U\vb{v} = \sum_{\vb{u}\in U} (\vb{v}\cdot \vb{u})\vb{u}$.

Subtracting $\proj_U\vb{v}_{w_p}$ removes the entire projection of the vector onto this subspace, including its component along $\vone$, making it orthogonal to $\vone$ as desired.
We then wish to add $\vtwo$, but since this is not orthogonal to $\vone$ we remove its projection along $\vone$ and rescale accordingly.
This means that in the subspace defined by $U$, $\widetilde{\vb{v}}^{\mtwo}_{w_p}$ is orthogonal to $\vone$ and has high cosine similarity with $\vtwo$.
This edited representation replaces $\vb{v}_{w_p}$ in the sentence encoding, which is then passed to the model as input.\looseness=-1

\subsection{Experimental Results}
\begin{table}
\footnotesize
    \setlength{\extrarowheight}{1pt} 
    \centering
    \begin{tabularx}{0.5\textwidth}{LL}\toprule        
        \multicolumn{2}{c}{Prompts}  \\\midrule
        a \textbf{bass} & a \textbf{crane} \\
        a man holds a \textbf{bass} & a \textbf{crane} by the ocean \\
        a \textbf{bass} is displayed on a wall & a \textbf{crane} surrounded by nature\\
        a \textbf{bat} & \textbf{glasses} on a table \\
        a \textbf{bat} and a baseball fly through the air & a \textbf{seal}\\
        a boy holds a black \textbf{bat} & a \textbf{seal} on an envelope\\
        a \textbf{bat} laying on the grass & a \textbf{trunk}\\
\bottomrule
    \end{tabularx}
    \caption{Prompts used to evaluate our editing procedure, with the word of interest in bold}
    \label{tab:prompts}
\end{table}
\begin{figure}[!t]
    \centering
    \includegraphics[width=0.45\textwidth]{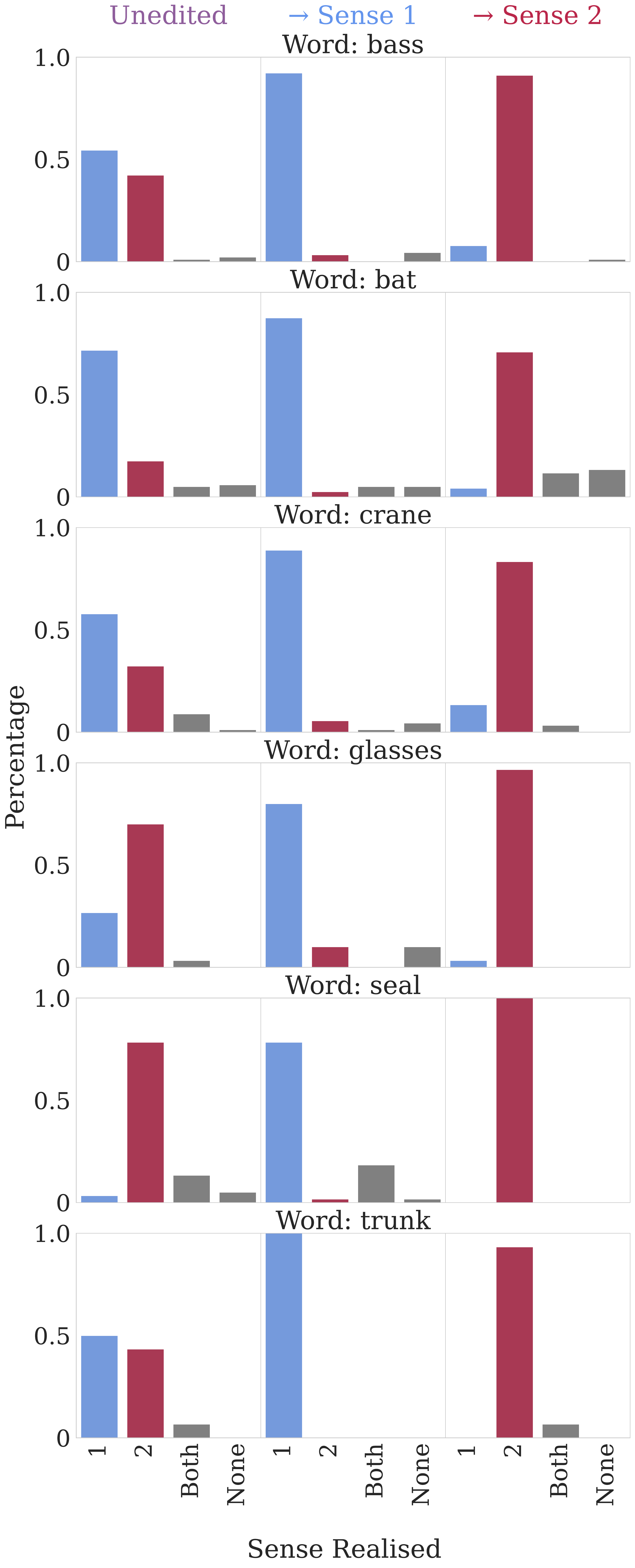}
    \caption{For each word, the proportion of images across all prompts realising each sense is compared across the \colamb{Unedited} encoding, the encoding shifted towards Sense 1 ({\color{sense1} $\rightarrow$ Sense 1}), and the encoding shifted towards Sense 2 ({\color{sense2} $\rightarrow$ Sense 2})}
    \label{fig:shifts}
\end{figure}
The sense editing procedure 
was performed on a set of prompts (shown in \cref{tab:prompts}) that were selected either because they had been observed to exhibit homonym duplication or because the sense that was represented in generated images was not always the same. 
In \cref{fig:bat_edit_body} and \cref{fig:bass_edit_body}, we show some example images generated using an unedited prompt and some using prompts edited to favour one sense over another.
More images are provided in \cref{subsec:app_meaning}.
In \cref{fig:shifts}, we display the proportion of images realising each sense before and after applying this procedure (with {\color{sense1} $\rightarrow$ Sense 1} and {\color{sense2} $\rightarrow$ Sense 2} denoting the encodings edited to favour Sense 1 and Sense 2 respectively).
Detailed numbers and details of statistical significance testing can be found in \cref{subsec:meaning_stats}.
We can see that this procedure is successful in shifting the represented sense towards the intended sense.
This suggests that the representations produced by CLIP do indeed incorporate a linear sum over representations of possible meanings, because a representation can be rewritten in terms of its projection into the subspace spanned by  $\vb{v}_{{\color{sense1}1}}$ and $\vb{v}_{{\color{sense2}2}}$ and a component in the nullspace of this projection.
It also shows that manipulating this sum affects the content of the resulting images.

We note that this is, of course, a simplified approach.
The directions obtained are an approximation based on a small number of handwritten sentences, and we intervene only on the polysemous word of interest, not on any of the other encodings in the sentence, which may also contain relevant information.
Nonetheless, that such a simple and approximate method is capable of achieving such results is a signal in itself.

\section{An Explanation for Homonym Duplication}

Combining the two claims that we have demonstrated, we arrive at a possible explanation of homonym duplication: words with multiple meanings are represented by a sum over multiple meanings, and both of these are then represented in the generated image.  
The difficulty inherent to dealing with words with potentially ambiguous meanings is by no means unique to text-to-image diffusion models, and, indeed, previous work has shown that similarly structured representations can arise through other methods \citet{arora-etal-2018-linear}.
So, it is natural to ask why the phenomenon of duplication has not been observed in text-based models that use such representations.
For example, consider a machine translation model translating the phrase \prompt{a bat} from English to French.
The English word \prompt{bat} can correspond to one of multiple words in French,\ryan{I would list them. } depending on which sense is intended.
Although it is possible that the model's representation may include contributions from both\ryan{You switched from multiple to both, which implies two. I would just name the two senses in this example.} possible meanings, when the decoder comes to generate an output word corresponding to \prompt{bat}, this representation will be converted into a distribution over output tokens and only one is sampled, corresponding to one of the two senses.
This functionally represents the model selecting that sense over the other.
When the next output token is sampled, the model's output distribution is affected by the previous output and accordingly is very unlikely to generate further tokens corresponding to the other sense as this would produce nonsensical sentences unlike anything seen in training.
Such models may produce the incorrect sense in a translation, but they have not been observed to produce both in the way diffusion models do.

In contrast, diffusion models use prompt representations to condition the denoising process.
This process produces probability distributions over noise to be removed from the image at each step.
This means that, unlike the translation model, the diffusion model is effectively never forced to sample from the possible meanings to select one over the other, and both can be represented in the resultant image.
We have seen this to be the case in \cref{sec:sumenc}.
This difference is likely why we see this unique behaviour in diffusion models and not in text-based models using similarly-structured representations.

\section{Stable Diffusion and \dalle}

\citet{rassin-2022-dalle-2} found that smaller models exhibit homonym duplication much less frequently than larger ones.
The superposition explanation is consistent with this finding: larger models have a larger representational capacity, and, thus, are capable of representing more features without interference.
As shown by \citet{elhage2022superposition}, models with more limited representational capacity may not represent features that do not prove to be sufficiently important.
This suggests that smaller models may cease to represent a meaning that is improbable, and, in effect, one meaning is selected, whereas larger models are capable of continuing to represent alternative possible meanings, even if they are modelled to be unlikely.\looseness=-1

It is also possible that this difference is a result of the two different processes followed by the models in training and in generating a new image.
For a training pair $(\vx, \vy)$,
Stable Diffusion uses a variational autoencoder \citep{kingmawelling} 
to obtain $\vz_0$, a representation of $\vx$ in latent space.
The denoising process is then trained to recover $\vz_0$ from a noised $\vz_T$, conditioned on $\clip(\vy)$.
Since the latent space is learned through autoencoding, it is likely that it will prioritise encoding information necessary to reconstruct the image, and there is no reason why distinct concepts described by a polysemous word should be encoded similarly.
Thus, the denoising process of Stable Diffusion may learn to discard this information.

\section{Related Work}

Previous work has investigated models' learned representations through the lens of superposition.
\citet{elhage2022superposition} conducted a controlled investigation into superposition in learned representations in feedforward neural networks.
They investigated the tendency for a learned feature in a network to correspond to multiple features in superposition, and demonstrated how this allows for efficient storage of important features, allowing a representation to encode more features than it has dimensions.
Previous work by \citet{arora-etal-2018-linear} has also found that learned representations of words can represent polysemous words as a linear superposition of possible meanings.\looseness=-1

Work modelling inference of the meaning of an ambiguous word also offers some insight.
\citet{erk-herbelot-2021-marry} model this process using a probabilistic graphical model, where meanings are latent variables that are inferred using the information provided by other words, such as the topical context and semantic restrictions.
From this perspective, beliefs about the meaning of an ambiguous word are always viewed as a distribution over possible meanings, even if it is skewed so that one is much more likely.
If we view computational models as similarly inferring a probability distribution over meanings, text-based decoding models are ultimately forced to sample from the resulting distribution, whereas diffusion models maintain both possibilities throughout the process of generating an image and thus can produce an image containing both.\looseness=-1

\section{Conclusion}
In this work, we set out a possible explanation for the phenomenon of homonym duplication in text-to-image diffusion models as described by \citet{rassin-2022-dalle-2}.
We showed that summing representations of two distinct prompts and using this as input to the denoising process often leads to images in which both prompts are realised.
This suggests that prompts are encoded in such a way that representations of distinct concepts can be summed without substantially interfering with each other.
It also demonstrates that the denoising process does not restrict diffusion models to representing only one meaning.\looseness=-1

We further described a procedure for manipulating which sense of a polysemous word appears in a generated image, based on the assumption that a representation of such a word contains linear contributions from each possible meaning.
We demonstrated this procedure on several prompts, including some that exhibit homonym duplication, finding that it substantially increases the number of generated images containing the targeted sense, even when this sense is only rarely produced using the unedited representation.
The success of this manipulation method provides support for the hypothesised structure of the representations.

We suggest that these facts combine to offer a plausible explanation of homonym duplication.

\bibliography{anthology,custom}
\bibliographystyle{acl_natbib}

\newpage
\appendix
\onecolumn

\section{More Generated Images}

In this section we provide a selection of additional images generated during our experiments. 

\subsection{Homonym Duplication}\label{subsec:moredup}

\begin{figure}[h]
    \centering
    \begin{subfigure}{0.45\textwidth}
        \centering
        \includegraphics[width=0.45\textwidth]{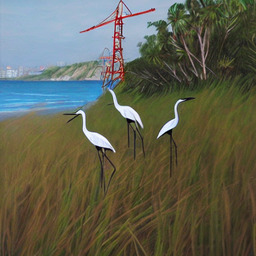}
        \includegraphics[width=0.45\textwidth]{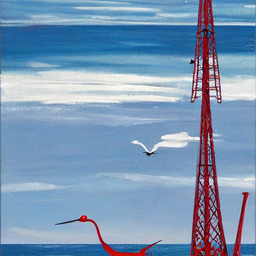}\\
        \includegraphics[width=0.45\textwidth]{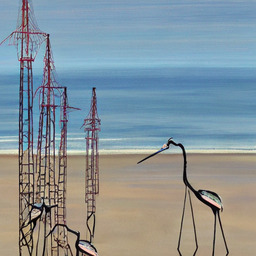}
        \includegraphics[width=0.45\textwidth]{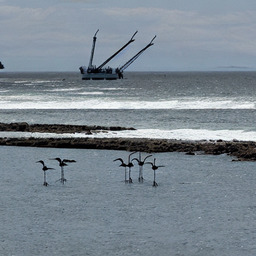}
        \caption{Prompt: \prompt{tall cranes by the ocean}}
        \label{fig:cranedup}
    \end{subfigure}
    \begin{subfigure}{0.45\textwidth}
        \centering
        \includegraphics[width=0.45\textwidth]{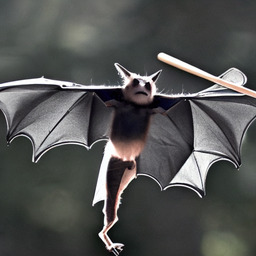}
        \includegraphics[width=0.45\textwidth]{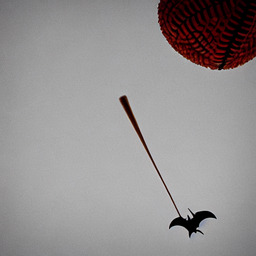}\\
        \includegraphics[width=0.45\textwidth]{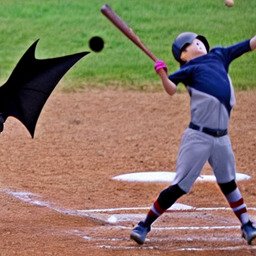}
        \includegraphics[width=0.45\textwidth]{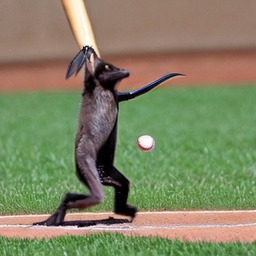}
        \caption{Prompt: \prompt{a bat and a baseball fly through the air}}
        \label{fig:batdup}
    \end{subfigure}
    \begin{subfigure}{0.45\textwidth}
        \centering
        \includegraphics[width=0.45\textwidth]{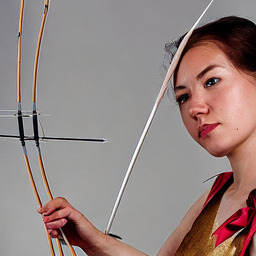}
        \includegraphics[width=0.45\textwidth]{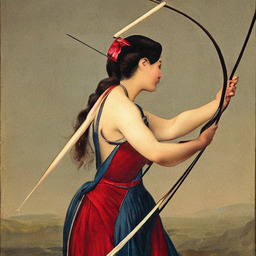}
        \caption{Prompt: \prompt{a woman with a silk bow and arrow}}
        \label{fig:bowdup}
    \end{subfigure}
    \caption{Examples of homonym duplication}
    \label{fig:homdup}
\end{figure}

\Cref{fig:homdup} displays further examples of homonym duplication seen in Stable Diffusion.

\subsection{Summing Encodings}

Here we include further examples of images generated from the weighted sum of two prompt encodings as described in \cref{sec:sumenc}.

\subsubsection{Both Concepts Realised}\label{subsubsec:realise_sum}

\begin{figure}
    \centering
    \begin{subfigure}{\textwidth}
        \centering
        \includegraphics[width=0.14\textwidth]{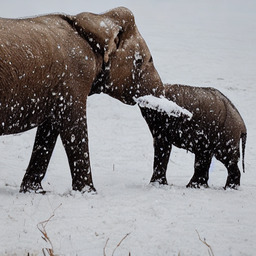}
        \includegraphics[width=0.14\textwidth]{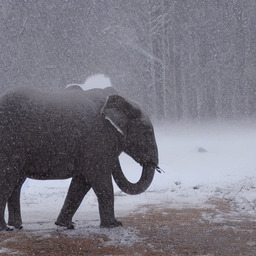}
        \includegraphics[width=0.14\textwidth]{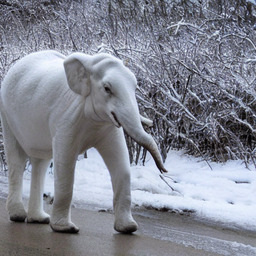}
        \includegraphics[width=0.14\textwidth]{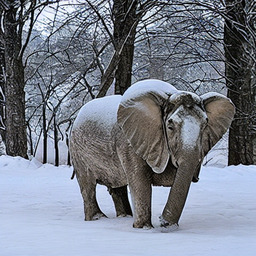} 
        \includegraphics[width=0.14\textwidth]{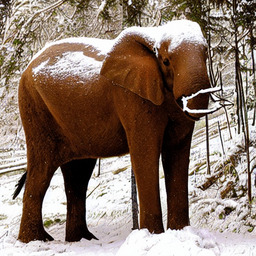}
        \includegraphics[width=0.14\textwidth]{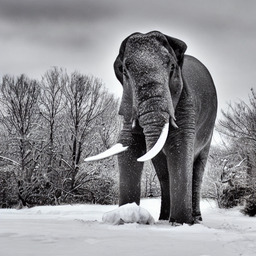}
        \caption{$(0.5\cdot \clip(\text{\prompt{elephant}})) + (0.5\cdot\clip(\text{\prompt{snow}}))$}
        \vspace*{2mm}
        \label{fig:elephantsnow}
    \end{subfigure}
    \begin{subfigure}{\textwidth}
        \centering
        \includegraphics[width=0.14\textwidth]{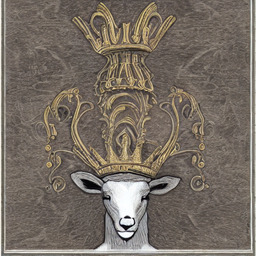}
        \includegraphics[width=0.14\textwidth]{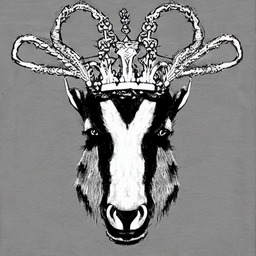}
        \includegraphics[width=0.14\textwidth]{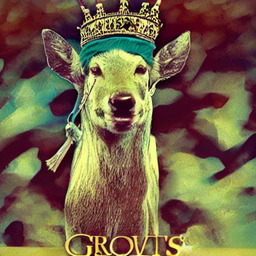}
        \includegraphics[width=0.14\textwidth]{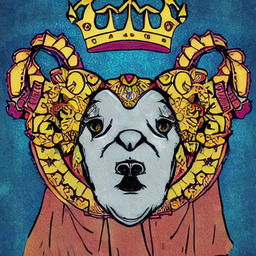} 
        \includegraphics[width=0.14\textwidth]{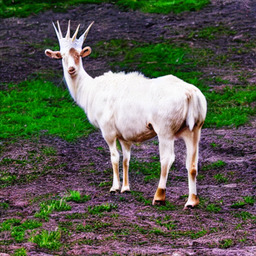}
        \includegraphics[width=0.14\textwidth]{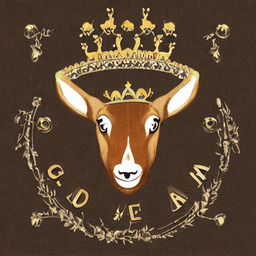}
        \caption{$(0.5\cdot \clip(\text{\prompt{goat}})) + (0.5\cdot\clip(\text{\prompt{crown}}))$}
        \vspace*{2mm}
        \label{fig:goatcrown}
    \end{subfigure}
     \begin{subfigure}{\textwidth}
        \centering
        \includegraphics[width=0.14\textwidth]{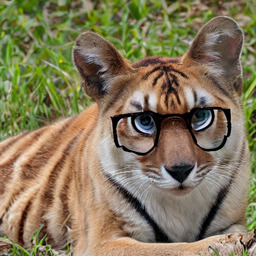}
        \includegraphics[width=0.14\textwidth]{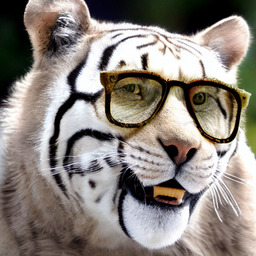} 
        \includegraphics[width=0.14\textwidth]{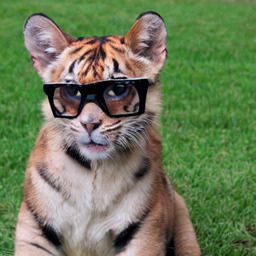}
        \includegraphics[width=0.14\textwidth]{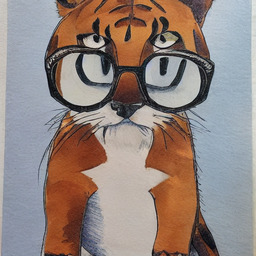}
        \includegraphics[width=0.14\textwidth]{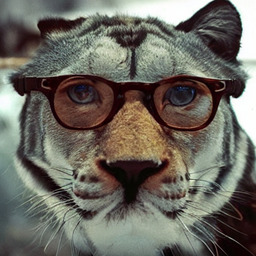} 
        \includegraphics[width=0.14\textwidth]{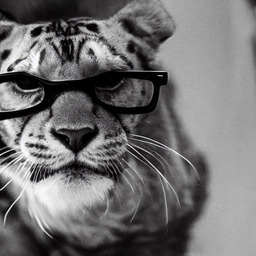}
        \caption{$(0.5\cdot \clip(\text{\prompt{tiger}})) + (0.5\cdot\clip(\text{\prompt{glasses}}))$}
        \vspace*{2mm}
        \label{fig:tigerglasses}
    \end{subfigure}
    \begin{subfigure}{\textwidth}
        \centering
        \includegraphics[width=0.14\textwidth]{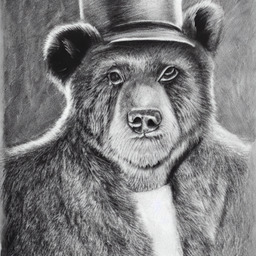}
        \includegraphics[width=0.14\textwidth]{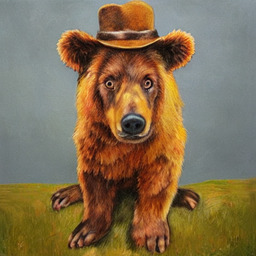} 
        \includegraphics[width=0.14\textwidth]{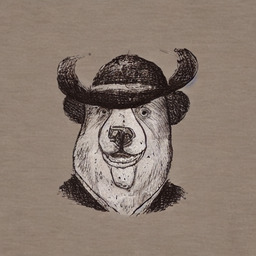}
        \includegraphics[width=0.14\textwidth]{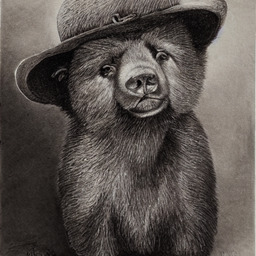}
        \includegraphics[width=0.14\textwidth]{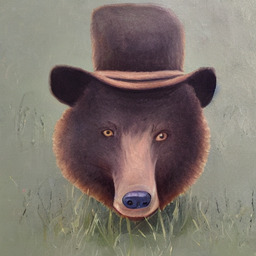} 
        \includegraphics[width=0.14\textwidth]{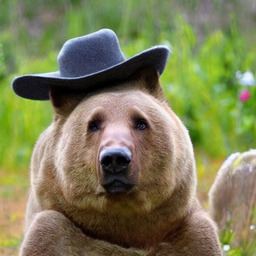}
        \caption{$(0.5\cdot \clip(\text{\prompt{bear}})) + (0.5\cdot\clip(\text{\prompt{hat}}))$}
        \label{fig:bearhat}
    \end{subfigure}
    \begin{subfigure}{\textwidth}
        \centering
        \includegraphics[width=0.14\textwidth]{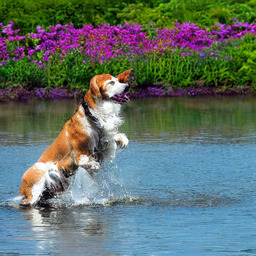}
        \includegraphics[width=0.14\textwidth]{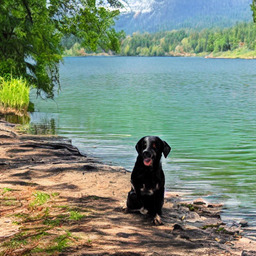}
        \includegraphics[width=0.14\textwidth]{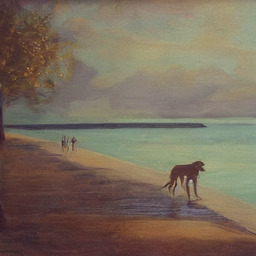}
        \includegraphics[width=0.14\textwidth]{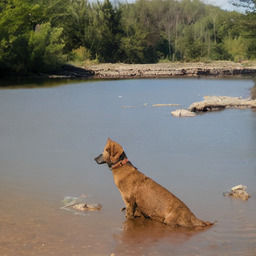}
        \includegraphics[width=0.14\textwidth]{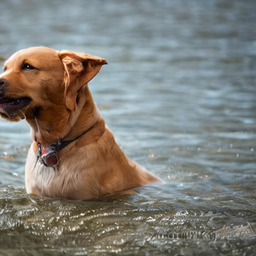} 
        \includegraphics[width=0.14\textwidth]{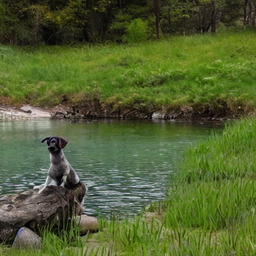}
        \caption{$(0.5\cdot \clip(\text{\prompt{dog}})) + (0.5\cdot\clip(\text{\prompt{lake}}))$}
        \vspace*{2mm}
        \label{fig:doglake}
    \end{subfigure}
    \begin{subfigure}{\textwidth}
        \centering
        \includegraphics[width=0.14\textwidth]{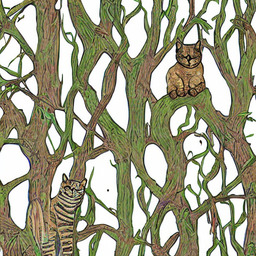}
        \includegraphics[width=0.14\textwidth]{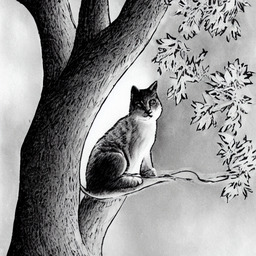}
        \includegraphics[width=0.14\textwidth]{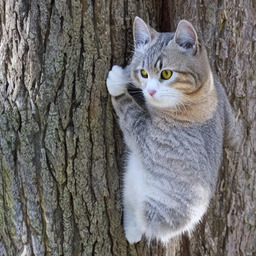}
        \includegraphics[width=0.14\textwidth]{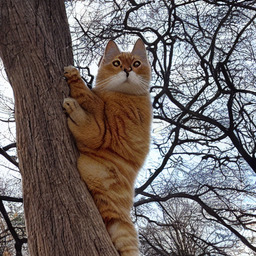}
        \includegraphics[width=0.14\textwidth]{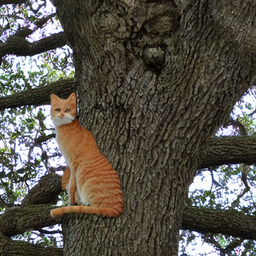} 
        \includegraphics[width=0.14\textwidth]{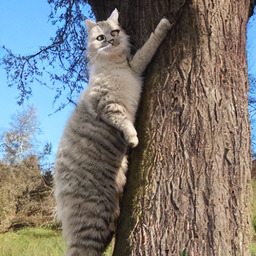}
        \caption{$(0.5\cdot \clip(\text{\prompt{cat}})) + (0.5\cdot\clip(\text{\prompt{tree}}))$}
        \vspace*{2mm}
        \label{fig:cattree}
    \end{subfigure}
    \begin{subfigure}{0.3\textwidth}
        \centering
        \includegraphics[width=0.45\textwidth]{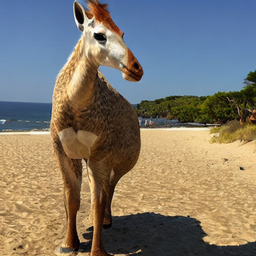}
        \includegraphics[width=0.45\textwidth]{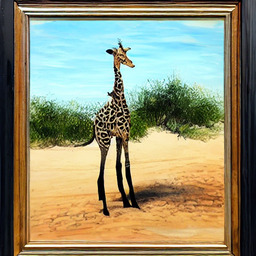}\\
        \includegraphics[width=0.45\textwidth]{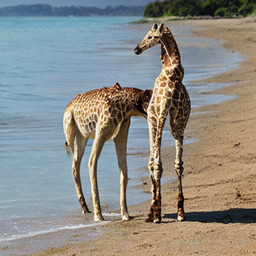}
        \includegraphics[width=0.45\textwidth]{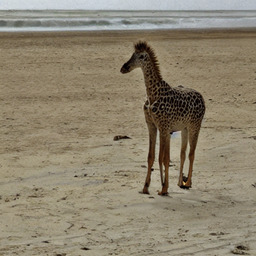}
        \caption{$(0.5\cdot \clip(\text{\prompt{giraffe}}))$\\ $+ (0.5\cdot\clip(\text{\prompt{beach}}))$}
        \label{fig:giraffebeach}
    \end{subfigure}
    \begin{subfigure}{0.3\textwidth}
        \centering
        \includegraphics[width=0.45\textwidth]{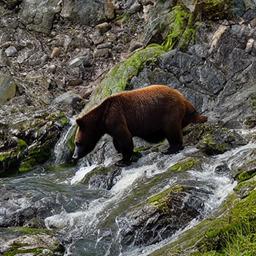}
        \includegraphics[width=0.45\textwidth]{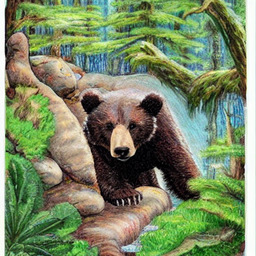}\\ 
        \includegraphics[width=0.45\textwidth]{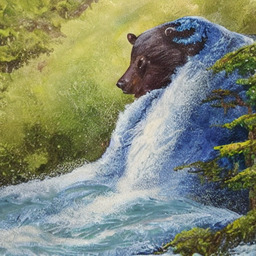}
        \includegraphics[width=0.45\textwidth]{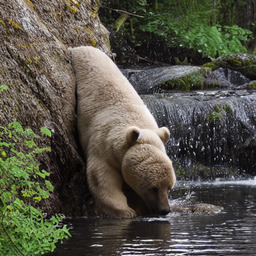}
        \caption{$(0.5\cdot \clip(\text{\prompt{bear}}))$\\ $+ (0.5\cdot\clip(\text{\prompt{waterfall}}))$}
        \label{fig:bearwaterfall}
    \end{subfigure}
    \begin{subfigure}{0.3\textwidth}
        \centering
        \includegraphics[width=0.45\textwidth]{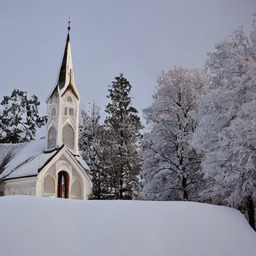}
        \includegraphics[width=0.45\textwidth]{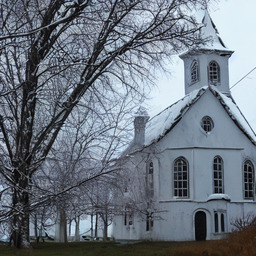}\\ 
        \includegraphics[width=0.45\textwidth]{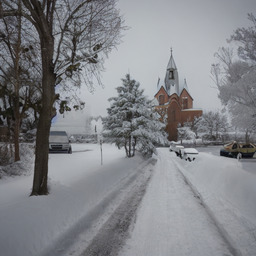}
        \includegraphics[width=0.45\textwidth]{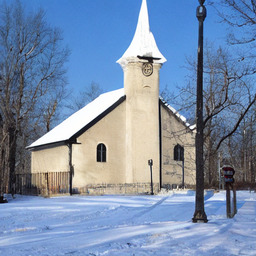}
        \caption{$(0.5\cdot \clip(\text{\prompt{snow}}))$\\ $+ (0.5\cdot\clip(\text{\prompt{church}}))$}
        \label{fig:snowchurch}
    \end{subfigure}
    \caption{Images generated from sums of encodings demonstrating the realisation of both concepts in the sum}
    \label{fig:sum}
\end{figure}

In \cref{fig:sum} we include examples of images where both concepts in the sum are realised in the final image.

\subsubsection{Leakage}

\begin{figure}
    \centering
    \begin{subfigure}{0.45\textwidth}
        \centering
        \includegraphics[width=0.3\textwidth]{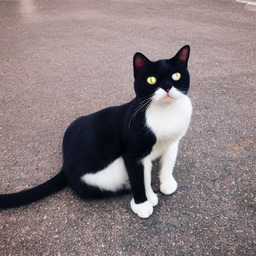}
        \includegraphics[width=0.3\textwidth]{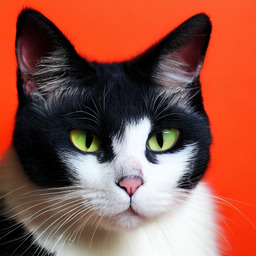} 
        \includegraphics[width=0.3\textwidth]{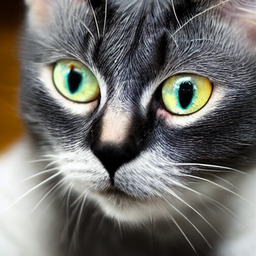}\\
        \includegraphics[width=0.3\textwidth]{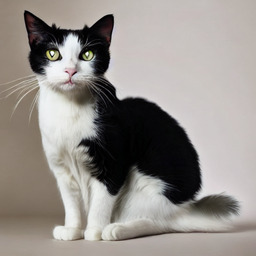}
        \includegraphics[width=0.3\textwidth]{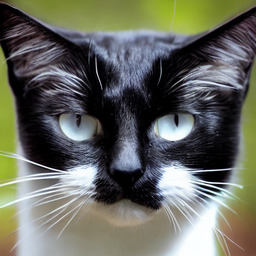} 
        \includegraphics[width=0.3\textwidth]{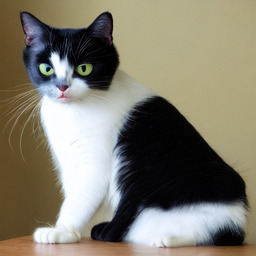}
        \caption{$(0.5\cdot \clip(\text{\prompt{a completely black cat}})) \\+ (0.5\cdot\clip(\text{\prompt{a completely white cat}}))$}
        \label{fig:blackwhite}
    \end{subfigure}
    \begin{subfigure}{0.45\textwidth}
        \centering
        \includegraphics[width=0.3\textwidth]{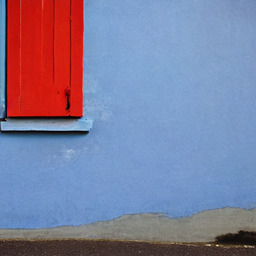}
        \includegraphics[width=0.3\textwidth]{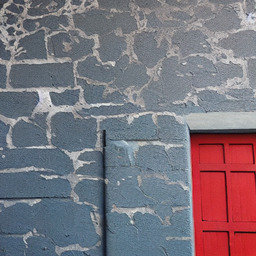} 
        \includegraphics[width=0.3\textwidth]{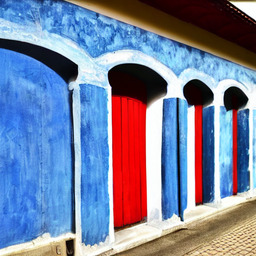}\\
        \includegraphics[width=0.3\textwidth]{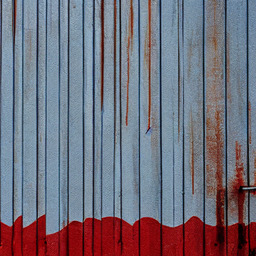}
        \includegraphics[width=0.3\textwidth]{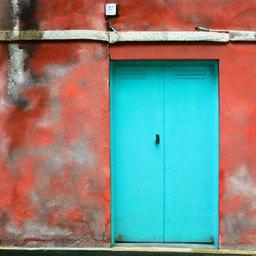} 
        \includegraphics[width=0.3\textwidth]{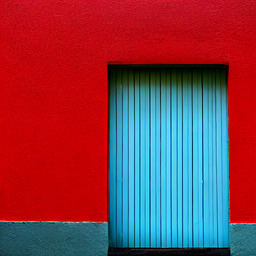}\\
        \caption{$(0.5\cdot \clip(\text{\prompt{a wall painted red}}))\\ + (0.5\cdot\clip(\text{\prompt{a wall painted blue}}))$}
        \label{fig:redblue}
    \end{subfigure}
    \caption{Images generated from sums of encodings that demonstrate concept leakage}
    \label{fig:leakage}
\end{figure}

In \cref{fig:leakage} we include some images that do not fully realise both concepts, but instead allow some kind of concept \prompt{leakage} as a result of the sum.
For example, the prompts \prompt{a completely black cat} and \prompt{a completely white cat}, both overwhelmingly generate images of cats of one colour, as described.
The weighted sum of the two does not generate one completely white cat and one completely black cat.
Instead it generates a cat that is both black and white.

Similarly, the sum of \prompt{a wall painted red} and \prompt{a wall painted blue} does not generate a red wall and a blue wall, or even (for the majority of the time) a wall that is both red and blue.
Instead, it tends to generate a wall that is one of the two colours described, but where something else in the image (such as a door) is of the other colour.

\subsection{Sense Editing}\label{subsec:app_meaning}

In this section we provide more example images demonstrating the effects of our linear subspace procedure as described in \cref{sec:superpos}.

For each prompt we display 15 images sampled from the image generated from the unedited prompt encoding, alongside 15 generated from the encoding after editing it to favour each sense.

\begin{figure}
    \centering
    \begin{subfigure}{\textwidth}
        \centering
        \includegraphics[width=0.15\textwidth]{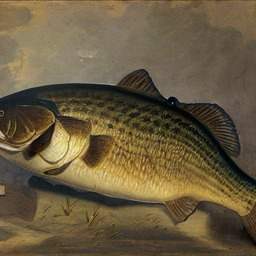}
        \includegraphics[width=0.15\textwidth]{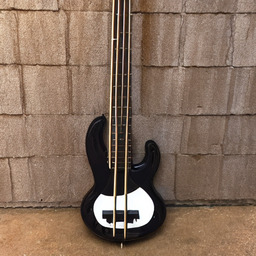}
        \includegraphics[width=0.15\textwidth]{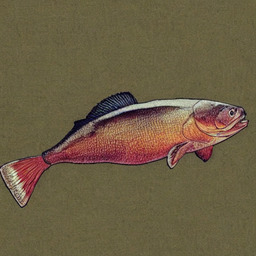}
        \includegraphics[width=0.15\textwidth]{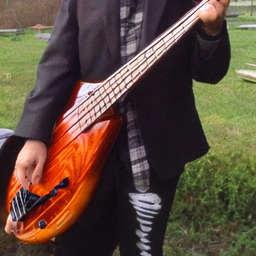}
        \includegraphics[width=0.15\textwidth]{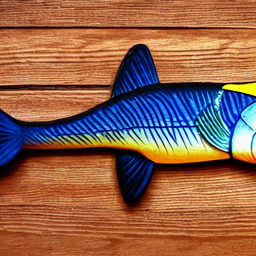}\\
        \includegraphics[width=0.15\textwidth]{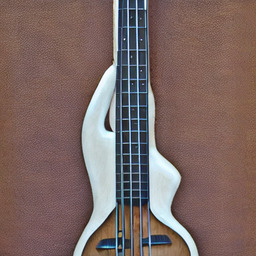}
        \includegraphics[width=0.15\textwidth]{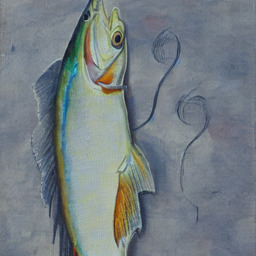}
        \includegraphics[width=0.15\textwidth]{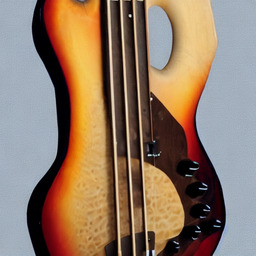}
        \includegraphics[width=0.15\textwidth]{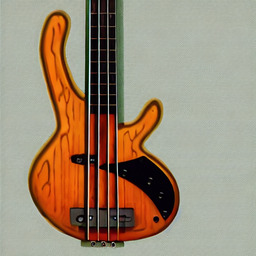}
        \includegraphics[width=0.15\textwidth]{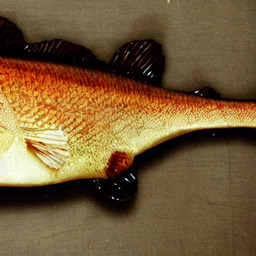}\\        \includegraphics[width=0.15\textwidth]{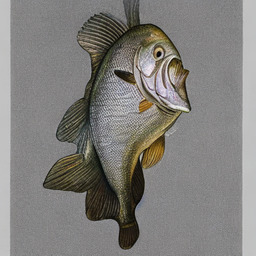}
        \includegraphics[width=0.15\textwidth]{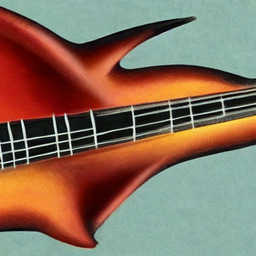}
        \includegraphics[width=0.15\textwidth]{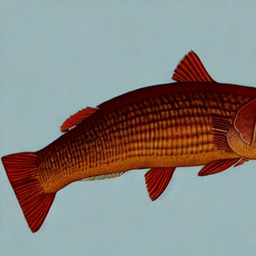}
        \includegraphics[width=0.15\textwidth]{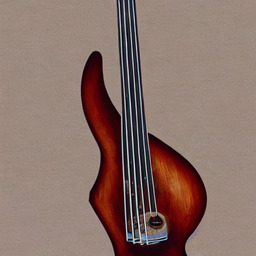}
        \includegraphics[width=0.15\textwidth]{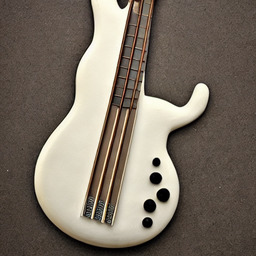}
        \caption{Unedited prompt encoding}
        \label{fig:bass_amb}
    \end{subfigure}
    \begin{subfigure}{0.45\textwidth}
        \centering
        \includegraphics[width=0.3\textwidth]{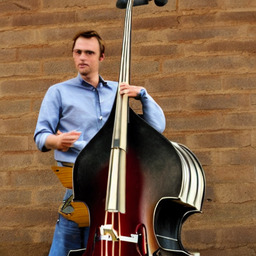}
        \includegraphics[width=0.3\textwidth]{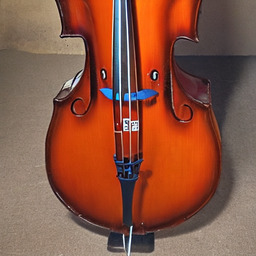}
        \includegraphics[width=0.3\textwidth]{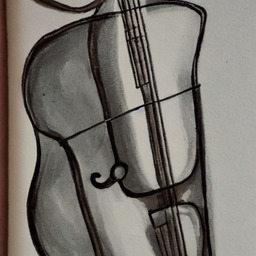}\\
        \includegraphics[width=0.3\textwidth]{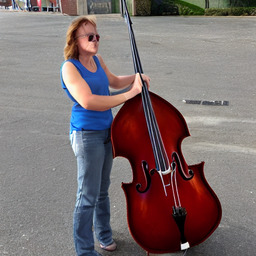}
        \includegraphics[width=0.3\textwidth]{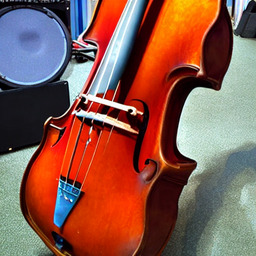}
        \includegraphics[width=0.3\textwidth]{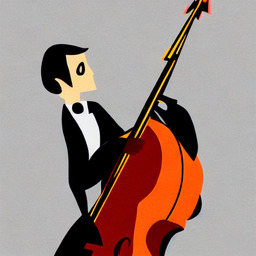}\\
        \includegraphics[width=0.3\textwidth]{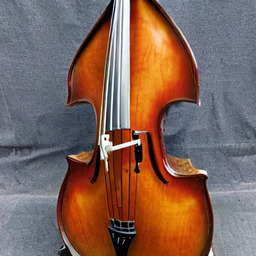}
        \includegraphics[width=0.3\textwidth]{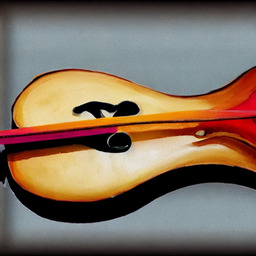}
        \includegraphics[width=0.3\textwidth]{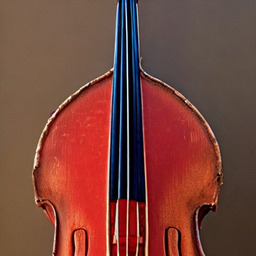}\\
        \includegraphics[width=0.3\textwidth]{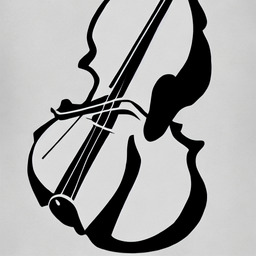}    \includegraphics[width=0.3\textwidth]{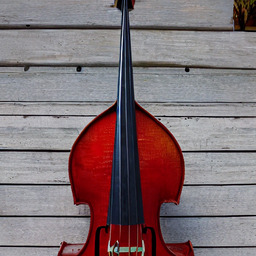}
        \includegraphics[width=0.3\textwidth]{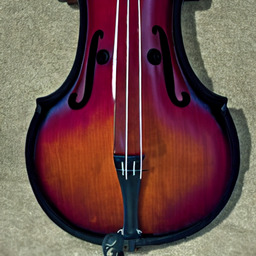}\\
        \includegraphics[width=0.3\textwidth]{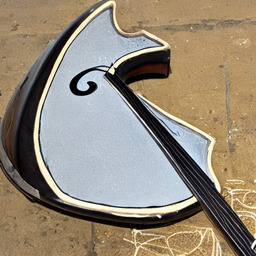}
        \includegraphics[width=0.3\textwidth]{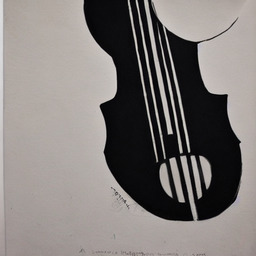}
        \includegraphics[width=0.3\textwidth]{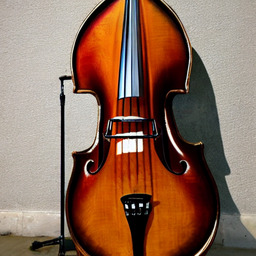}
        \caption{Encoding edited to favour music-related sense}
        \label{fig:bass_1}
    \end{subfigure}
    \begin{subfigure}{0.45\textwidth}
        \centering
        \includegraphics[width=0.3\textwidth]{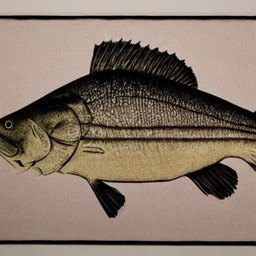}
        \includegraphics[width=0.3\textwidth]{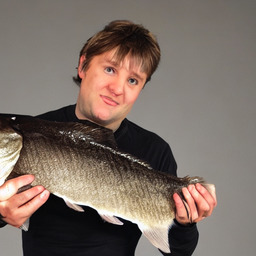}
        \includegraphics[width=0.3\textwidth]{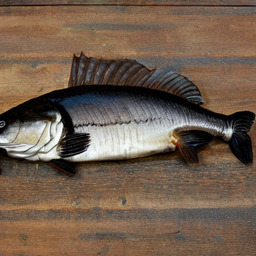}\\
        \includegraphics[width=0.3\textwidth]{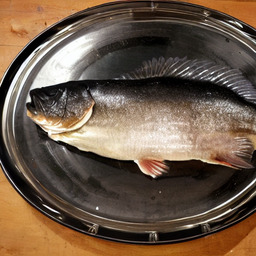}
        \includegraphics[width=0.3\textwidth]{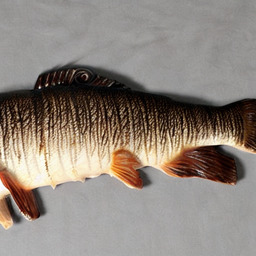}
        \includegraphics[width=0.3\textwidth]{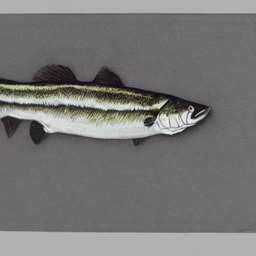}\\
        \includegraphics[width=0.3\textwidth]{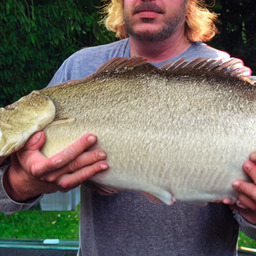}
        \includegraphics[width=0.3\textwidth]{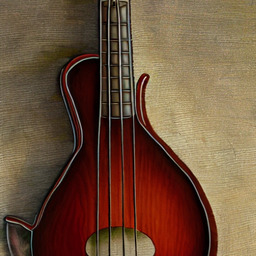}
        \includegraphics[width=0.3\textwidth]{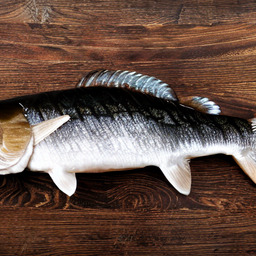}\\
        \includegraphics[width=0.3\textwidth]{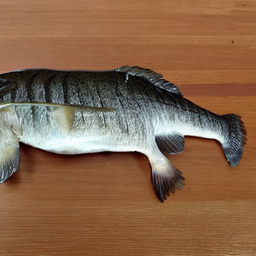}    \includegraphics[width=0.3\textwidth]{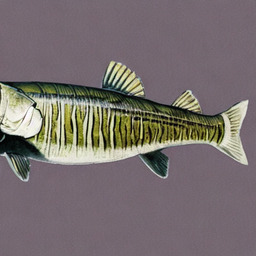}
        \includegraphics[width=0.3\textwidth]{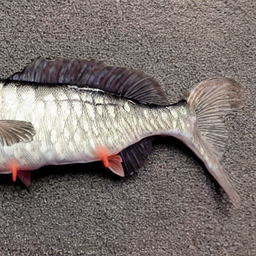}\\
        \includegraphics[width=0.3\textwidth]{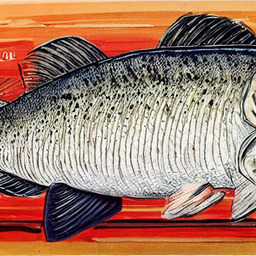}
        \includegraphics[width=0.3\textwidth]{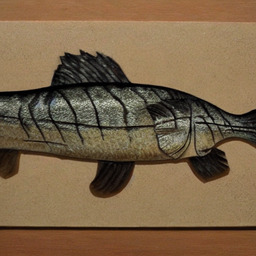}
        \includegraphics[width=0.3\textwidth]{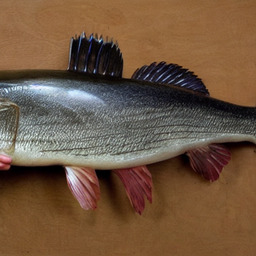}
        \caption{Encoding edited to favour fish sense}
        \label{fig:bass_2}
    \end{subfigure}
    \caption{Prompt: \prompt{a bass}}
    \label{fig:abass}
\end{figure}

\begin{figure}
    \centering
    \begin{subfigure}{\textwidth}
        \centering
        \includegraphics[width=0.15\textwidth]{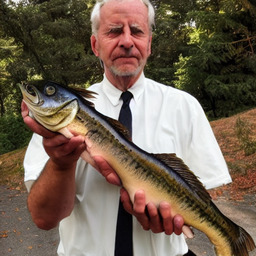}
        \includegraphics[width=0.15\textwidth]{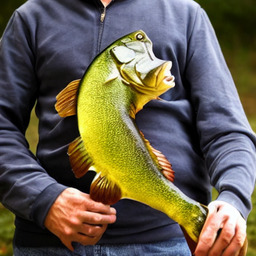}
        \includegraphics[width=0.15\textwidth]{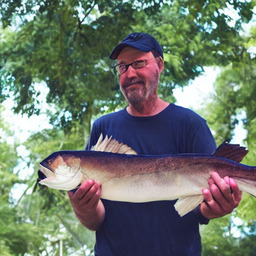}
        \includegraphics[width=0.15\textwidth]{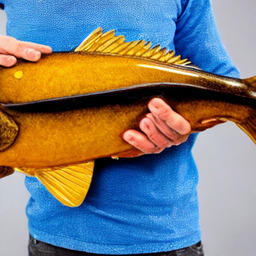}
        \includegraphics[width=0.15\textwidth]{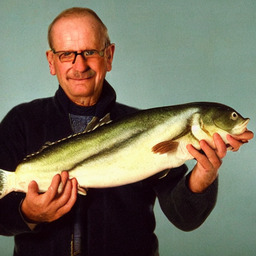}\\
        \includegraphics[width=0.15\textwidth]{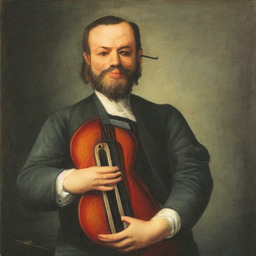}
        \includegraphics[width=0.15\textwidth]{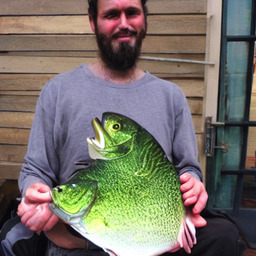}
        \includegraphics[width=0.15\textwidth]{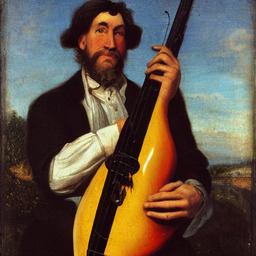}
        \includegraphics[width=0.15\textwidth]{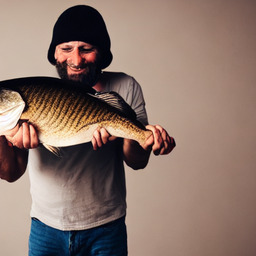}
        \includegraphics[width=0.15\textwidth]{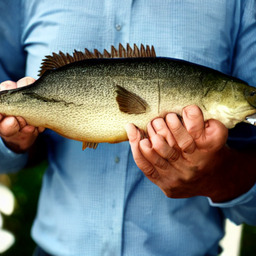}\\        \includegraphics[width=0.15\textwidth]{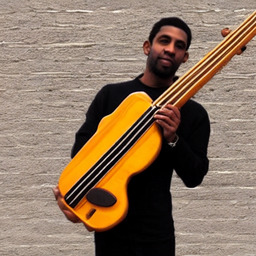}
        \includegraphics[width=0.15\textwidth]{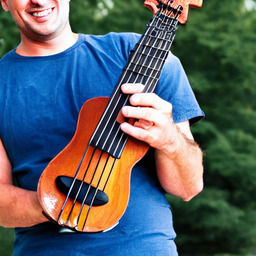}
        \includegraphics[width=0.15\textwidth]{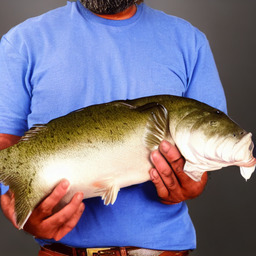}
        \includegraphics[width=0.15\textwidth]{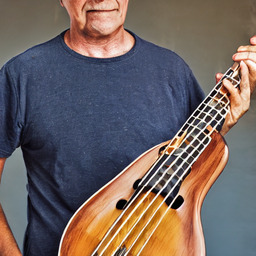}
        \includegraphics[width=0.15\textwidth]{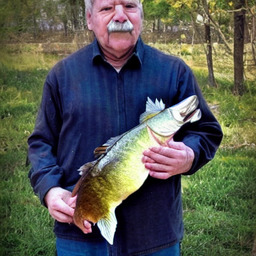}
        \caption{Unedited prompt encoding}
        \label{fig:bass_man_amb}
    \end{subfigure}
    \begin{subfigure}{0.45\textwidth}
        \centering
        \includegraphics[width=0.3\textwidth]{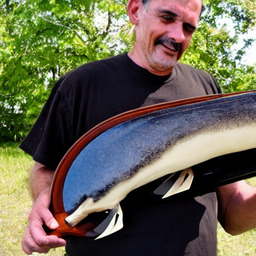}
        \includegraphics[width=0.3\textwidth]{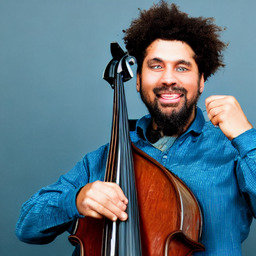}
        \includegraphics[width=0.3\textwidth]{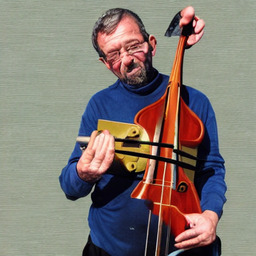}\\
        \includegraphics[width=0.3\textwidth]{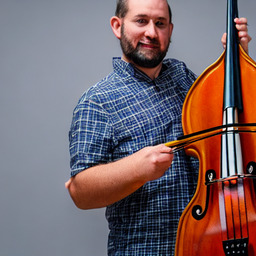}
        \includegraphics[width=0.3\textwidth]{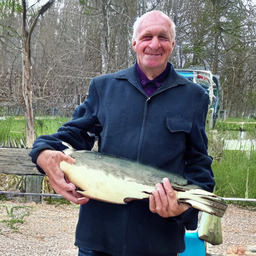}
        \includegraphics[width=0.3\textwidth]{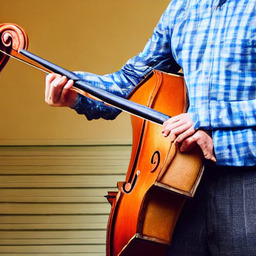}\\
        \includegraphics[width=0.3\textwidth]{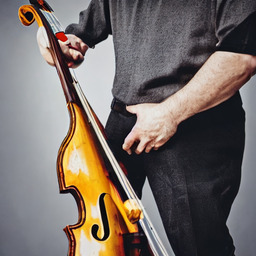}
        \includegraphics[width=0.3\textwidth]{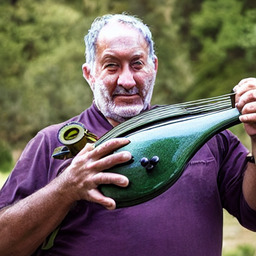}
        \includegraphics[width=0.3\textwidth]{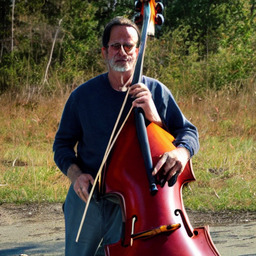}\\
        \includegraphics[width=0.3\textwidth]{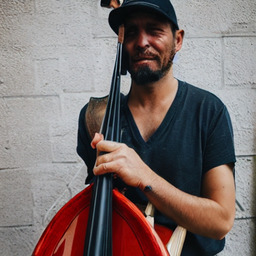}    \includegraphics[width=0.3\textwidth]{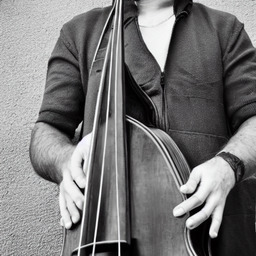}
        \includegraphics[width=0.3\textwidth]{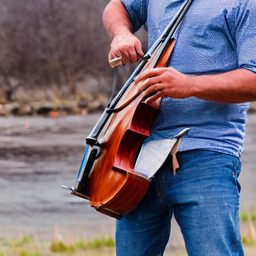}\\
        \includegraphics[width=0.3\textwidth]{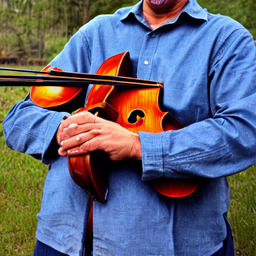}
        \includegraphics[width=0.3\textwidth]{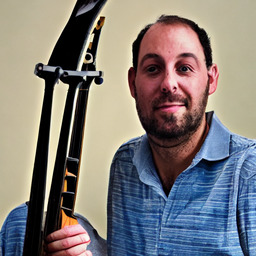}
        \includegraphics[width=0.3\textwidth]{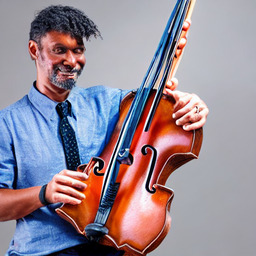}
        \caption{Encoding edited to favour music-related sense}
        \label{fig:bass_man_1}
    \end{subfigure}
    \begin{subfigure}{0.45\textwidth}
        \centering
        \includegraphics[width=0.3\textwidth]{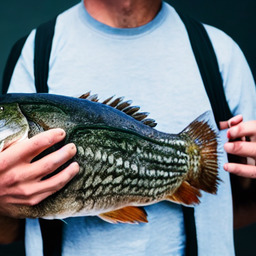}
        \includegraphics[width=0.3\textwidth]{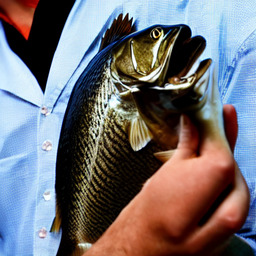}
        \includegraphics[width=0.3\textwidth]{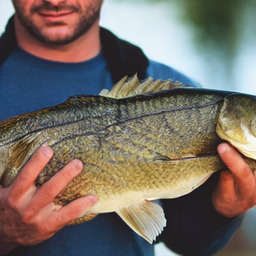}\\
        \includegraphics[width=0.3\textwidth]{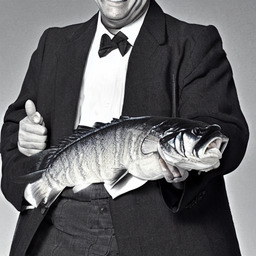}
        \includegraphics[width=0.3\textwidth]{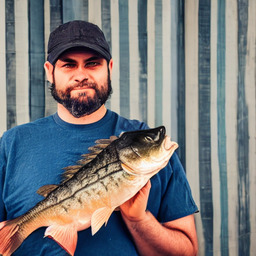}
        \includegraphics[width=0.3\textwidth]{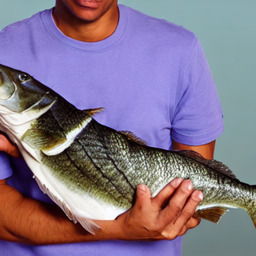}\\
        \includegraphics[width=0.3\textwidth]{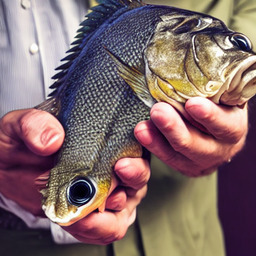}
        \includegraphics[width=0.3\textwidth]{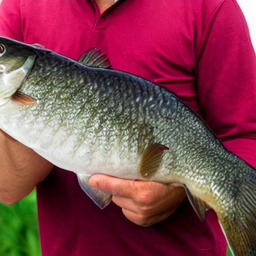}
        \includegraphics[width=0.3\textwidth]{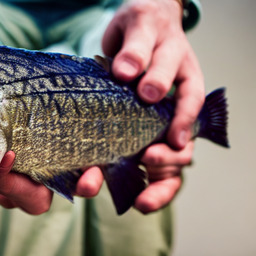}\\
        \includegraphics[width=0.3\textwidth]{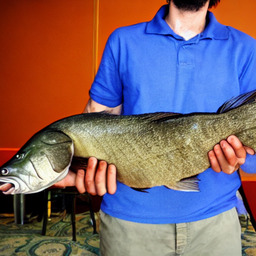}    \includegraphics[width=0.3\textwidth]{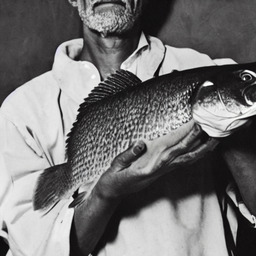}
        \includegraphics[width=0.3\textwidth]{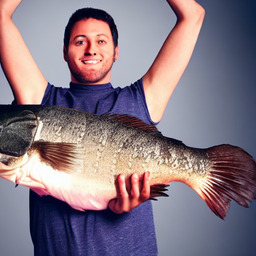}\\
        \includegraphics[width=0.3\textwidth]{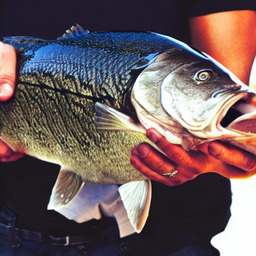}
        \includegraphics[width=0.3\textwidth]{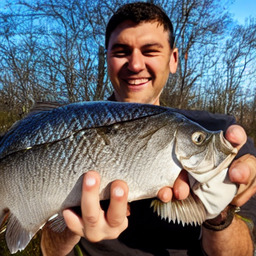}
        \includegraphics[width=0.3\textwidth]{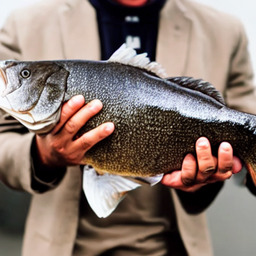}
        \caption{Encoding edited to favour fish sense}
        \label{fig:bass_man_2}
    \end{subfigure}
    \caption{Prompt: \prompt{a man holding a bass}}
    \label{fig:bass_man}
\end{figure}

\begin{figure}
    \centering
    \begin{subfigure}{\textwidth}
        \centering
        \includegraphics[width=0.15\textwidth]{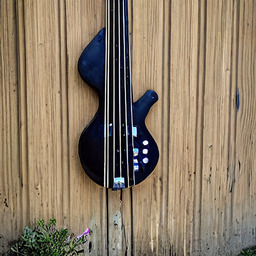}
        \includegraphics[width=0.15\textwidth]{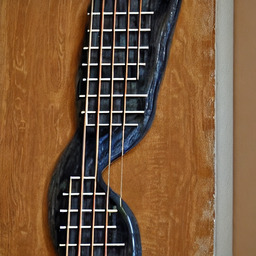}
        \includegraphics[width=0.15\textwidth]{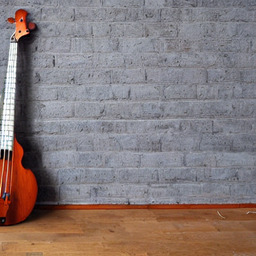}
        \includegraphics[width=0.15\textwidth]{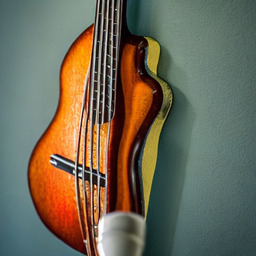}
        \includegraphics[width=0.15\textwidth]{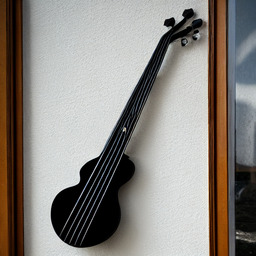}\\
        \includegraphics[width=0.15\textwidth]{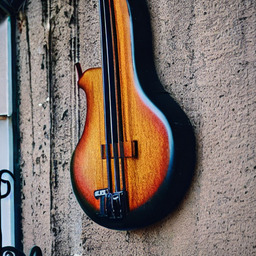}
        \includegraphics[width=0.15\textwidth]{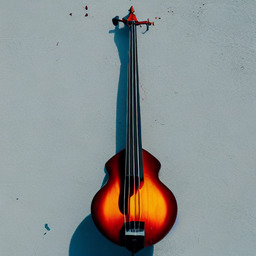}
        \includegraphics[width=0.15\textwidth]{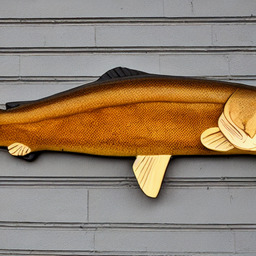}
        \includegraphics[width=0.15\textwidth]{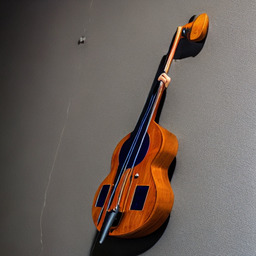}
        \includegraphics[width=0.15\textwidth]{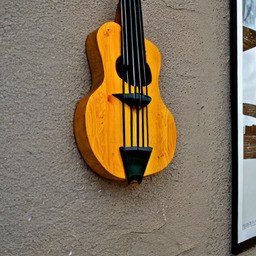}\\        \includegraphics[width=0.15\textwidth]{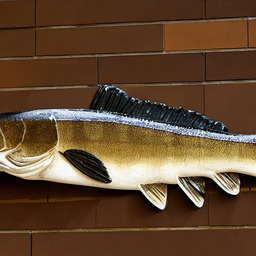}
        \includegraphics[width=0.15\textwidth]{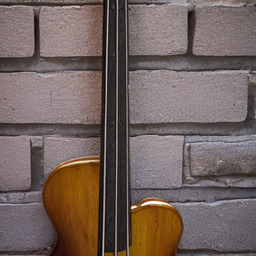}
        \includegraphics[width=0.15\textwidth]{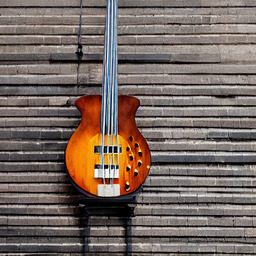}
        \includegraphics[width=0.15\textwidth]{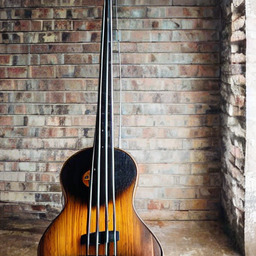}
        \includegraphics[width=0.15\textwidth]{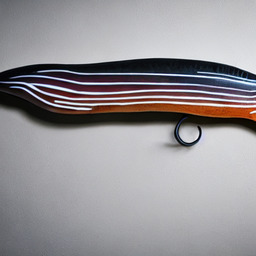}
        \caption{Unedited prompt encoding}
        \label{fig:bass_wall_amb}
    \end{subfigure}
    \begin{subfigure}{0.45\textwidth}
        \centering
        \includegraphics[width=0.3\textwidth]{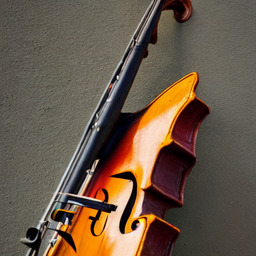}
        \includegraphics[width=0.3\textwidth]{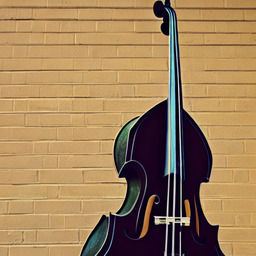}
        \includegraphics[width=0.3\textwidth]{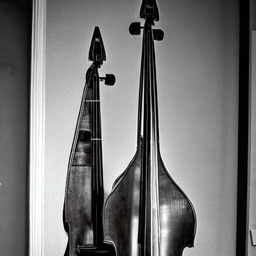}\\
        \includegraphics[width=0.3\textwidth]{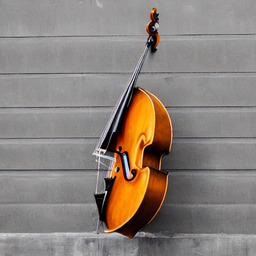}
        \includegraphics[width=0.3\textwidth]{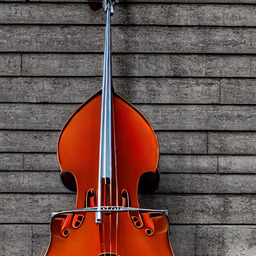}
        \includegraphics[width=0.3\textwidth]{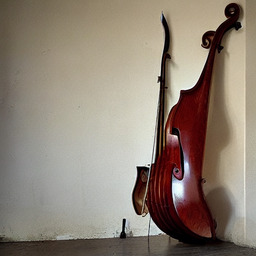}\\
        \includegraphics[width=0.3\textwidth]{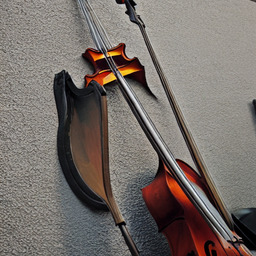}
        \includegraphics[width=0.3\textwidth]{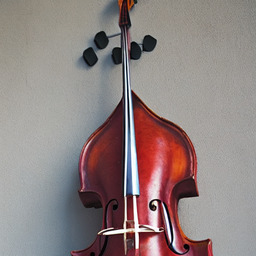}
        \includegraphics[width=0.3\textwidth]{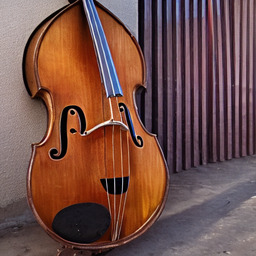}\\
        \includegraphics[width=0.3\textwidth]{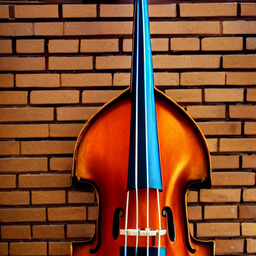}    \includegraphics[width=0.3\textwidth]{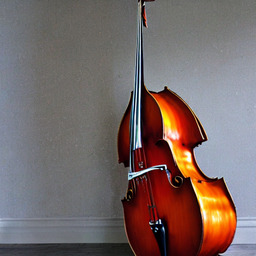}
        \includegraphics[width=0.3\textwidth]{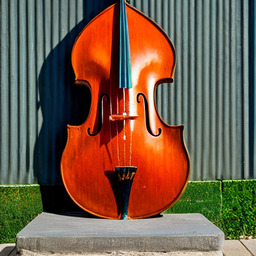}\\
        \includegraphics[width=0.3\textwidth]{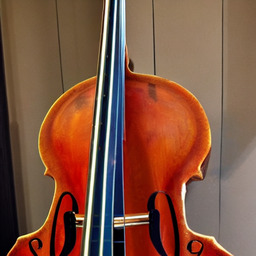}
        \includegraphics[width=0.3\textwidth]{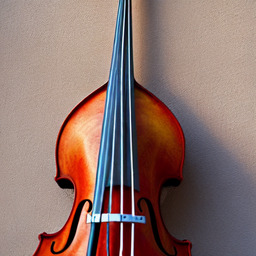}
        \includegraphics[width=0.3\textwidth]{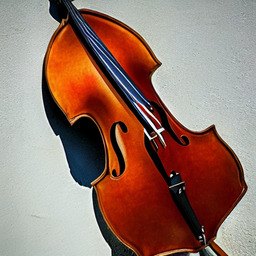}
        \caption{Encoding edited to favour music-related sense}
        \label{fig:bass_wall_1}
    \end{subfigure}
    \begin{subfigure}{0.45\textwidth}
        \centering
        \includegraphics[width=0.3\textwidth]{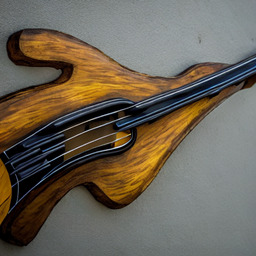}
        \includegraphics[width=0.3\textwidth]{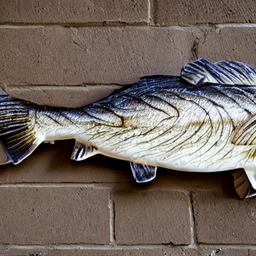}
        \includegraphics[width=0.3\textwidth]{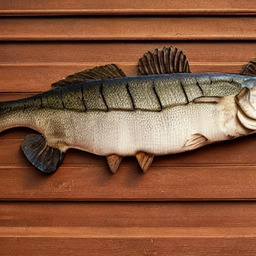}\\
        \includegraphics[width=0.3\textwidth]{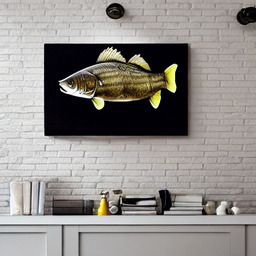}
        \includegraphics[width=0.3\textwidth]{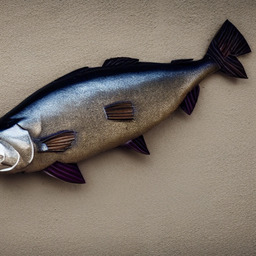}
        \includegraphics[width=0.3\textwidth]{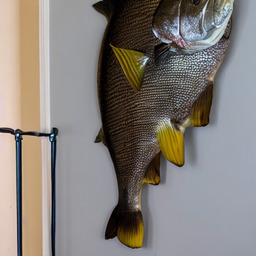}\\
        \includegraphics[width=0.3\textwidth]{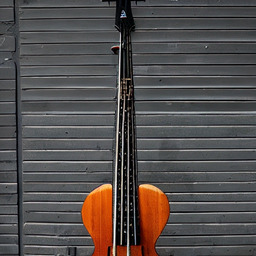}
        \includegraphics[width=0.3\textwidth]{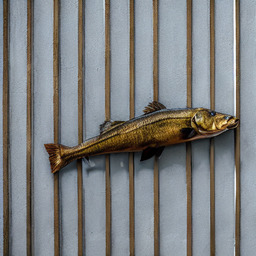}
        \includegraphics[width=0.3\textwidth]{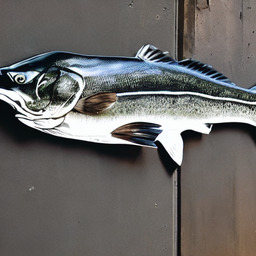}\\
        \includegraphics[width=0.3\textwidth]{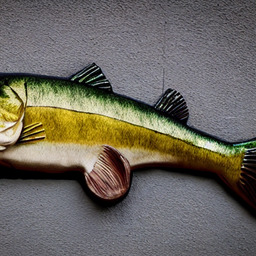}    \includegraphics[width=0.3\textwidth]{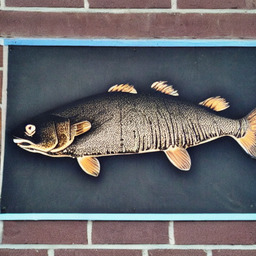}
        \includegraphics[width=0.3\textwidth]{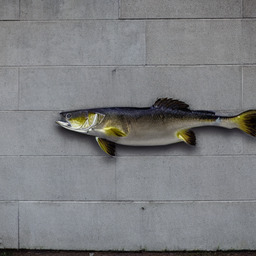}\\
        \includegraphics[width=0.3\textwidth]{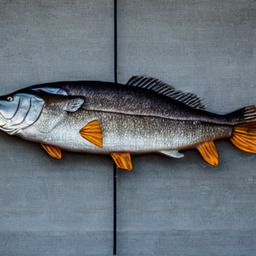}
        \includegraphics[width=0.3\textwidth]{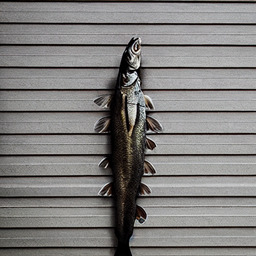}
        \includegraphics[width=0.3\textwidth]{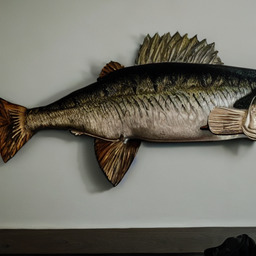}
        \caption{Encoding edited to favour fish sense}
        \label{fig:bass_wall_2}
    \end{subfigure}
    \caption{Prompt: \prompt{a bass displayed on a wall}}
    \label{fig:bass_wall}
\end{figure}

\begin{figure}
    \centering
    \begin{subfigure}{\textwidth}
        \centering
        \includegraphics[width=0.15\textwidth]{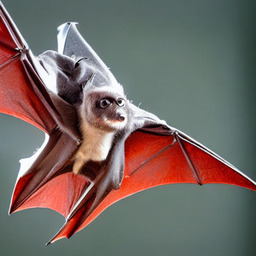}
        \includegraphics[width=0.15\textwidth]{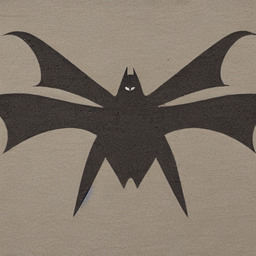}
        \includegraphics[width=0.15\textwidth]{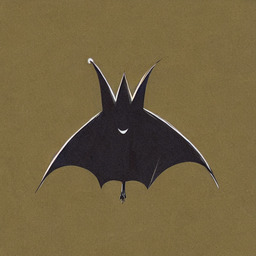}
        \includegraphics[width=0.15\textwidth]{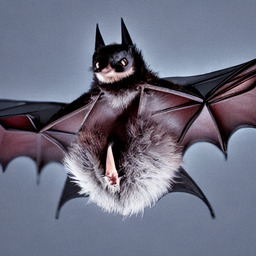}
        \includegraphics[width=0.15\textwidth]{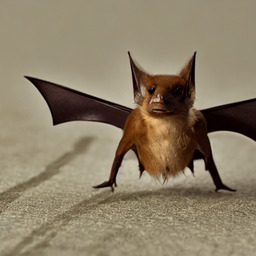}\\
        \includegraphics[width=0.15\textwidth]{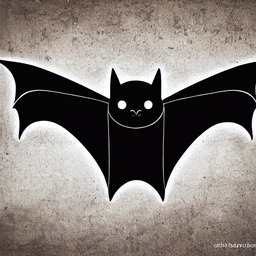}
        \includegraphics[width=0.15\textwidth]{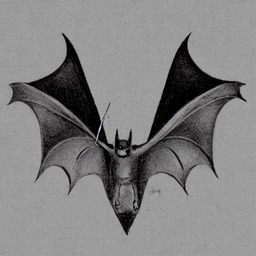}
        \includegraphics[width=0.15\textwidth]{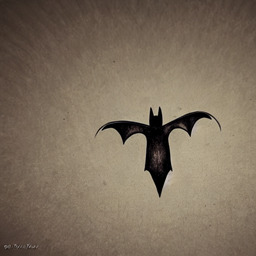}
        \includegraphics[width=0.15\textwidth]{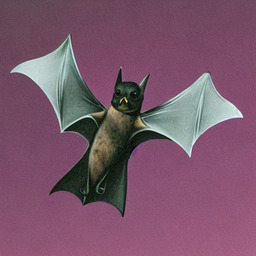}
        \includegraphics[width=0.15\textwidth]{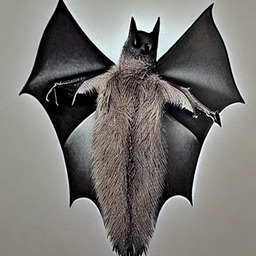}\\        \includegraphics[width=0.15\textwidth]{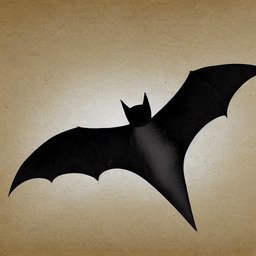}
        \includegraphics[width=0.15\textwidth]{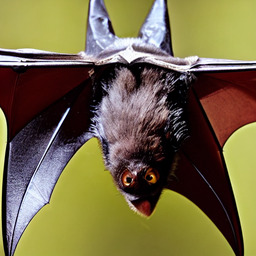}
        \includegraphics[width=0.15\textwidth]{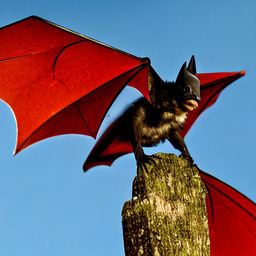}
        \includegraphics[width=0.15\textwidth]{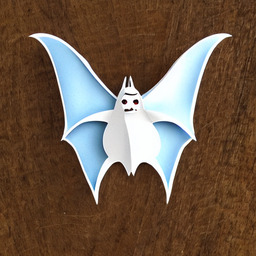}
        \includegraphics[width=0.15\textwidth]{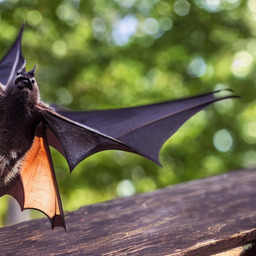}
        \caption{Unedited prompt encoding}
        \label{fig:bat_amb}
    \end{subfigure}
    \begin{subfigure}{0.45\textwidth}
        \centering
        \includegraphics[width=0.3\textwidth]{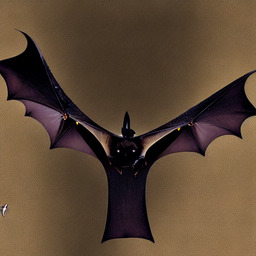}
        \includegraphics[width=0.3\textwidth]{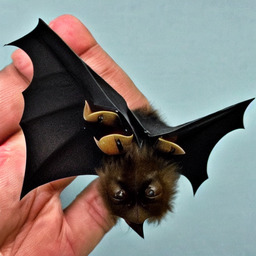}
        \includegraphics[width=0.3\textwidth]{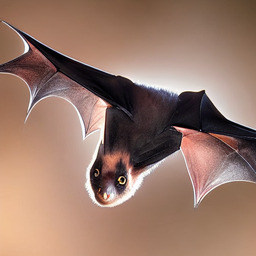}\\
        \includegraphics[width=0.3\textwidth]{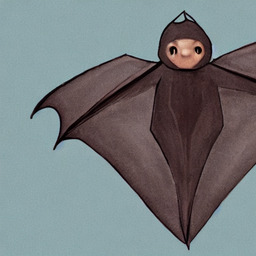}
        \includegraphics[width=0.3\textwidth]{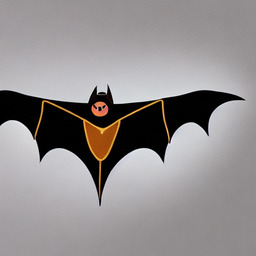}
        \includegraphics[width=0.3\textwidth]{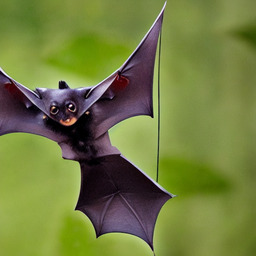}\\
        \includegraphics[width=0.3\textwidth]{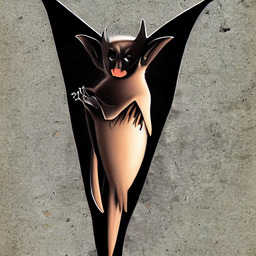}
        \includegraphics[width=0.3\textwidth]{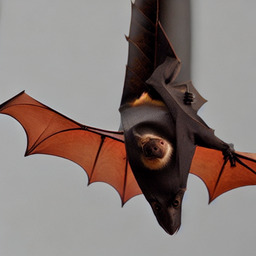}
        \includegraphics[width=0.3\textwidth]{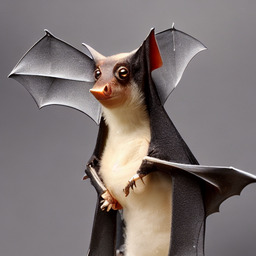}\\
        \includegraphics[width=0.3\textwidth]{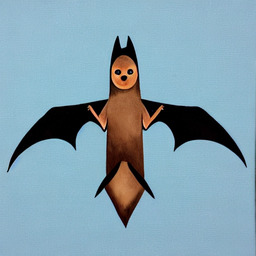}    \includegraphics[width=0.3\textwidth]{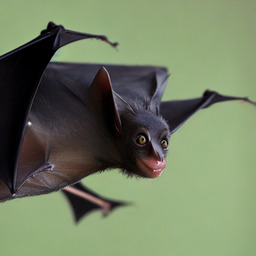}
        \includegraphics[width=0.3\textwidth]{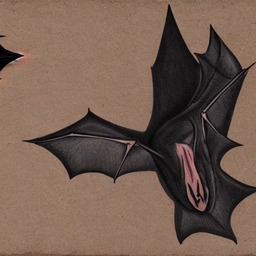}\\
        \includegraphics[width=0.3\textwidth]{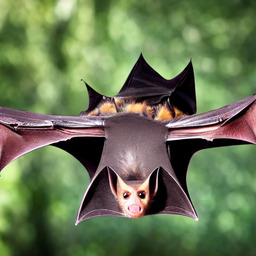}
        \includegraphics[width=0.3\textwidth]{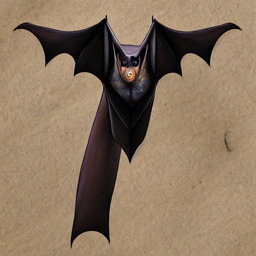}
        \includegraphics[width=0.3\textwidth]{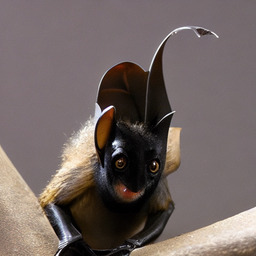}
        \caption{Encoding edited to favour animal sense}
        \label{fig:bat_2}
    \end{subfigure}
    \begin{subfigure}{0.45\textwidth}
        \centering
        \includegraphics[width=0.3\textwidth]{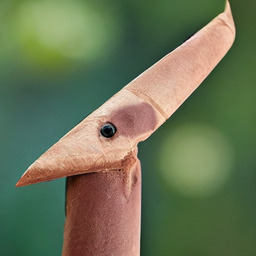}
        \includegraphics[width=0.3\textwidth]{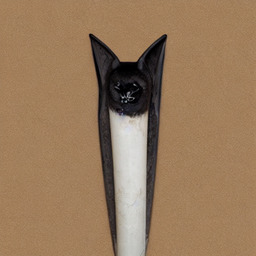}
        \includegraphics[width=0.3\textwidth]{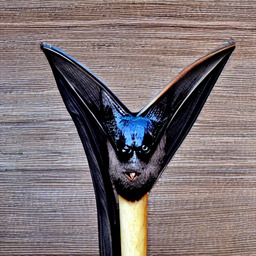}\\
        \includegraphics[width=0.3\textwidth]{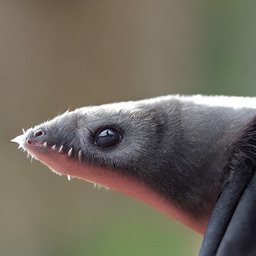}
        \includegraphics[width=0.3\textwidth]{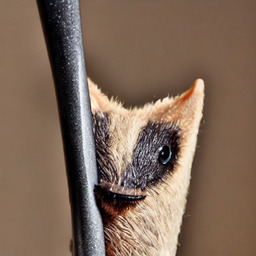}
        \includegraphics[width=0.3\textwidth]{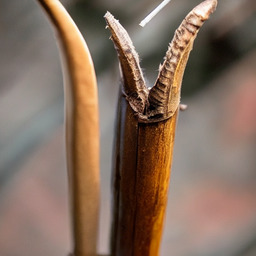}\\
        \includegraphics[width=0.3\textwidth]{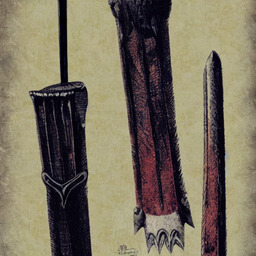}
        \includegraphics[width=0.3\textwidth]{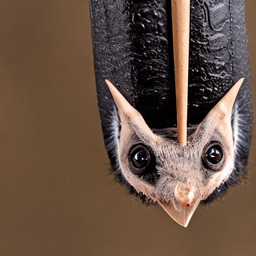}
        \includegraphics[width=0.3\textwidth]{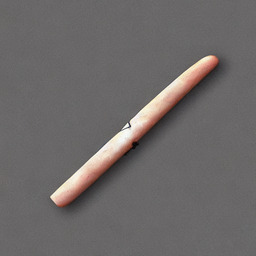}\\
        \includegraphics[width=0.3\textwidth]{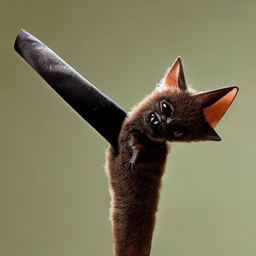}    \includegraphics[width=0.3\textwidth]{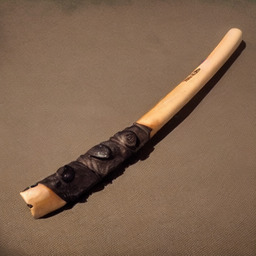}
        \includegraphics[width=0.3\textwidth]{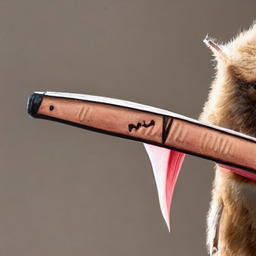}\\
        \includegraphics[width=0.3\textwidth]{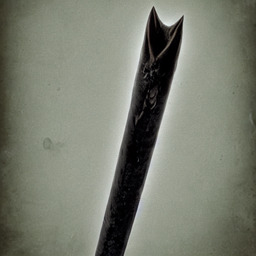}
        \includegraphics[width=0.3\textwidth]{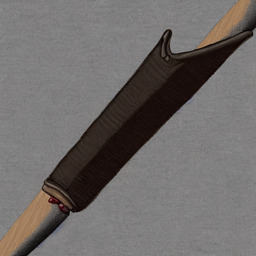}
        \includegraphics[width=0.3\textwidth]{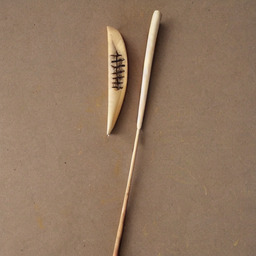}
        \caption{Encoding edited to favour sports-related sense}
        \label{fig:bat_1}
    \end{subfigure}
    \caption{Prompt: \prompt{a bat}}
    \label{fig:abat}
\end{figure}

\begin{figure}
    \centering
    \begin{subfigure}{\textwidth}
        \centering
        \includegraphics[width=0.15\textwidth]{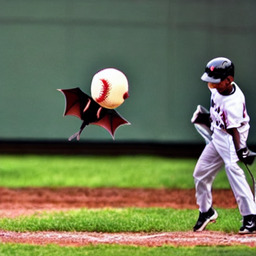}
        \includegraphics[width=0.15\textwidth]{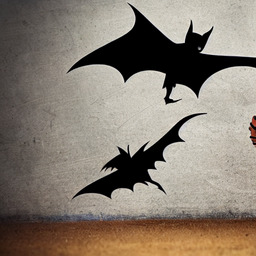}
        \includegraphics[width=0.15\textwidth]{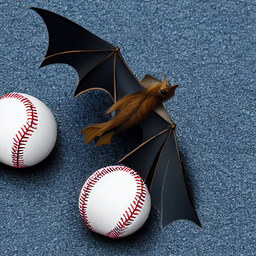}
        \includegraphics[width=0.15\textwidth]{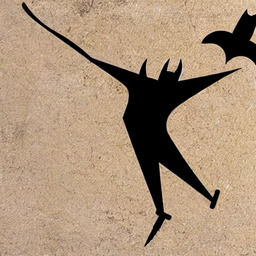}
        \includegraphics[width=0.15\textwidth]{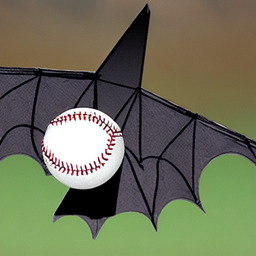}\\
        \includegraphics[width=0.15\textwidth]{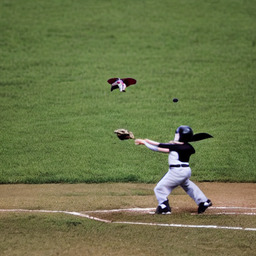}
        \includegraphics[width=0.15\textwidth]{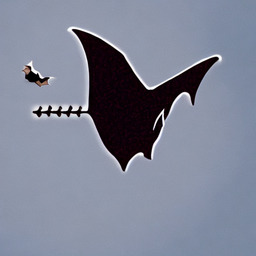}
        \includegraphics[width=0.15\textwidth]{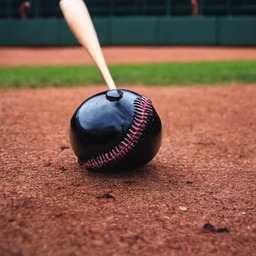}
        \includegraphics[width=0.15\textwidth]{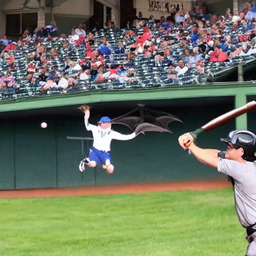}
        \includegraphics[width=0.15\textwidth]{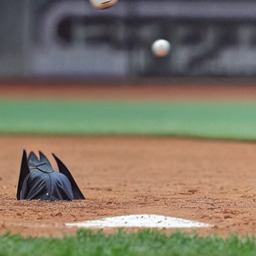}\\        \includegraphics[width=0.15\textwidth]{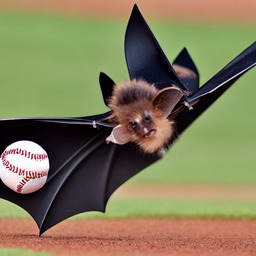}
        \includegraphics[width=0.15\textwidth]{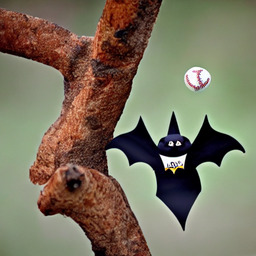}
        \includegraphics[width=0.15\textwidth]{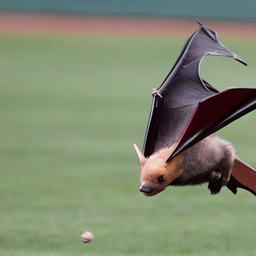}
        \includegraphics[width=0.15\textwidth]{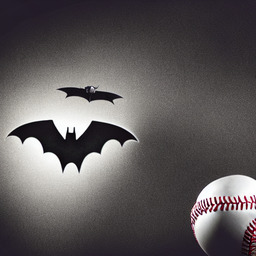}
        \includegraphics[width=0.15\textwidth]{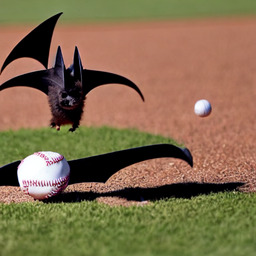}
        \caption{Unedited prompt encoding}
        \label{fig:bat_fly_amb}
    \end{subfigure}
    \begin{subfigure}{0.45\textwidth}
        \centering
        \includegraphics[width=0.3\textwidth]{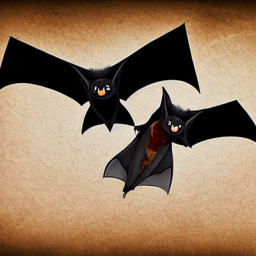}
        \includegraphics[width=0.3\textwidth]{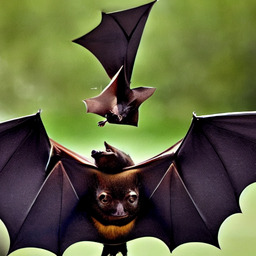}
        \includegraphics[width=0.3\textwidth]{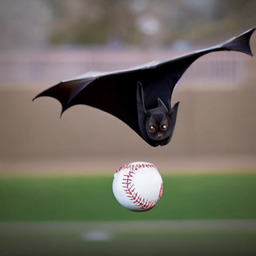}\\
        \includegraphics[width=0.3\textwidth]{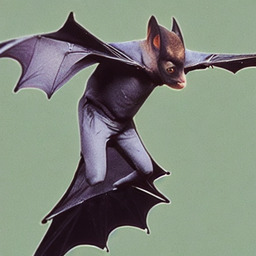}
        \includegraphics[width=0.3\textwidth]{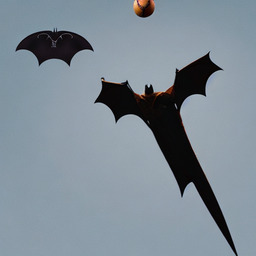}
        \includegraphics[width=0.3\textwidth]{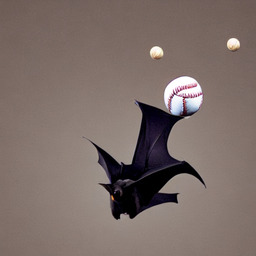}\\
        \includegraphics[width=0.3\textwidth]{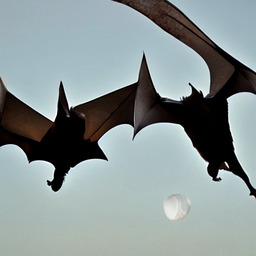}
        \includegraphics[width=0.3\textwidth]{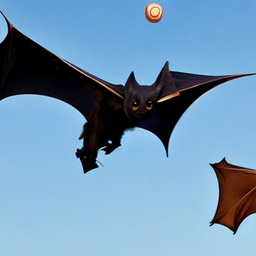}
        \includegraphics[width=0.3\textwidth]{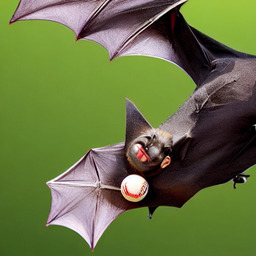}\\
        \includegraphics[width=0.3\textwidth]{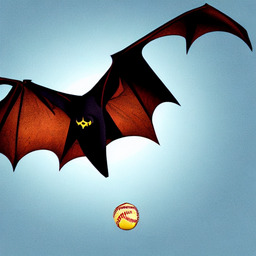}    \includegraphics[width=0.3\textwidth]{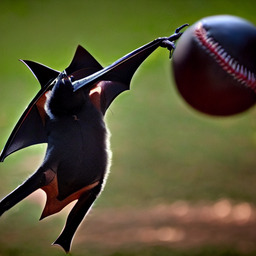}
        \includegraphics[width=0.3\textwidth]{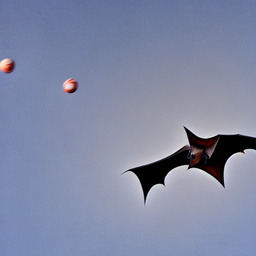}\\
        \includegraphics[width=0.3\textwidth]{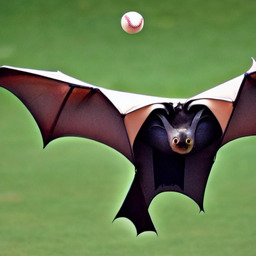}
        \includegraphics[width=0.3\textwidth]{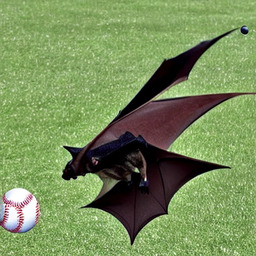}
        \includegraphics[width=0.3\textwidth]{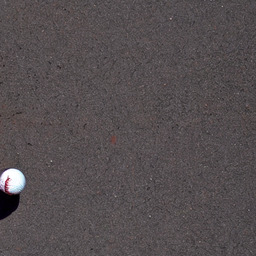}
        \caption{Encoding edited to favour animal sense}
        \label{fig:bat_fly_2}
    \end{subfigure}
    \begin{subfigure}{0.45\textwidth}
        \centering
        \includegraphics[width=0.3\textwidth]{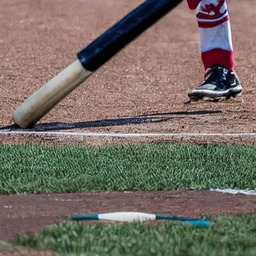}
        \includegraphics[width=0.3\textwidth]{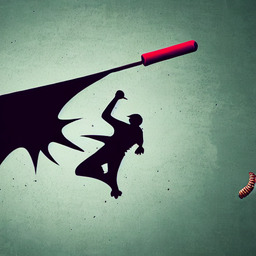}
        \includegraphics[width=0.3\textwidth]{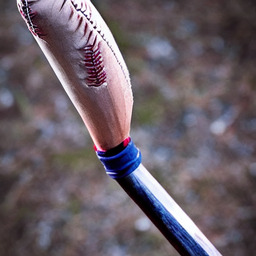}\\
        \includegraphics[width=0.3\textwidth]{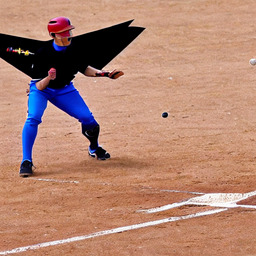}
        \includegraphics[width=0.3\textwidth]{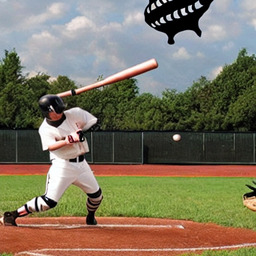}
        \includegraphics[width=0.3\textwidth]{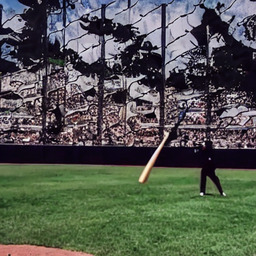}\\
        \includegraphics[width=0.3\textwidth]{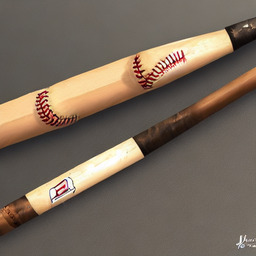}
        \includegraphics[width=0.3\textwidth]{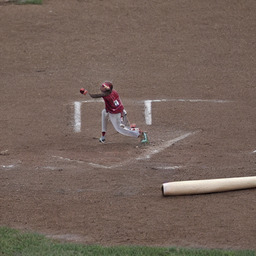}
        \includegraphics[width=0.3\textwidth]{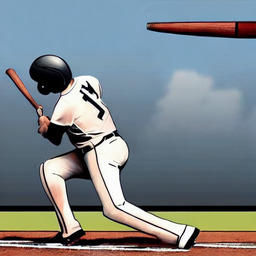}\\
        \includegraphics[width=0.3\textwidth]{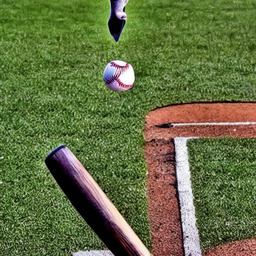}    \includegraphics[width=0.3\textwidth]{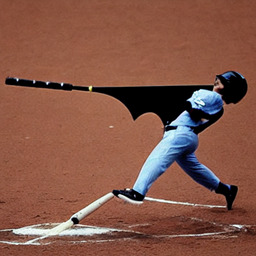}
        \includegraphics[width=0.3\textwidth]{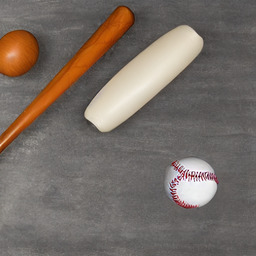}\\
        \includegraphics[width=0.3\textwidth]{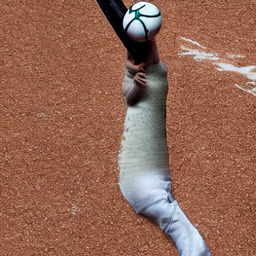}
        \includegraphics[width=0.3\textwidth]{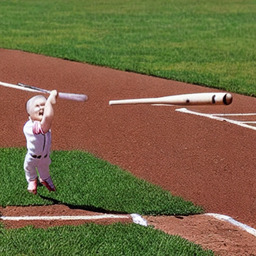}
        \includegraphics[width=0.3\textwidth]{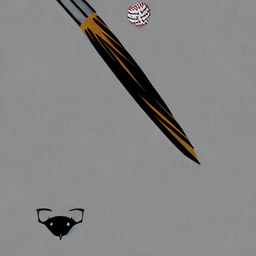}
        \caption{Encoding edited to favour sports-related sense}
        \label{fig:bat_fly_1}
    \end{subfigure}
    \caption{Prompt: \prompt{a bat and a baseball fly through the air}}
    \label{fig:bat_fly}
\end{figure}

\begin{figure}
    \centering
    \begin{subfigure}{\textwidth}
        \centering
        \includegraphics[width=0.15\textwidth]{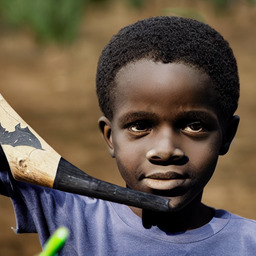}
        \includegraphics[width=0.15\textwidth]{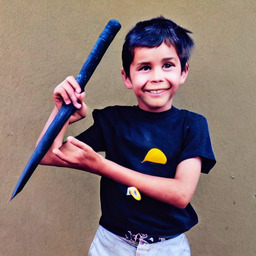}
        \includegraphics[width=0.15\textwidth]{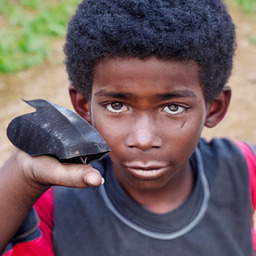}
        \includegraphics[width=0.15\textwidth]{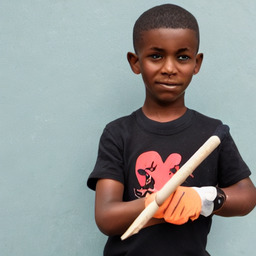}
        \includegraphics[width=0.15\textwidth]{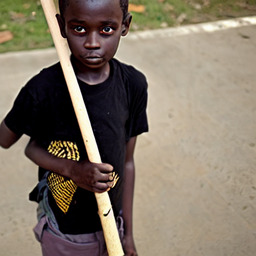}\\
        \includegraphics[width=0.15\textwidth]{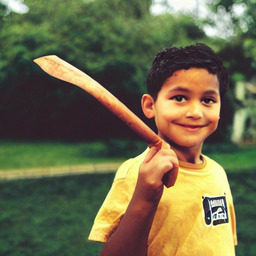}
        \includegraphics[width=0.15\textwidth]{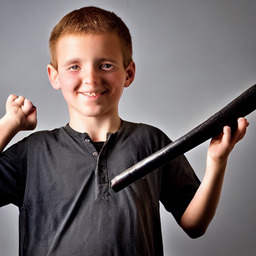}
        \includegraphics[width=0.15\textwidth]{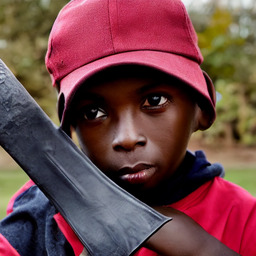}
        \includegraphics[width=0.15\textwidth]{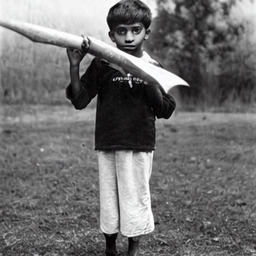}
        \includegraphics[width=0.15\textwidth]{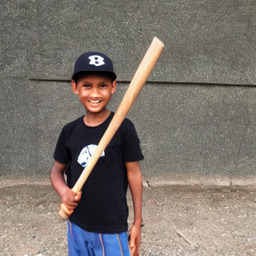}\\        \includegraphics[width=0.15\textwidth]{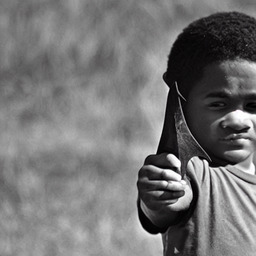}
        \includegraphics[width=0.15\textwidth]{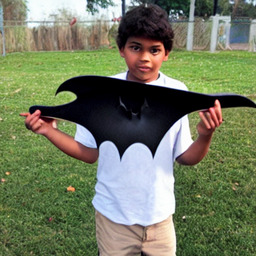}
        \includegraphics[width=0.15\textwidth]{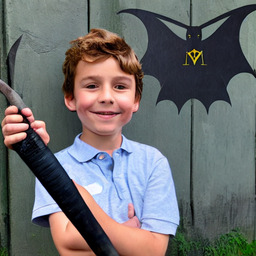}
        \includegraphics[width=0.15\textwidth]{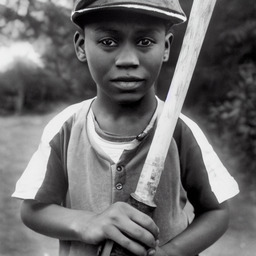}
        \includegraphics[width=0.15\textwidth]{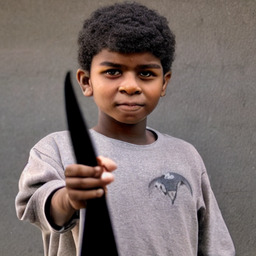}
        \caption{Unedited prompt encoding}
        \label{fig:boy_bat_amb}
    \end{subfigure}
    \begin{subfigure}{0.45\textwidth}
        \centering
        \includegraphics[width=0.3\textwidth]{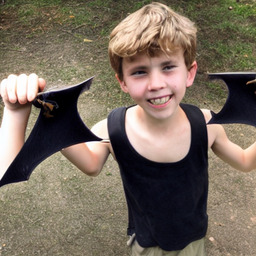}
        \includegraphics[width=0.3\textwidth]{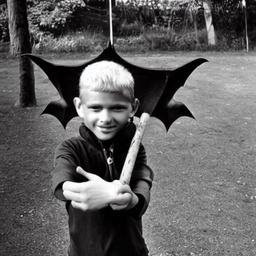}
        \includegraphics[width=0.3\textwidth]{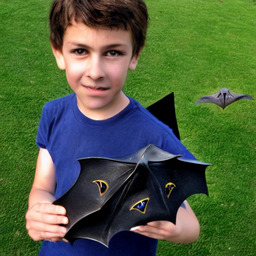}\\
        \includegraphics[width=0.3\textwidth]{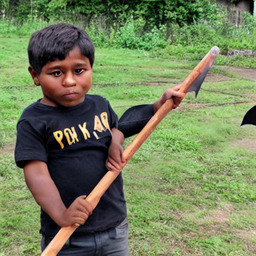}
        \includegraphics[width=0.3\textwidth]{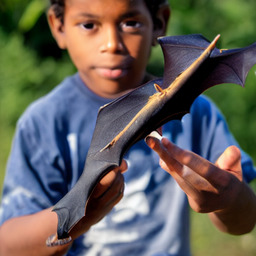}
        \includegraphics[width=0.3\textwidth]{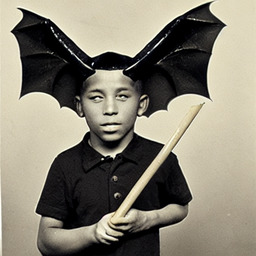}\\
        \includegraphics[width=0.3\textwidth]{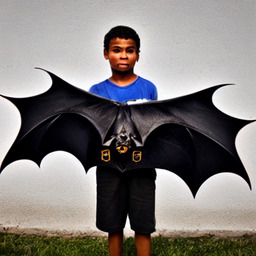}
        \includegraphics[width=0.3\textwidth]{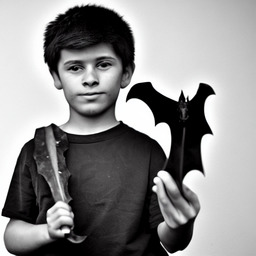}
        \includegraphics[width=0.3\textwidth]{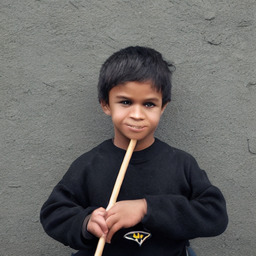}\\
        \includegraphics[width=0.3\textwidth]{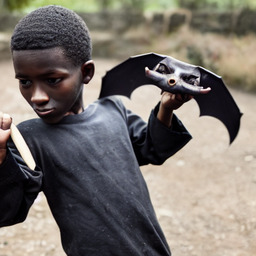}    \includegraphics[width=0.3\textwidth]{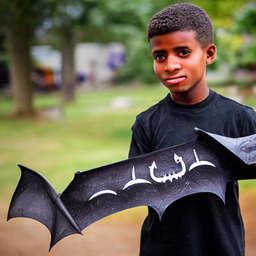}
        \includegraphics[width=0.3\textwidth]{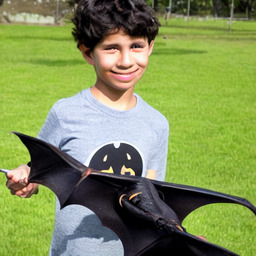}\\
        \includegraphics[width=0.3\textwidth]{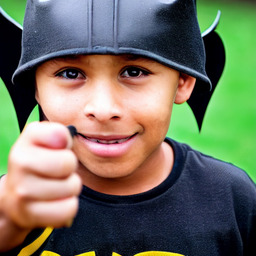}
        \includegraphics[width=0.3\textwidth]{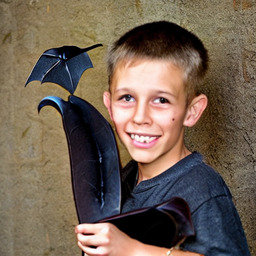}
        \includegraphics[width=0.3\textwidth]{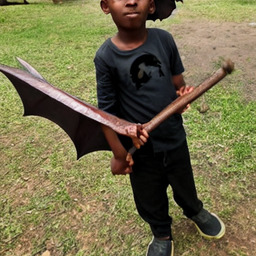}
        \caption{Encoding edited to favour animal sense}
        \label{fig:boy_bat_2}
    \end{subfigure}
    \begin{subfigure}{0.45\textwidth}
        \centering
        \includegraphics[width=0.3\textwidth]{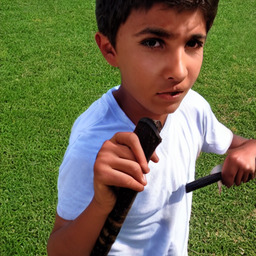}
        \includegraphics[width=0.3\textwidth]{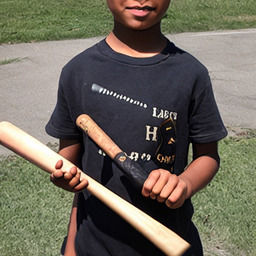}
        \includegraphics[width=0.3\textwidth]{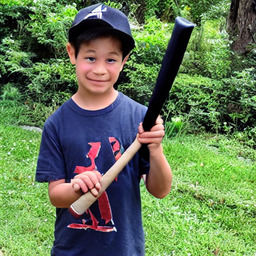}\\
        \includegraphics[width=0.3\textwidth]{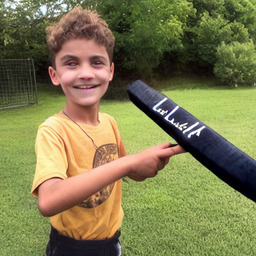}
        \includegraphics[width=0.3\textwidth]{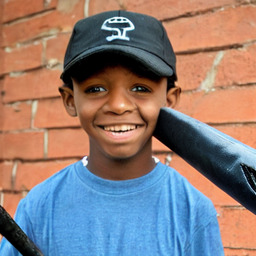}
        \includegraphics[width=0.3\textwidth]{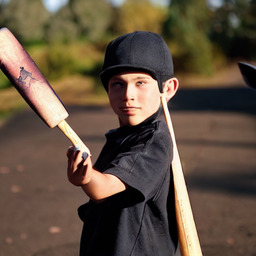}\\
        \includegraphics[width=0.3\textwidth]{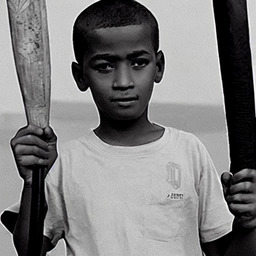}
        \includegraphics[width=0.3\textwidth]{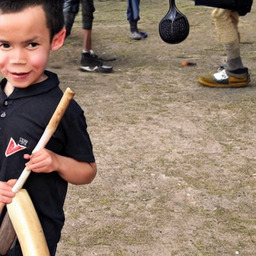}
        \includegraphics[width=0.3\textwidth]{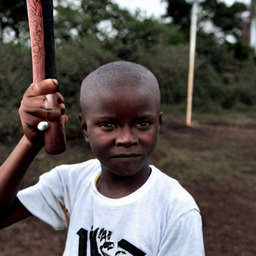}\\
        \includegraphics[width=0.3\textwidth]{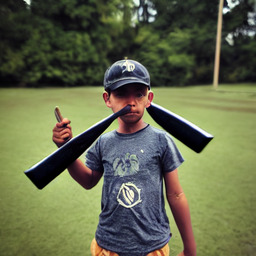}    \includegraphics[width=0.3\textwidth]{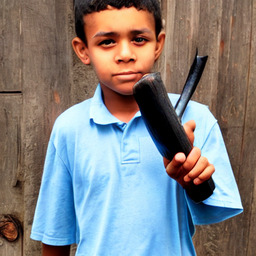}
        \includegraphics[width=0.3\textwidth]{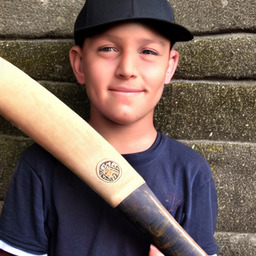}\\
        \includegraphics[width=0.3\textwidth]{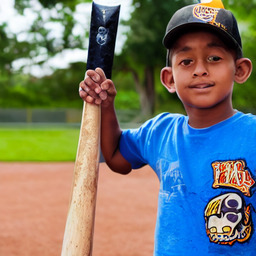}
        \includegraphics[width=0.3\textwidth]{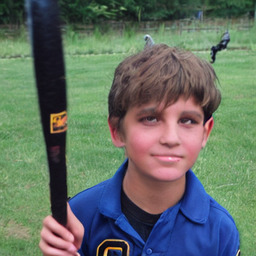}
        \includegraphics[width=0.3\textwidth]{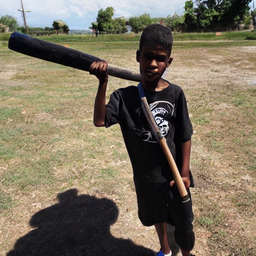}
        \caption{Encoding edited to favour sports-related sense}
        \label{fig:boy_bat_1}
    \end{subfigure}
    \caption{Prompt: \prompt{a boy holds a black bat}}
    \label{fig:batboy}
\end{figure}

\begin{figure}
    \centering
    \begin{subfigure}{\textwidth}
        \centering
        \includegraphics[width=0.15\textwidth]{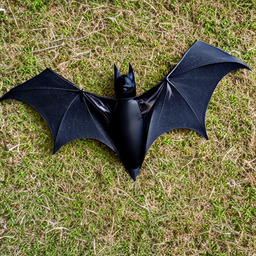}
        \includegraphics[width=0.15\textwidth]{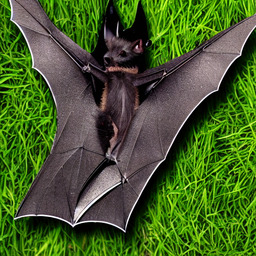}
        \includegraphics[width=0.15\textwidth]{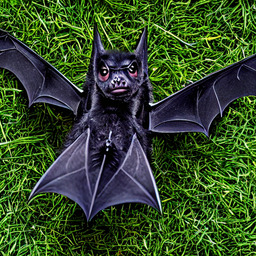}
        \includegraphics[width=0.15\textwidth]{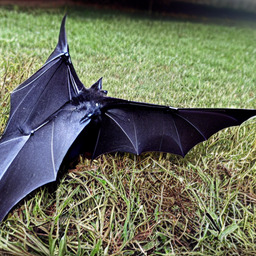}
        \includegraphics[width=0.15\textwidth]{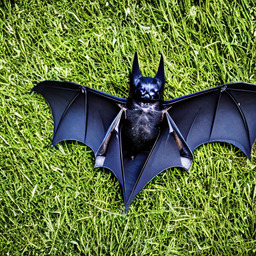}\\
        \includegraphics[width=0.15\textwidth]{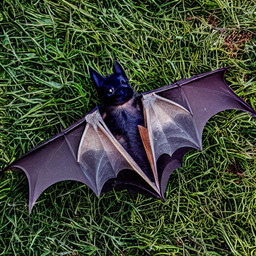}
        \includegraphics[width=0.15\textwidth]{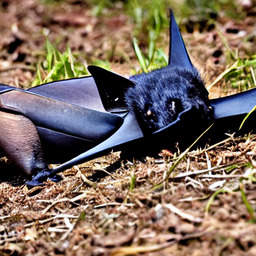}
        \includegraphics[width=0.15\textwidth]{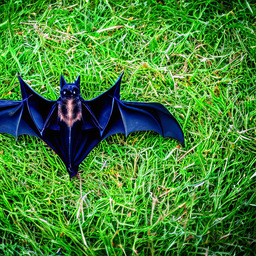}
        \includegraphics[width=0.15\textwidth]{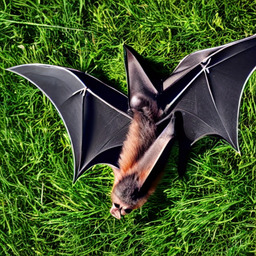}
        \includegraphics[width=0.15\textwidth]{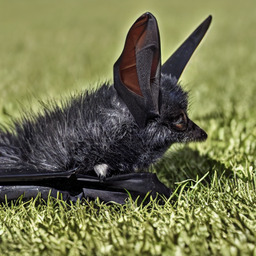}\\        \includegraphics[width=0.15\textwidth]{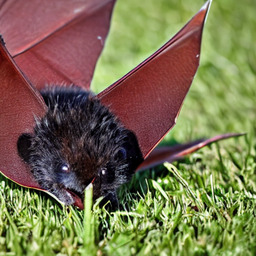}
        \includegraphics[width=0.15\textwidth]{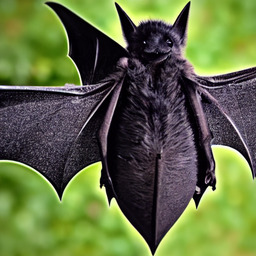}
        \includegraphics[width=0.15\textwidth]{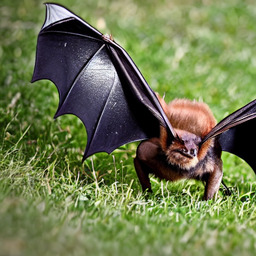}
        \includegraphics[width=0.15\textwidth]{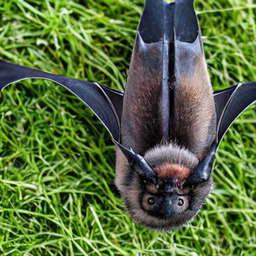}
        \includegraphics[width=0.15\textwidth]{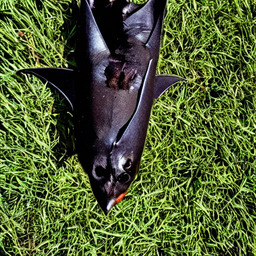}
        \caption{Unedited prompt encoding}
        \label{fig:bat_grass_amb}
    \end{subfigure}
    \begin{subfigure}{0.45\textwidth}
        \centering
        \includegraphics[width=0.3\textwidth]{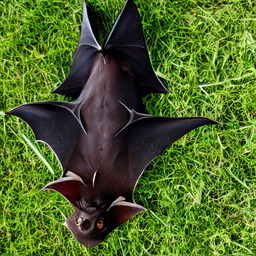}
        \includegraphics[width=0.3\textwidth]{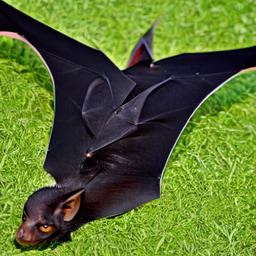}
        \includegraphics[width=0.3\textwidth]{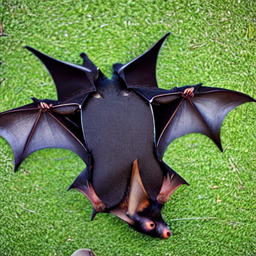}\\
        \includegraphics[width=0.3\textwidth]{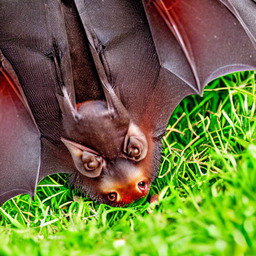}
        \includegraphics[width=0.3\textwidth]{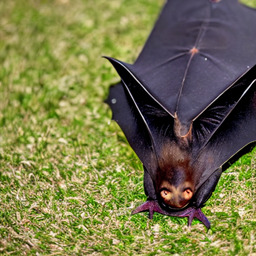}
        \includegraphics[width=0.3\textwidth]{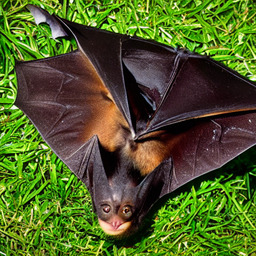}\\
        \includegraphics[width=0.3\textwidth]{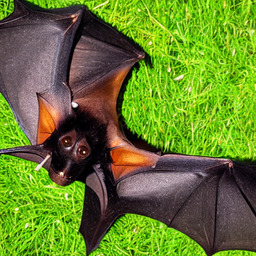}
        \includegraphics[width=0.3\textwidth]{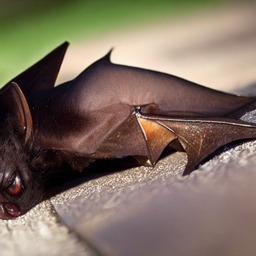}
        \includegraphics[width=0.3\textwidth]{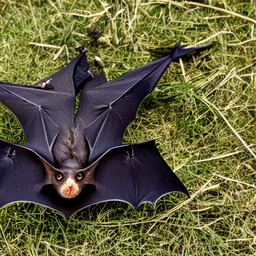}\\
        \includegraphics[width=0.3\textwidth]{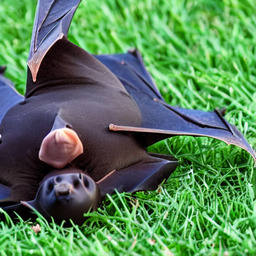}    \includegraphics[width=0.3\textwidth]{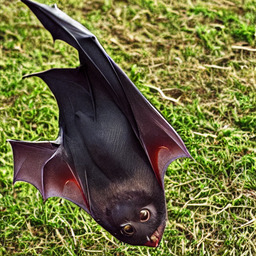}
        \includegraphics[width=0.3\textwidth]{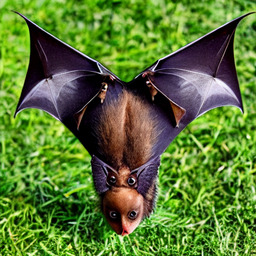}\\
        \includegraphics[width=0.3\textwidth]{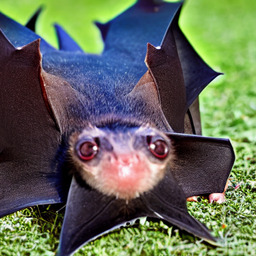}
        \includegraphics[width=0.3\textwidth]{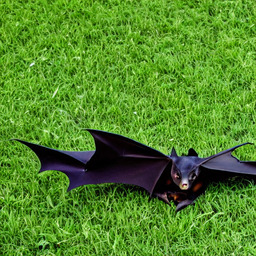}
        \includegraphics[width=0.3\textwidth]{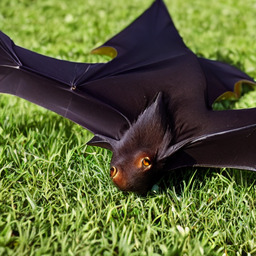}
        \caption{Encoding edited to favour animal sense}
        \label{fig:bat_grass_2}
    \end{subfigure}
    \begin{subfigure}{0.45\textwidth}
        \centering
        \includegraphics[width=0.3\textwidth]{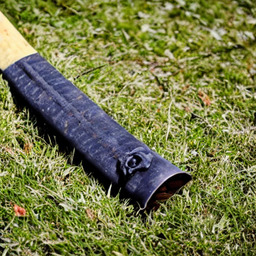}
        \includegraphics[width=0.3\textwidth]{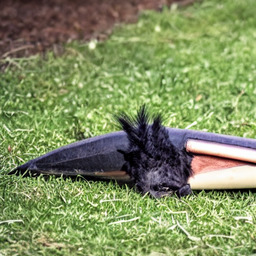}
        \includegraphics[width=0.3\textwidth]{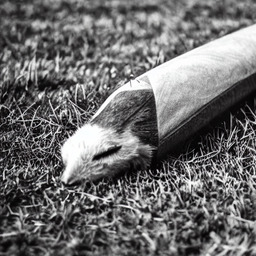}\\
        \includegraphics[width=0.3\textwidth]{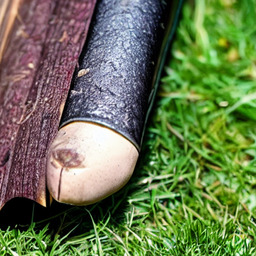}
        \includegraphics[width=0.3\textwidth]{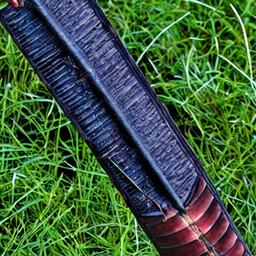}
        \includegraphics[width=0.3\textwidth]{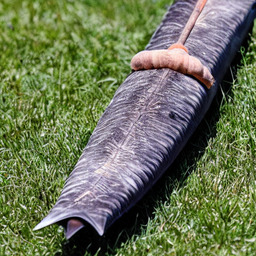}\\
        \includegraphics[width=0.3\textwidth]{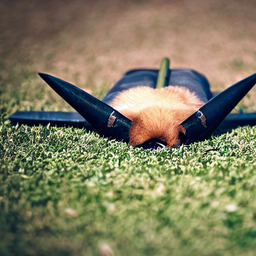}
        \includegraphics[width=0.3\textwidth]{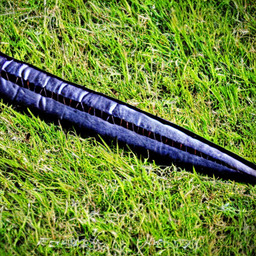}
        \includegraphics[width=0.3\textwidth]{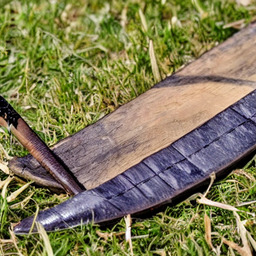}\\
        \includegraphics[width=0.3\textwidth]{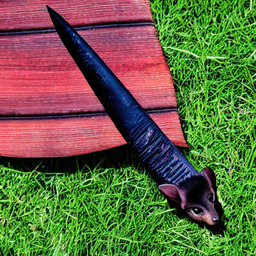}    \includegraphics[width=0.3\textwidth]{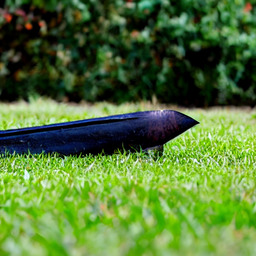}
        \includegraphics[width=0.3\textwidth]{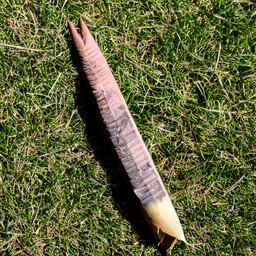}\\
        \includegraphics[width=0.3\textwidth]{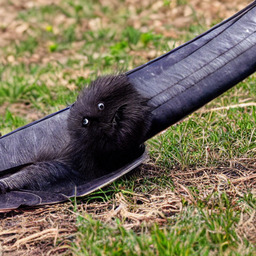}
        \includegraphics[width=0.3\textwidth]{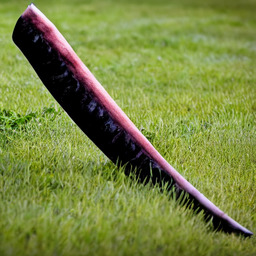}
        \includegraphics[width=0.3\textwidth]{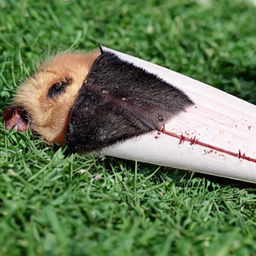}
        \caption{Encoding edited to favour sports-related sense}
        \label{fig:bat_grass_1}
    \end{subfigure}
    \caption{Prompt: \prompt{a bat laying on the grass}}
    \label{fig:abat_grass}
\end{figure}

\begin{figure}
    \centering
    \begin{subfigure}{\textwidth}
        \centering
        \includegraphics[width=0.15\textwidth]{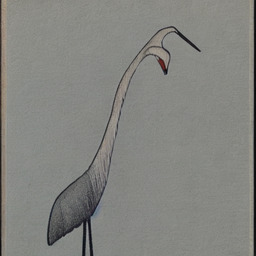}
        \includegraphics[width=0.15\textwidth]{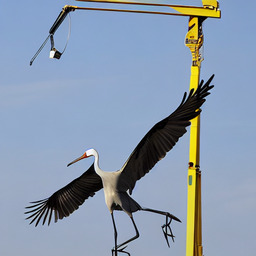}
        \includegraphics[width=0.15\textwidth]{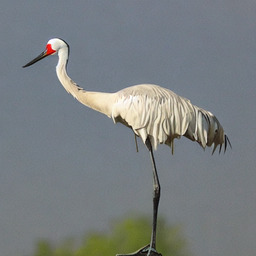}
        \includegraphics[width=0.15\textwidth]{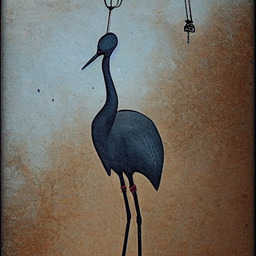}
        \includegraphics[width=0.15\textwidth]{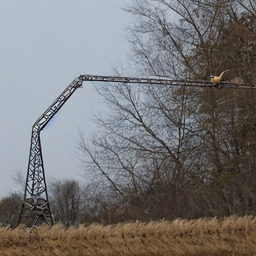}\\
        \includegraphics[width=0.15\textwidth]{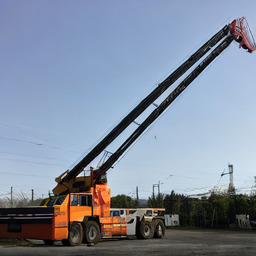}
        \includegraphics[width=0.15\textwidth]{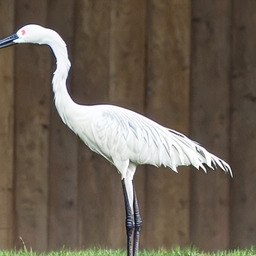}
        \includegraphics[width=0.15\textwidth]{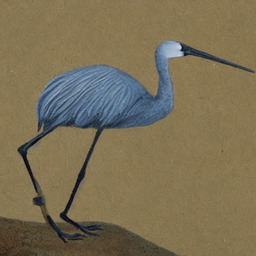}
        \includegraphics[width=0.15\textwidth]{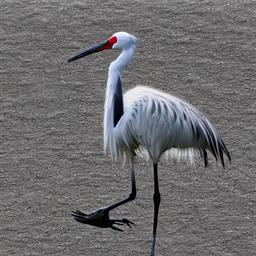}
        \includegraphics[width=0.15\textwidth]{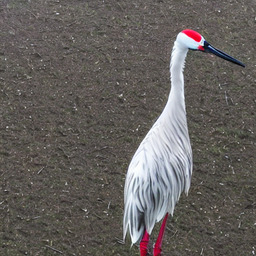}\\        \includegraphics[width=0.15\textwidth]{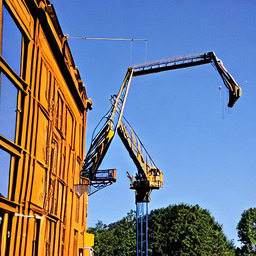}
        \includegraphics[width=0.15\textwidth]{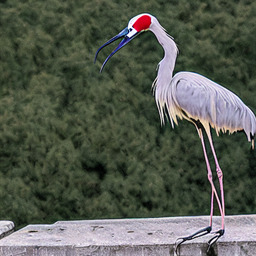}
        \includegraphics[width=0.15\textwidth]{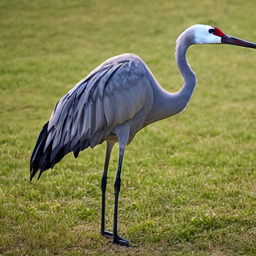}
        \includegraphics[width=0.15\textwidth]{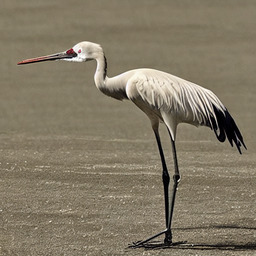}
        \includegraphics[width=0.15\textwidth]{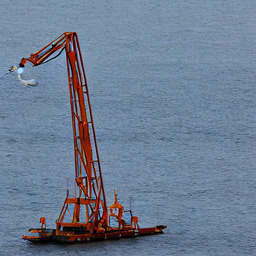}
        \caption{Unedited prompt encoding}
        \label{fig:crane_amb}
    \end{subfigure}
    \begin{subfigure}{0.45\textwidth}
        \centering
        \includegraphics[width=0.3\textwidth]{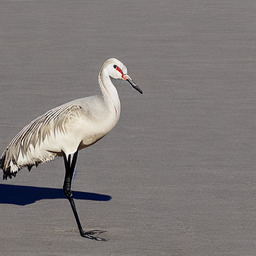}
        \includegraphics[width=0.3\textwidth]{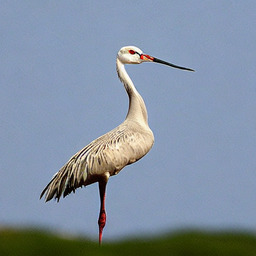}
        \includegraphics[width=0.3\textwidth]{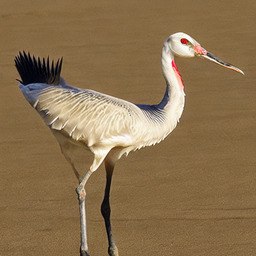}\\
        \includegraphics[width=0.3\textwidth]{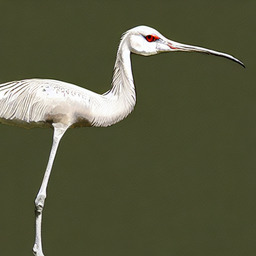}
        \includegraphics[width=0.3\textwidth]{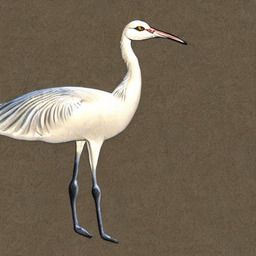}
        \includegraphics[width=0.3\textwidth]{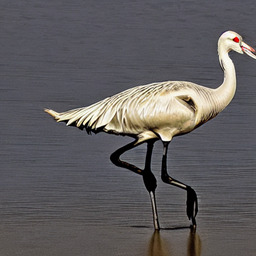}\\
        \includegraphics[width=0.3\textwidth]{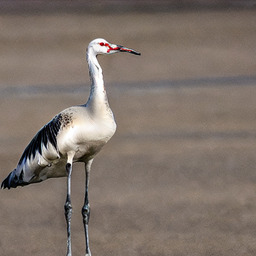}
        \includegraphics[width=0.3\textwidth]{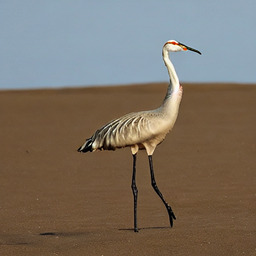}
        \includegraphics[width=0.3\textwidth]{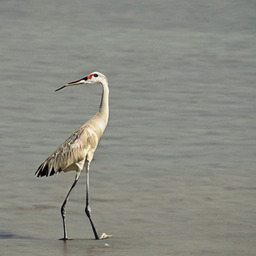}\\
        \includegraphics[width=0.3\textwidth]{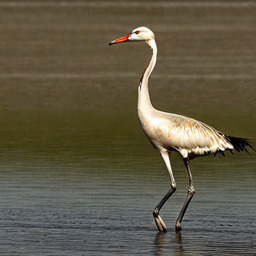}    \includegraphics[width=0.3\textwidth]{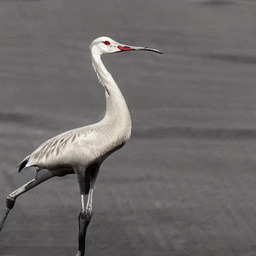}
        \includegraphics[width=0.3\textwidth]{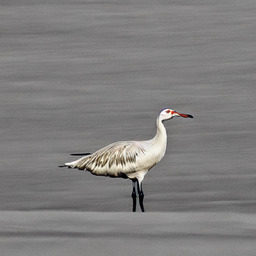}\\
        \includegraphics[width=0.3\textwidth]{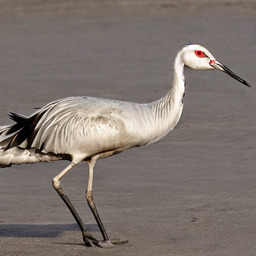}
        \includegraphics[width=0.3\textwidth]{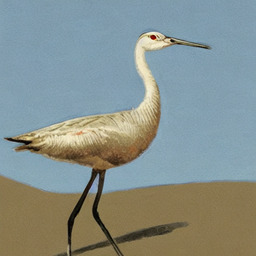}
        \includegraphics[width=0.3\textwidth]{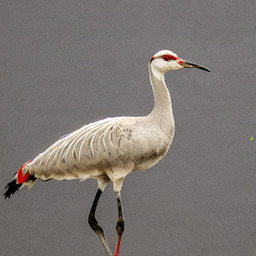}
        \caption{Encoding edited to favour bird sense}
        \label{fig:crane_2}
    \end{subfigure}
    \begin{subfigure}{0.45\textwidth}
        \centering
        \includegraphics[width=0.3\textwidth]{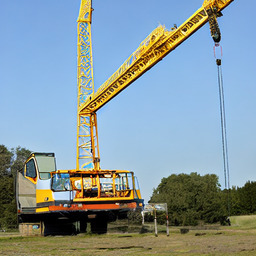}
        \includegraphics[width=0.3\textwidth]{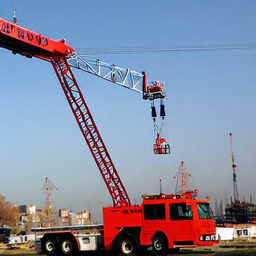}
        \includegraphics[width=0.3\textwidth]{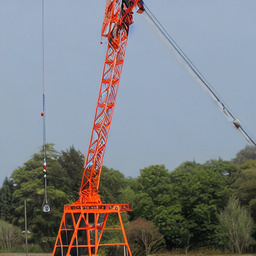}\\
        \includegraphics[width=0.3\textwidth]{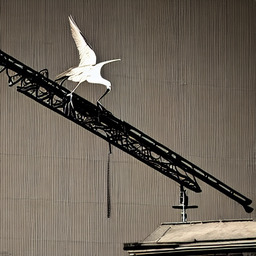}
        \includegraphics[width=0.3\textwidth]{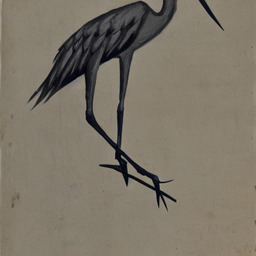}
        \includegraphics[width=0.3\textwidth]{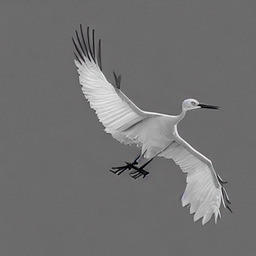}\\
        \includegraphics[width=0.3\textwidth]{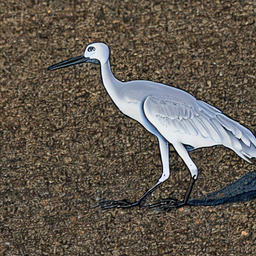}
        \includegraphics[width=0.3\textwidth]{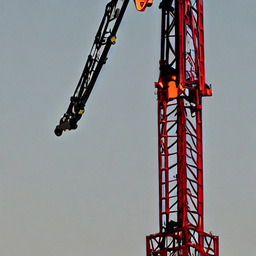}
        \includegraphics[width=0.3\textwidth]{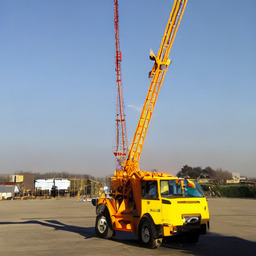}\\
        \includegraphics[width=0.3\textwidth]{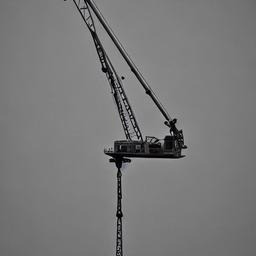}    \includegraphics[width=0.3\textwidth]{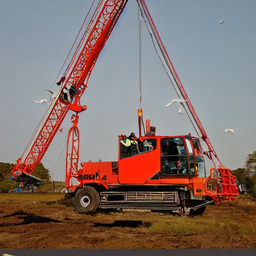}
        \includegraphics[width=0.3\textwidth]{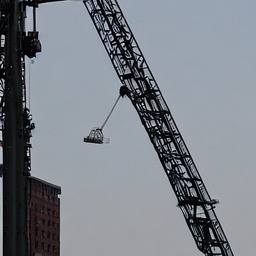}\\
        \includegraphics[width=0.3\textwidth]{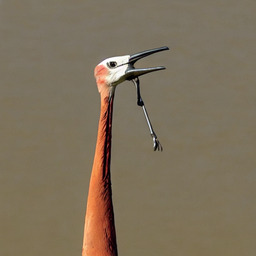}
        \includegraphics[width=0.3\textwidth]{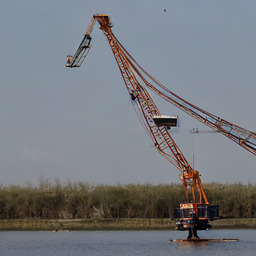}
        \includegraphics[width=0.3\textwidth]{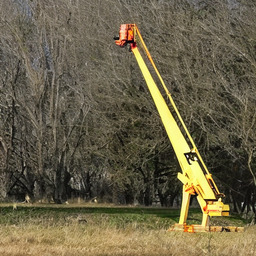}
        \caption{Encoding edited to favour construction-related sense}
        \label{fig:crane_1}
    \end{subfigure}
    \caption{Prompt: \prompt{a crane}}
    \label{fig:crane}
\end{figure}

\begin{figure}
    \centering
    \begin{subfigure}{\textwidth}
        \centering
        \includegraphics[width=0.15\textwidth]{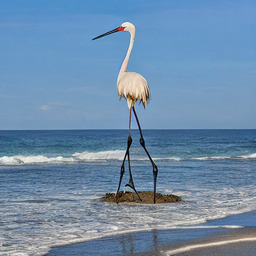}
        \includegraphics[width=0.15\textwidth]{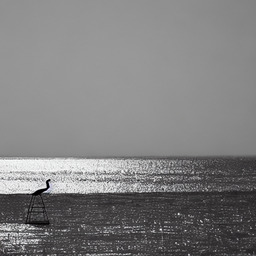}
        \includegraphics[width=0.15\textwidth]{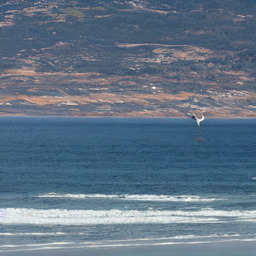}
        \includegraphics[width=0.15\textwidth]{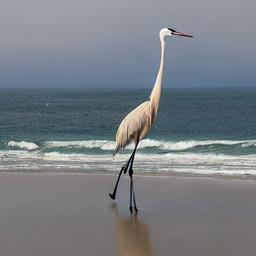}
        \includegraphics[width=0.15\textwidth]{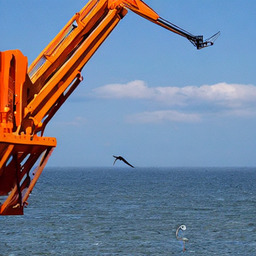}\\
        \includegraphics[width=0.15\textwidth]{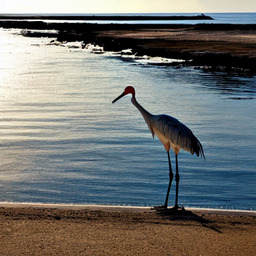}
        \includegraphics[width=0.15\textwidth]{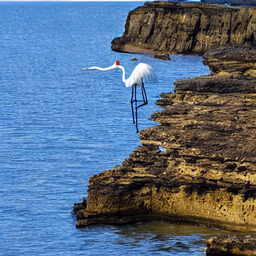}
        \includegraphics[width=0.15\textwidth]{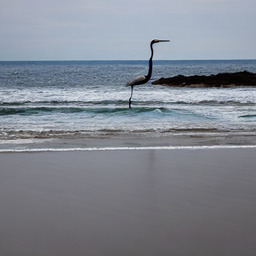}
        \includegraphics[width=0.15\textwidth]{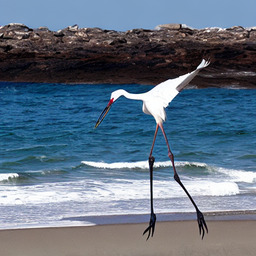}
        \includegraphics[width=0.15\textwidth]{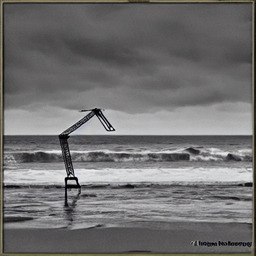}\\        \includegraphics[width=0.15\textwidth]{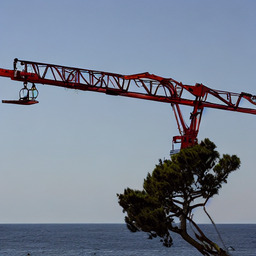}
        \includegraphics[width=0.15\textwidth]{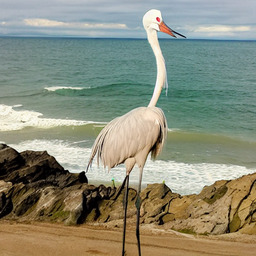}
        \includegraphics[width=0.15\textwidth]{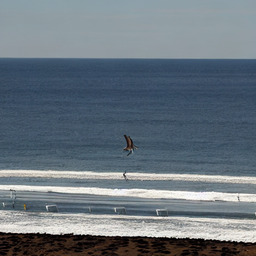}
        \includegraphics[width=0.15\textwidth]{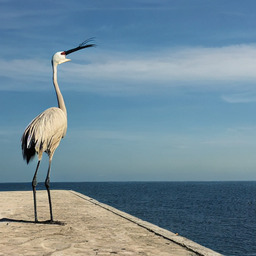}
        \includegraphics[width=0.15\textwidth]{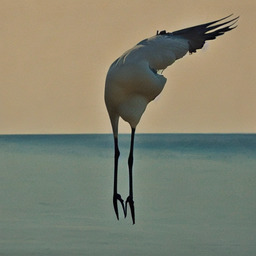}
        \caption{Unedited prompt encoding}
        \label{fig:crane_ocean_amb}
    \end{subfigure}
    \begin{subfigure}{0.45\textwidth}
        \centering
        \includegraphics[width=0.3\textwidth]{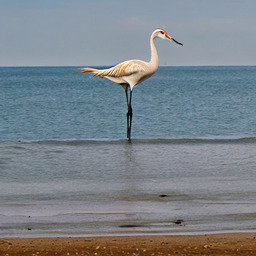}
        \includegraphics[width=0.3\textwidth]{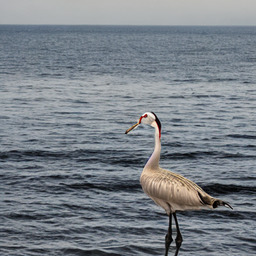}
        \includegraphics[width=0.3\textwidth]{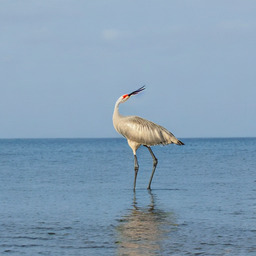}\\
        \includegraphics[width=0.3\textwidth]{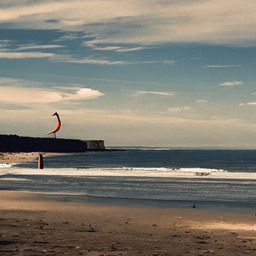}
        \includegraphics[width=0.3\textwidth]{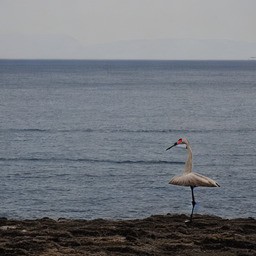}
        \includegraphics[width=0.3\textwidth]{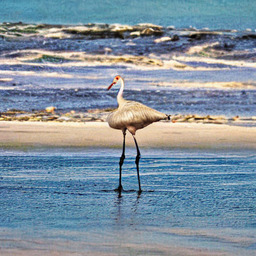}\\
        \includegraphics[width=0.3\textwidth]{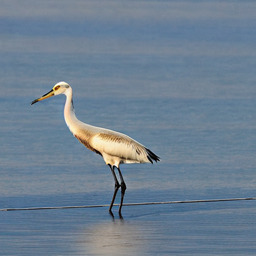}
        \includegraphics[width=0.3\textwidth]{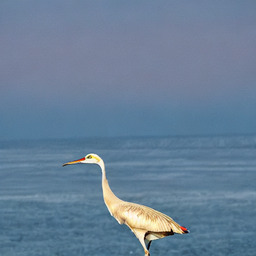}
        \includegraphics[width=0.3\textwidth]{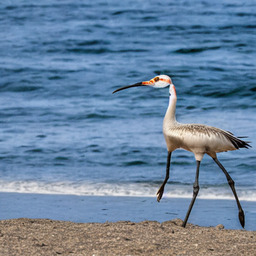}\\
        \includegraphics[width=0.3\textwidth]{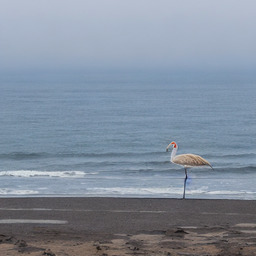}    \includegraphics[width=0.3\textwidth]{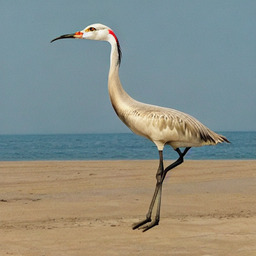}
        \includegraphics[width=0.3\textwidth]{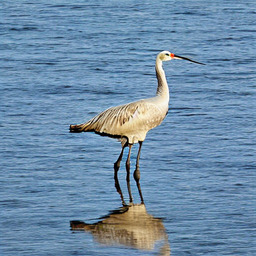}\\
        \includegraphics[width=0.3\textwidth]{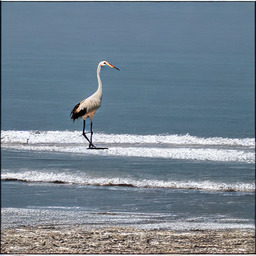}
        \includegraphics[width=0.3\textwidth]{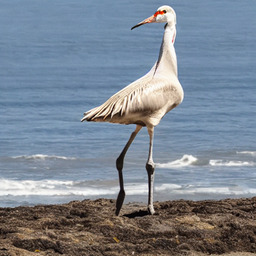}
        \includegraphics[width=0.3\textwidth]{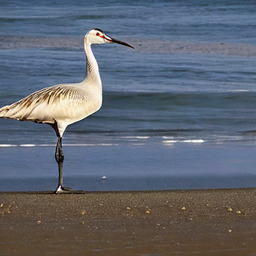}
        \caption{Encoding edited to favour bird sense}
        \label{fig:crane_ocean_2}
    \end{subfigure}
    \begin{subfigure}{0.45\textwidth}
        \centering
        \includegraphics[width=0.3\textwidth]{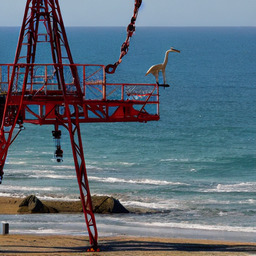}
        \includegraphics[width=0.3\textwidth]{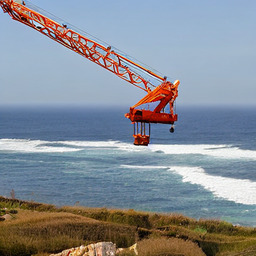}
        \includegraphics[width=0.3\textwidth]{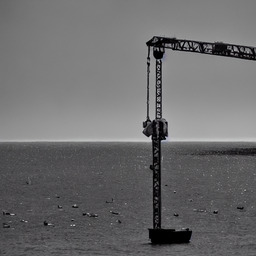}\\
        \includegraphics[width=0.3\textwidth]{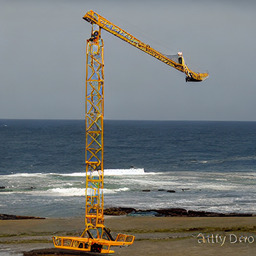}
        \includegraphics[width=0.3\textwidth]{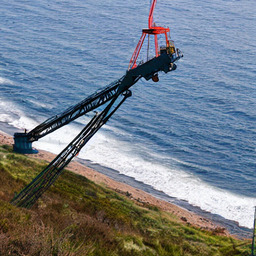}
        \includegraphics[width=0.3\textwidth]{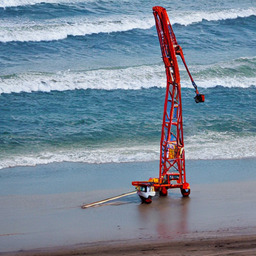}\\
        \includegraphics[width=0.3\textwidth]{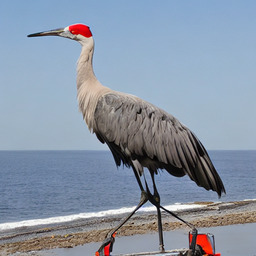}
        \includegraphics[width=0.3\textwidth]{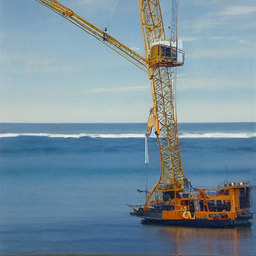}
        \includegraphics[width=0.3\textwidth]{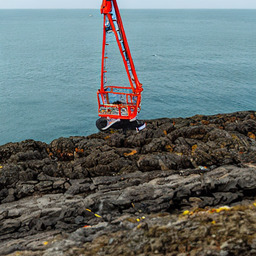}\\
        \includegraphics[width=0.3\textwidth]{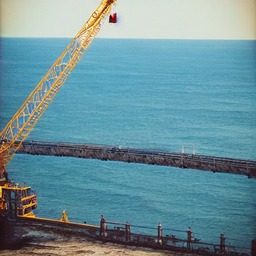}    \includegraphics[width=0.3\textwidth]{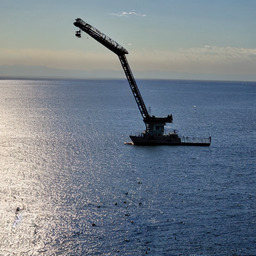}
        \includegraphics[width=0.3\textwidth]{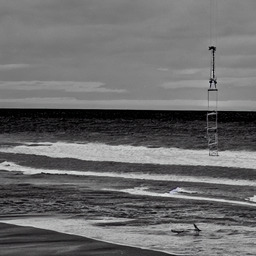}\\
        \includegraphics[width=0.3\textwidth]{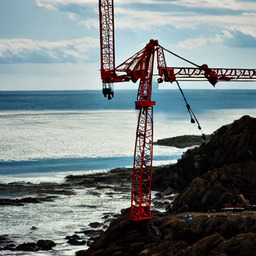}
        \includegraphics[width=0.3\textwidth]{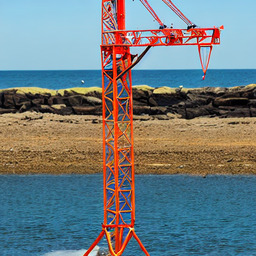}
        \includegraphics[width=0.3\textwidth]{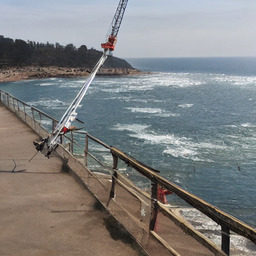}
        \caption{Encoding edited to favour construction-related sense}
        \label{fig:crane_ocean_1}
    \end{subfigure}
    \caption{Prompt: \prompt{a crane by the ocean}}
    \label{fig:crane_ocean}
\end{figure}

\begin{figure}
    \centering
    \begin{subfigure}{\textwidth}
        \centering
        \includegraphics[width=0.15\textwidth]{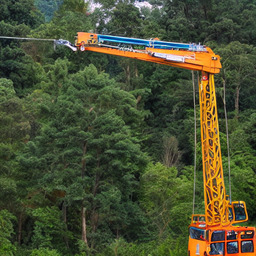}
        \includegraphics[width=0.15\textwidth]{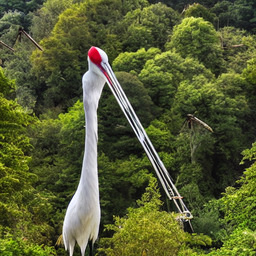}
        \includegraphics[width=0.15\textwidth]{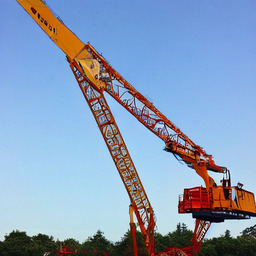}
        \includegraphics[width=0.15\textwidth]{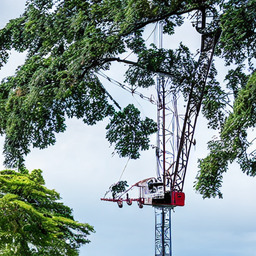}
        \includegraphics[width=0.15\textwidth]{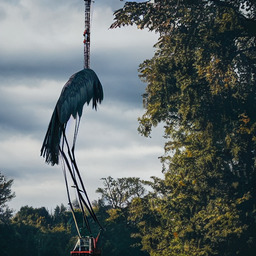}\\
        \includegraphics[width=0.15\textwidth]{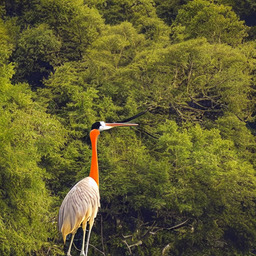}
        \includegraphics[width=0.15\textwidth]{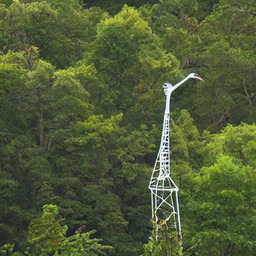}
        \includegraphics[width=0.15\textwidth]{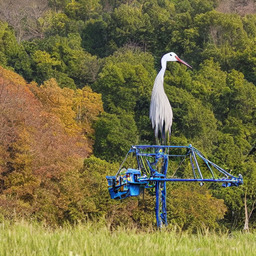}
        \includegraphics[width=0.15\textwidth]{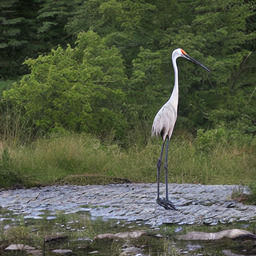}
        \includegraphics[width=0.15\textwidth]{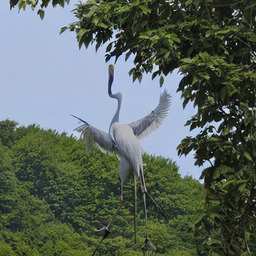}\\        \includegraphics[width=0.15\textwidth]{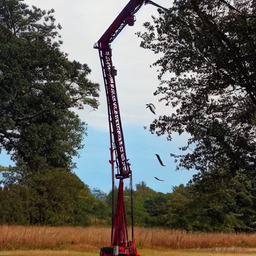}
        \includegraphics[width=0.15\textwidth]{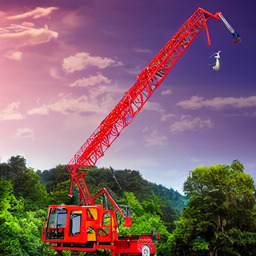}
        \includegraphics[width=0.15\textwidth]{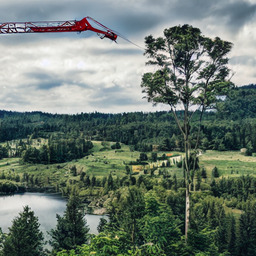}
        \includegraphics[width=0.15\textwidth]{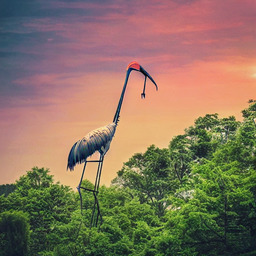}
        \includegraphics[width=0.15\textwidth]{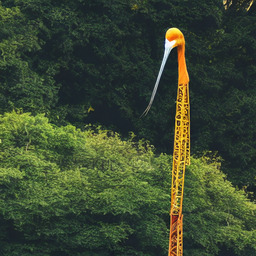}
        \caption{Unedited prompt encoding}
        \label{fig:crane_nature_amb}
    \end{subfigure}
    \begin{subfigure}{0.45\textwidth}
        \centering
        \includegraphics[width=0.3\textwidth]{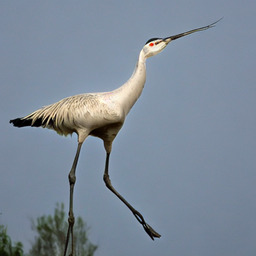}
        \includegraphics[width=0.3\textwidth]{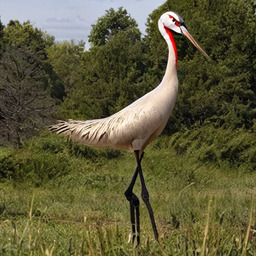}
        \includegraphics[width=0.3\textwidth]{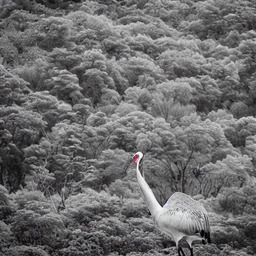}\\
        \includegraphics[width=0.3\textwidth]{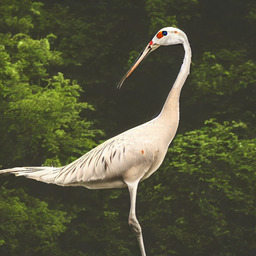}
        \includegraphics[width=0.3\textwidth]{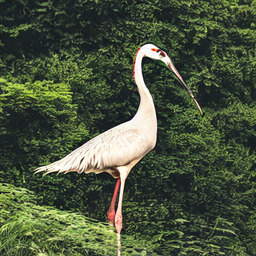}
        \includegraphics[width=0.3\textwidth]{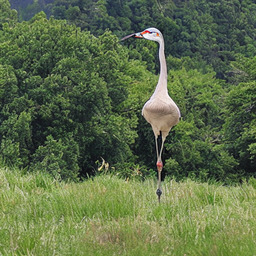}\\
        \includegraphics[width=0.3\textwidth]{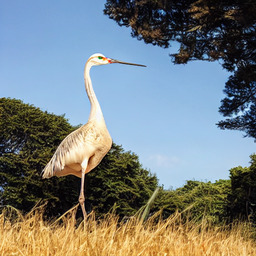}
        \includegraphics[width=0.3\textwidth]{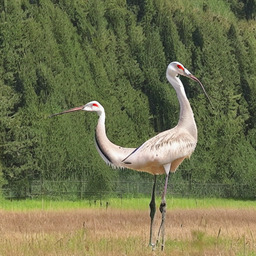}
        \includegraphics[width=0.3\textwidth]{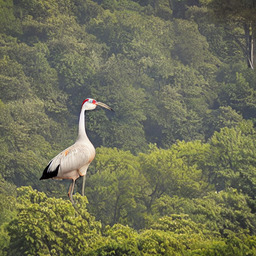}\\
        \includegraphics[width=0.3\textwidth]{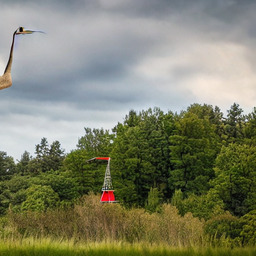}    \includegraphics[width=0.3\textwidth]{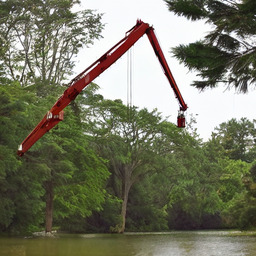}
        \includegraphics[width=0.3\textwidth]{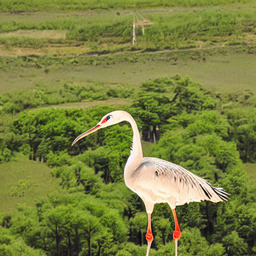}\\
        \includegraphics[width=0.3\textwidth]{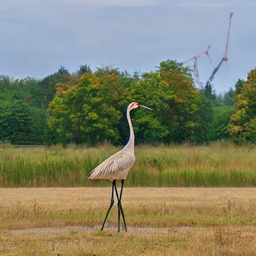}
        \includegraphics[width=0.3\textwidth]{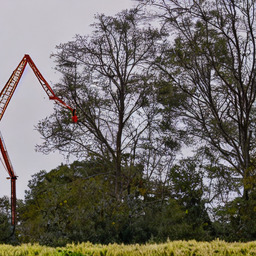}
        \includegraphics[width=0.3\textwidth]{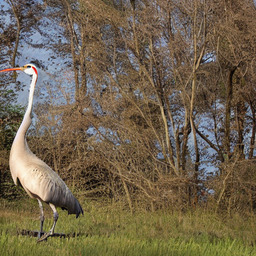}
        \caption{Encoding edited to favour bird sense}
        \label{fig:crane_nature_2}
    \end{subfigure}
    \begin{subfigure}{0.45\textwidth}
        \centering
        \includegraphics[width=0.3\textwidth]{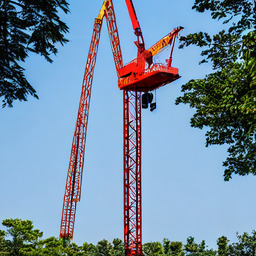}
        \includegraphics[width=0.3\textwidth]{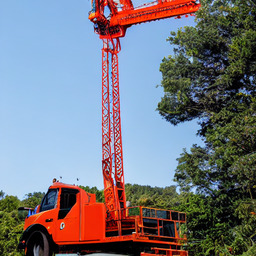}
        \includegraphics[width=0.3\textwidth]{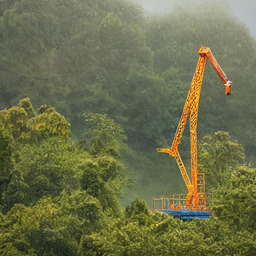}\\
        \includegraphics[width=0.3\textwidth]{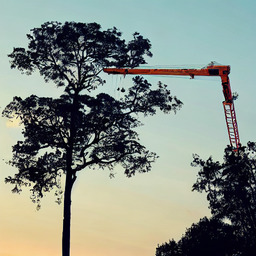}
        \includegraphics[width=0.3\textwidth]{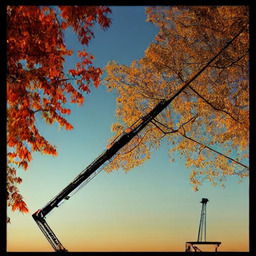}
        \includegraphics[width=0.3\textwidth]{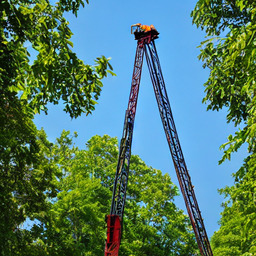}\\
        \includegraphics[width=0.3\textwidth]{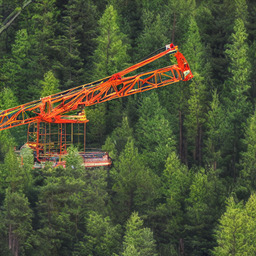}
        \includegraphics[width=0.3\textwidth]{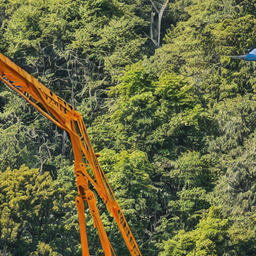}
        \includegraphics[width=0.3\textwidth]{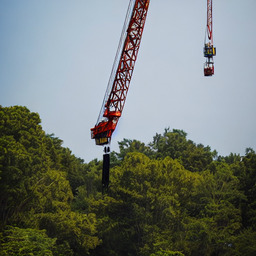}\\
        \includegraphics[width=0.3\textwidth]{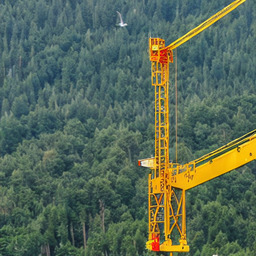}    \includegraphics[width=0.3\textwidth]{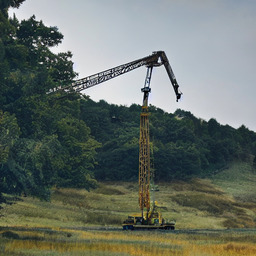}
        \includegraphics[width=0.3\textwidth]{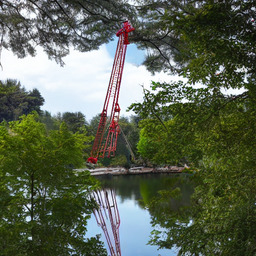}\\
        \includegraphics[width=0.3\textwidth]{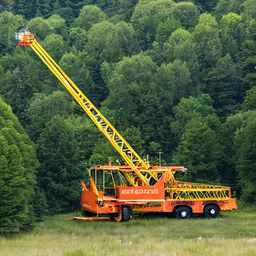}
        \includegraphics[width=0.3\textwidth]{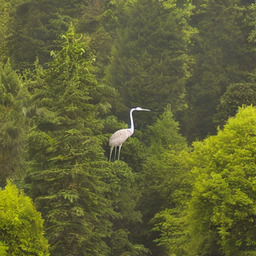}
        \includegraphics[width=0.3\textwidth]{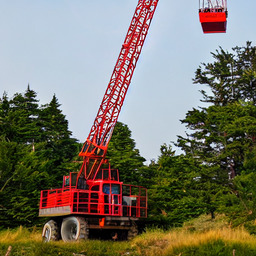}
        \caption{Encoding edited to favour construction-related sense}
        \label{fig:crane_nature_1}
    \end{subfigure}
    \caption{Prompt: \prompt{a crane surrounded by nature}}
    \label{fig:crane_nature}
\end{figure}

\begin{figure}
    \centering
    \begin{subfigure}{\textwidth}
        \centering
        \includegraphics[width=0.15\textwidth]{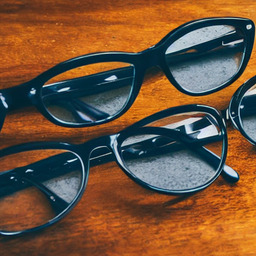}
        \includegraphics[width=0.15\textwidth]{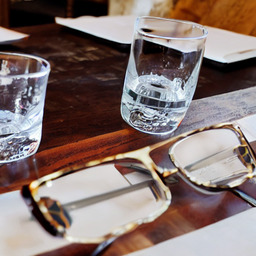}
        \includegraphics[width=0.15\textwidth]{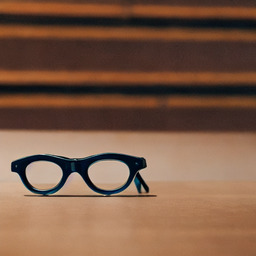}
        \includegraphics[width=0.15\textwidth]{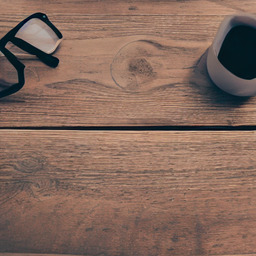}
        \includegraphics[width=0.15\textwidth]{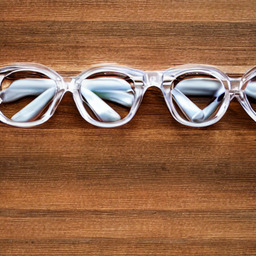}\\
        \includegraphics[width=0.15\textwidth]{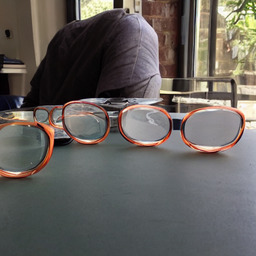}
        \includegraphics[width=0.15\textwidth]{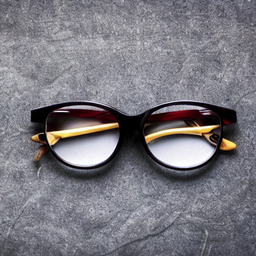}
        \includegraphics[width=0.15\textwidth]{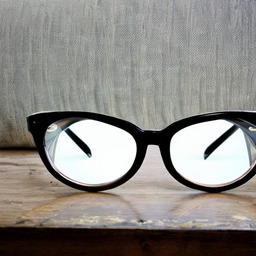}
        \includegraphics[width=0.15\textwidth]{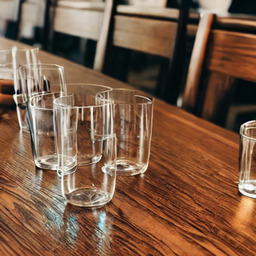}
        \includegraphics[width=0.15\textwidth]{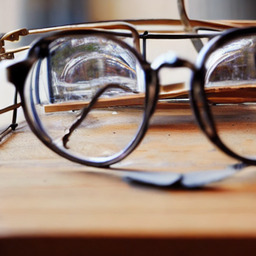}\\        \includegraphics[width=0.15\textwidth]{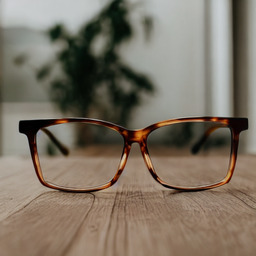}
        \includegraphics[width=0.15\textwidth]{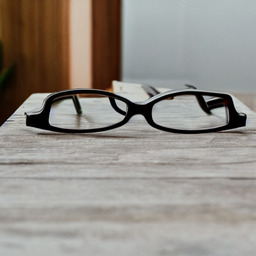}
        \includegraphics[width=0.15\textwidth]{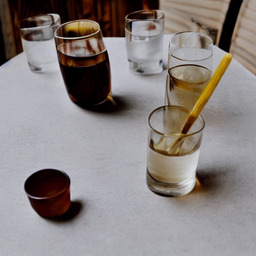}
        \includegraphics[width=0.15\textwidth]{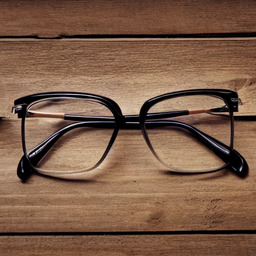}
        \includegraphics[width=0.15\textwidth]{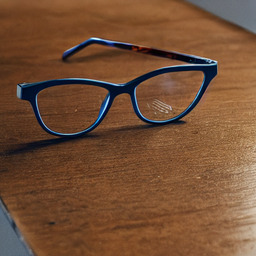}
        \caption{Unedited prompt encoding}
        \label{fig:glasses_amb}
    \end{subfigure}
    \begin{subfigure}{0.45\textwidth}
        \centering
        \includegraphics[width=0.3\textwidth]{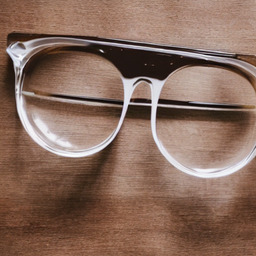}
        \includegraphics[width=0.3\textwidth]{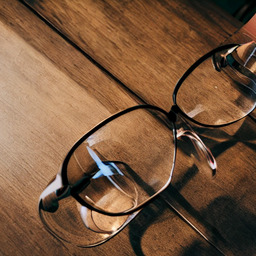}
        \includegraphics[width=0.3\textwidth]{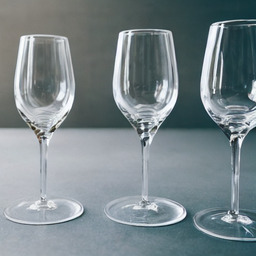}\\
        \includegraphics[width=0.3\textwidth]{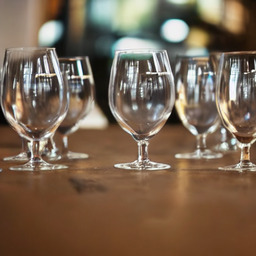}
        \includegraphics[width=0.3\textwidth]{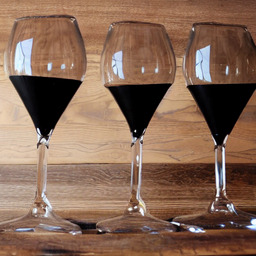}
        \includegraphics[width=0.3\textwidth]{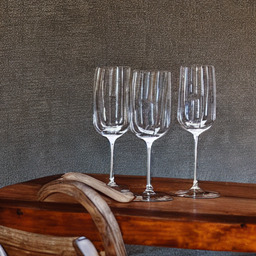}\\
        \includegraphics[width=0.3\textwidth]{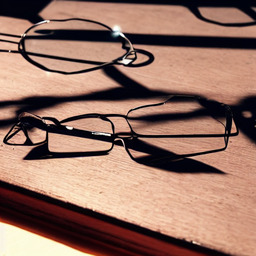}
        \includegraphics[width=0.3\textwidth]{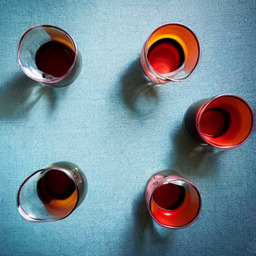}
        \includegraphics[width=0.3\textwidth]{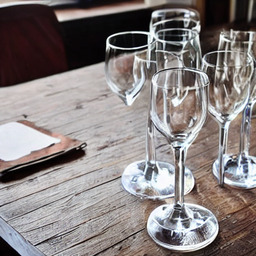}\\
        \includegraphics[width=0.3\textwidth]{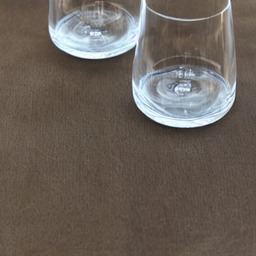}    \includegraphics[width=0.3\textwidth]{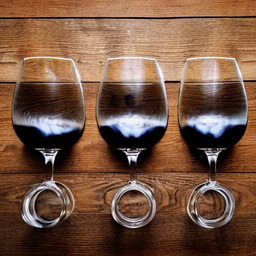}
        \includegraphics[width=0.3\textwidth]{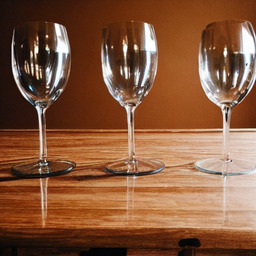}\\
        \includegraphics[width=0.3\textwidth]{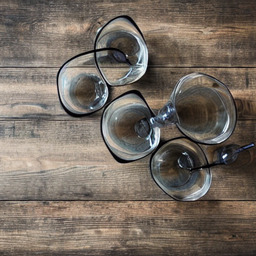}
        \includegraphics[width=0.3\textwidth]{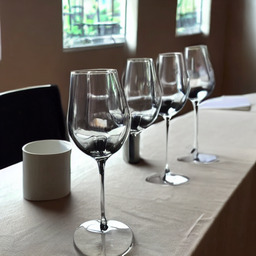}
        \includegraphics[width=0.3\textwidth]{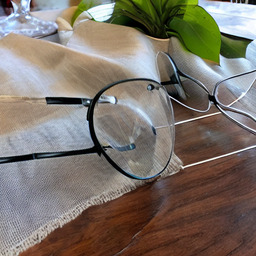}
        \caption{Encoding edited to favour drinking-related sense}
        \label{fig:glasses_1}
    \end{subfigure}
    \begin{subfigure}{0.45\textwidth}
        \centering
        \includegraphics[width=0.3\textwidth]{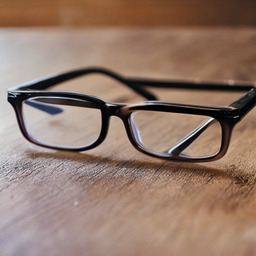}
        \includegraphics[width=0.3\textwidth]{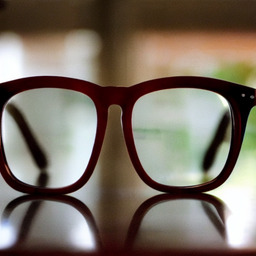}
        \includegraphics[width=0.3\textwidth]{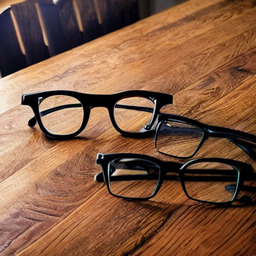}\\
        \includegraphics[width=0.3\textwidth]{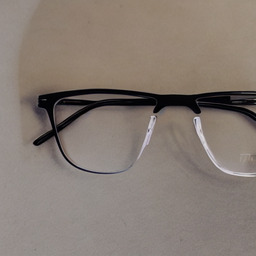}
        \includegraphics[width=0.3\textwidth]{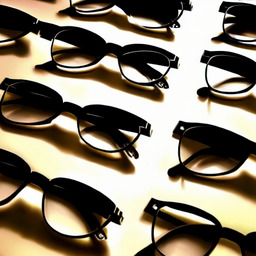}
        \includegraphics[width=0.3\textwidth]{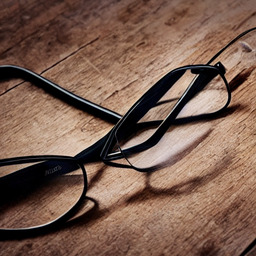}\\
        \includegraphics[width=0.3\textwidth]{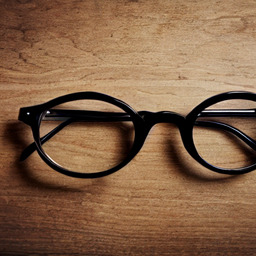}
        \includegraphics[width=0.3\textwidth]{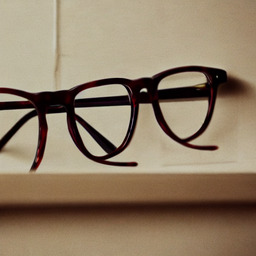}
        \includegraphics[width=0.3\textwidth]{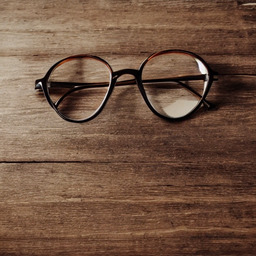}\\
        \includegraphics[width=0.3\textwidth]{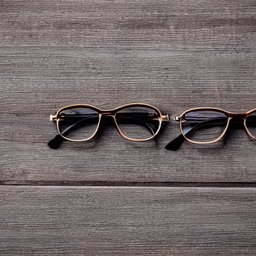}    \includegraphics[width=0.3\textwidth]{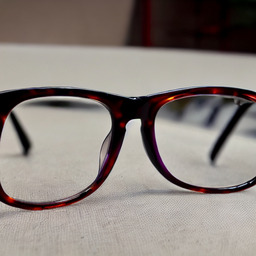}
        \includegraphics[width=0.3\textwidth]{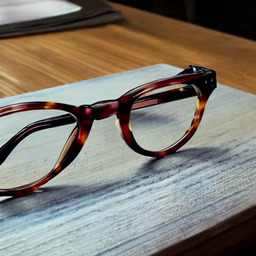}\\
        \includegraphics[width=0.3\textwidth]{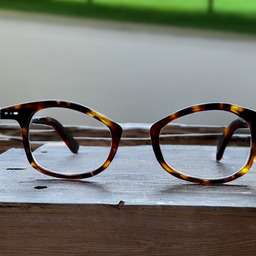}
        \includegraphics[width=0.3\textwidth]{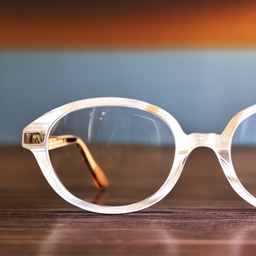}
        \includegraphics[width=0.3\textwidth]{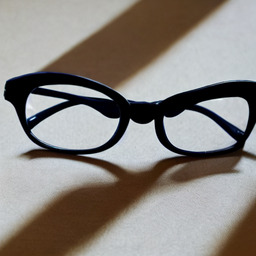}
        \caption{Encoding edited to favour eyeglasses sense}
        \label{fig:glasses_2}
    \end{subfigure}
    \caption{Prompt: \prompt{glasses on a table}}
    \label{fig:glasses}
\end{figure}
\begin{figure}
    \centering
    \begin{subfigure}{\textwidth}
        \centering
        \includegraphics[width=0.15\textwidth]{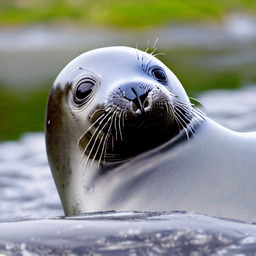}
        \includegraphics[width=0.15\textwidth]{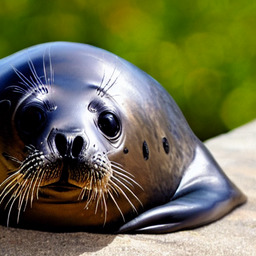}
        \includegraphics[width=0.15\textwidth]{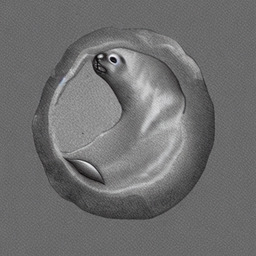}
        \includegraphics[width=0.15\textwidth]{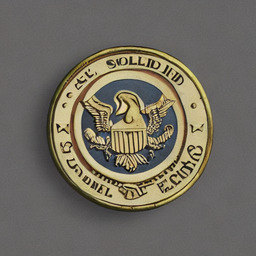}
        \includegraphics[width=0.15\textwidth]{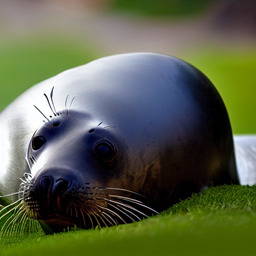}\\
        \includegraphics[width=0.15\textwidth]{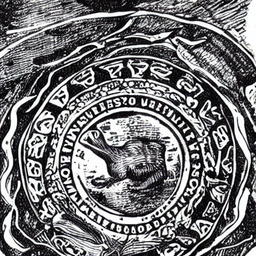}
        \includegraphics[width=0.15\textwidth]{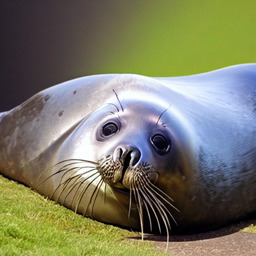}
        \includegraphics[width=0.15\textwidth]{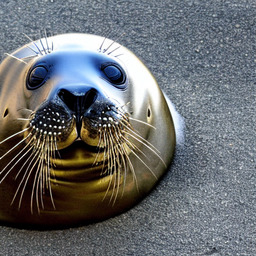}
        \includegraphics[width=0.15\textwidth]{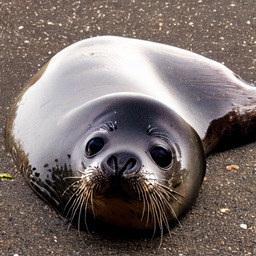}
        \includegraphics[width=0.15\textwidth]{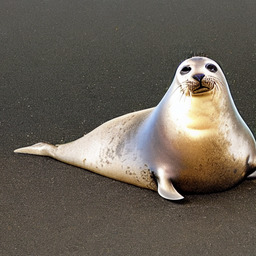}\\        \includegraphics[width=0.15\textwidth]{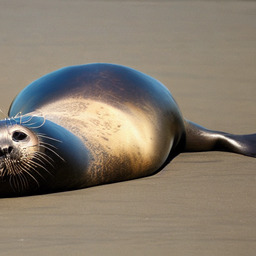}
        \includegraphics[width=0.15\textwidth]{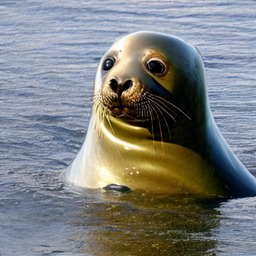}
        \includegraphics[width=0.15\textwidth]{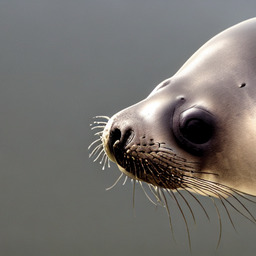}
        \includegraphics[width=0.15\textwidth]{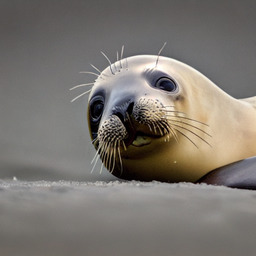}
        \includegraphics[width=0.15\textwidth]{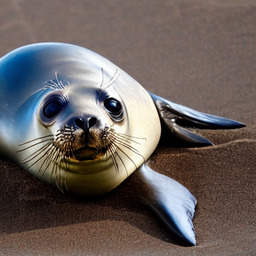}
        \caption{Unedited prompt encoding}
        \label{fig:seal_amb}
    \end{subfigure}
    \begin{subfigure}{0.45\textwidth}
        \centering
        \includegraphics[width=0.3\textwidth]{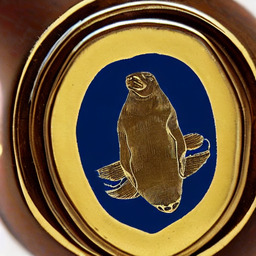}
        \includegraphics[width=0.3\textwidth]{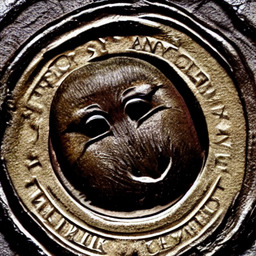}
        \includegraphics[width=0.3\textwidth]{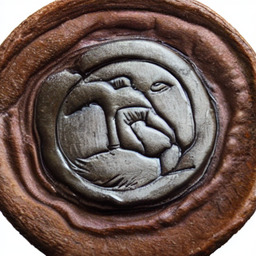}\\
        \includegraphics[width=0.3\textwidth]{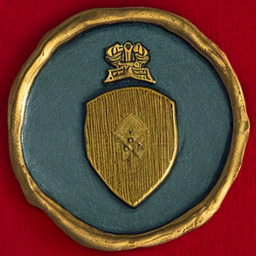}
        \includegraphics[width=0.3\textwidth]{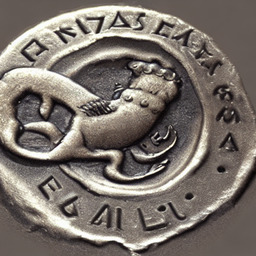}
        \includegraphics[width=0.3\textwidth]{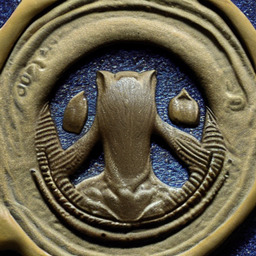}\\
        \includegraphics[width=0.3\textwidth]{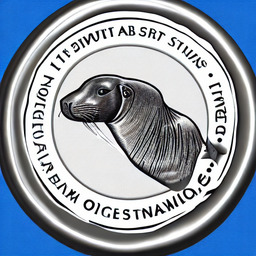}
        \includegraphics[width=0.3\textwidth]{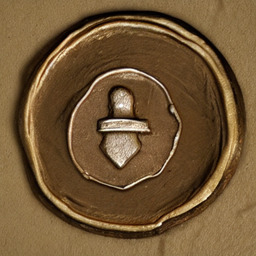}
        \includegraphics[width=0.3\textwidth]{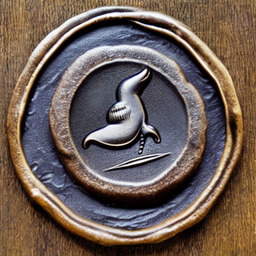}\\
        \includegraphics[width=0.3\textwidth]{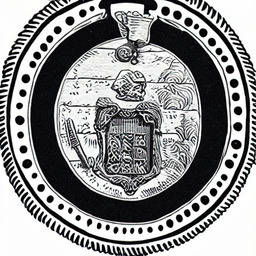}    \includegraphics[width=0.3\textwidth]{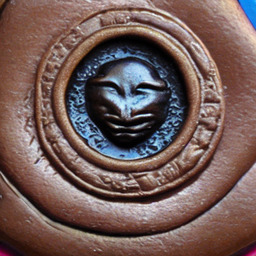}
        \includegraphics[width=0.3\textwidth]{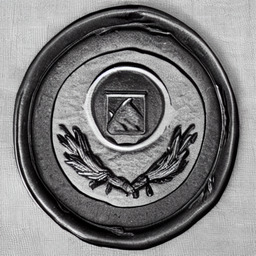}\\
        \includegraphics[width=0.3\textwidth]{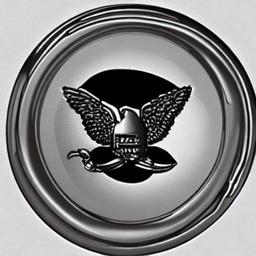}
        \includegraphics[width=0.3\textwidth]{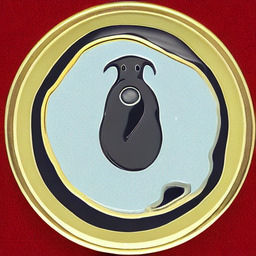}
        \includegraphics[width=0.3\textwidth]{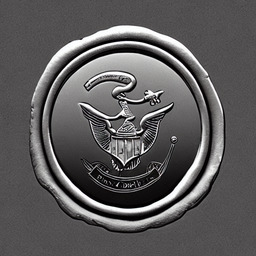}
        \caption{Encoding edited to favour sense of wax or official seal}
        \label{fig:seal_1}
    \end{subfigure}
    \begin{subfigure}{0.45\textwidth}
        \centering
        \includegraphics[width=0.3\textwidth]{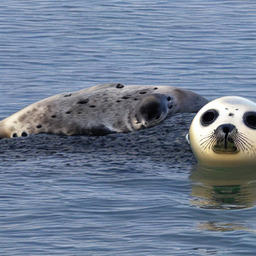}
        \includegraphics[width=0.3\textwidth]{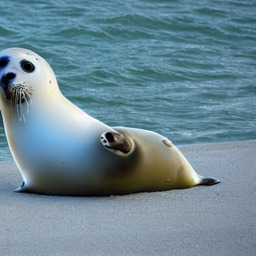}
        \includegraphics[width=0.3\textwidth]{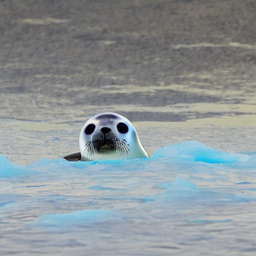}\\
        \includegraphics[width=0.3\textwidth]{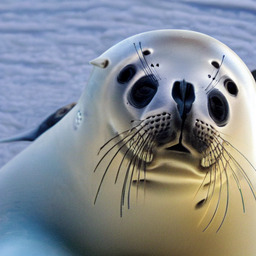}
        \includegraphics[width=0.3\textwidth]{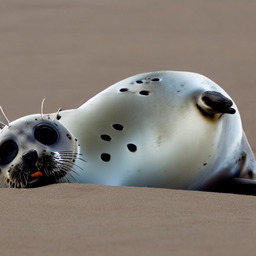}
        \includegraphics[width=0.3\textwidth]{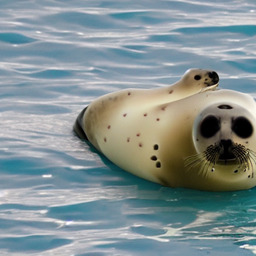}\\
        \includegraphics[width=0.3\textwidth]{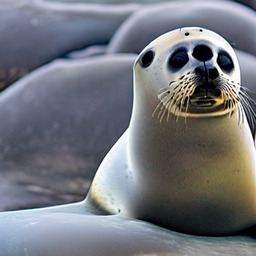}
        \includegraphics[width=0.3\textwidth]{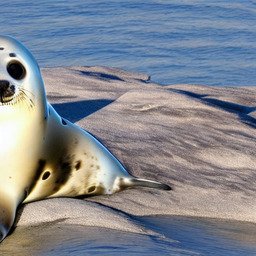}
        \includegraphics[width=0.3\textwidth]{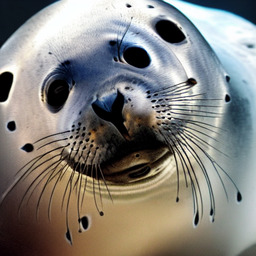}\\
        \includegraphics[width=0.3\textwidth]{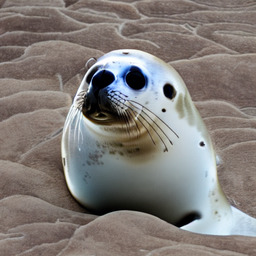}    \includegraphics[width=0.3\textwidth]{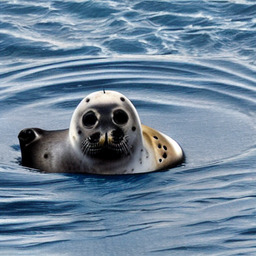}
        \includegraphics[width=0.3\textwidth]{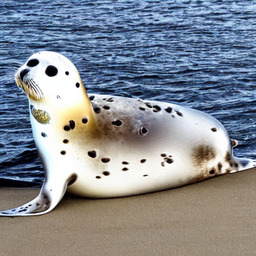}\\
        \includegraphics[width=0.3\textwidth]{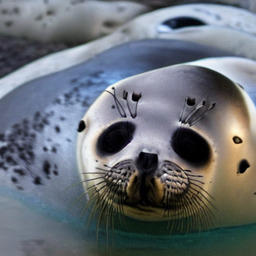}
        \includegraphics[width=0.3\textwidth]{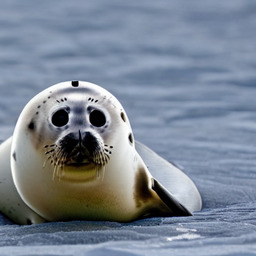}
        \includegraphics[width=0.3\textwidth]{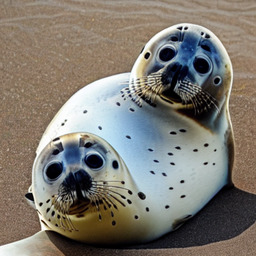}
        \caption{Encoding edited to favour animal sense}
        \label{fig:seal_2}
    \end{subfigure}
    \caption{Prompt: \prompt{a seal}}
    \label{fig:aseal}
\end{figure}
\begin{figure}
    \centering
    \begin{subfigure}{\textwidth}
        \centering
        \includegraphics[width=0.15\textwidth]{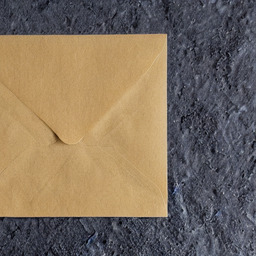}
        \includegraphics[width=0.15\textwidth]{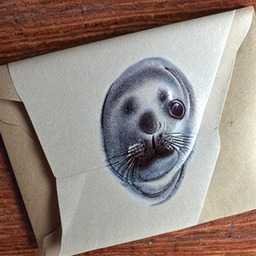}
        \includegraphics[width=0.15\textwidth]{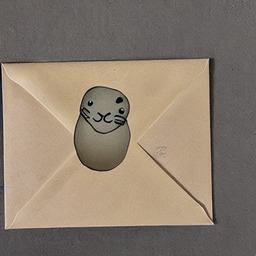}
        \includegraphics[width=0.15\textwidth]{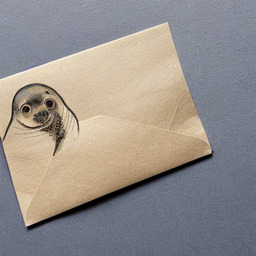}
        \includegraphics[width=0.15\textwidth]{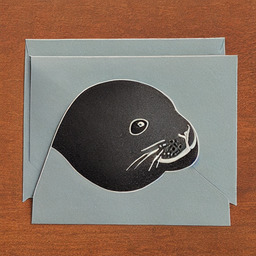}\\
        \includegraphics[width=0.15\textwidth]{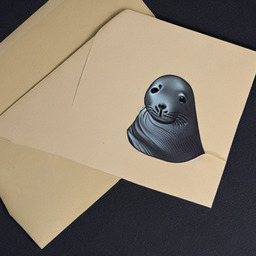}
        \includegraphics[width=0.15\textwidth]{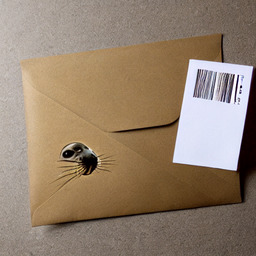}
        \includegraphics[width=0.15\textwidth]{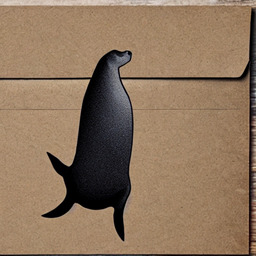}
        \includegraphics[width=0.15\textwidth]{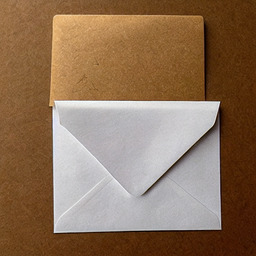}
        \includegraphics[width=0.15\textwidth]{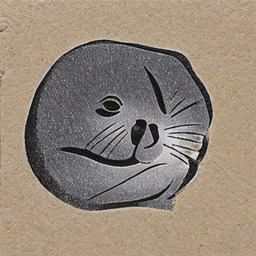}\\        \includegraphics[width=0.15\textwidth]{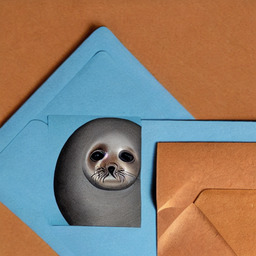}
        \includegraphics[width=0.15\textwidth]{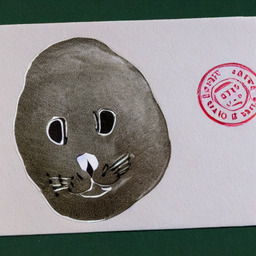}
        \includegraphics[width=0.15\textwidth]{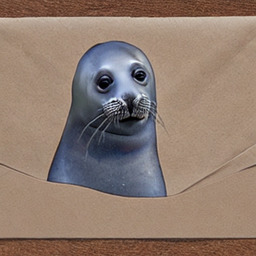}
        \includegraphics[width=0.15\textwidth]{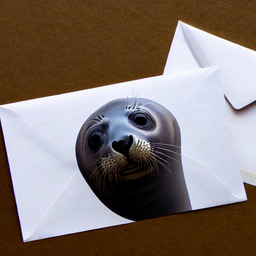}
        \includegraphics[width=0.15\textwidth]{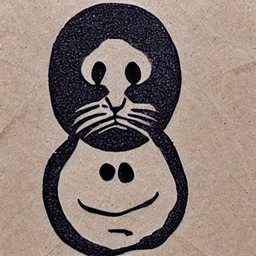}
        \caption{Unedited prompt encoding}
        \label{fig:seal_env_amb}
    \end{subfigure}
    \begin{subfigure}{0.45\textwidth}
        \centering
        \includegraphics[width=0.3\textwidth]{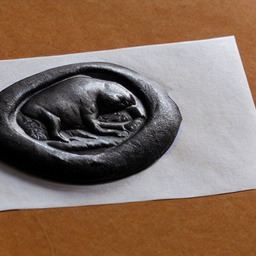}
        \includegraphics[width=0.3\textwidth]{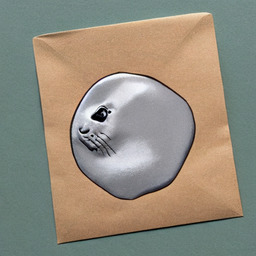}
        \includegraphics[width=0.3\textwidth]{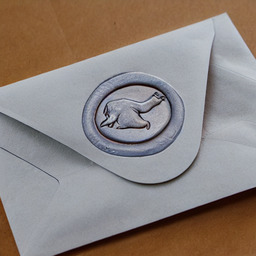}\\
        \includegraphics[width=0.3\textwidth]{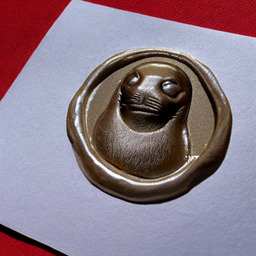}
        \includegraphics[width=0.3\textwidth]{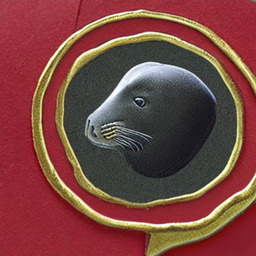}
        \includegraphics[width=0.3\textwidth]{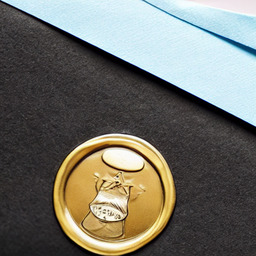}\\
        \includegraphics[width=0.3\textwidth]{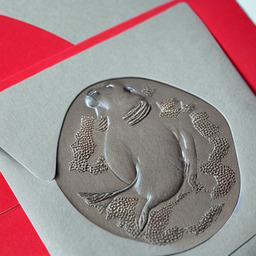}
        \includegraphics[width=0.3\textwidth]{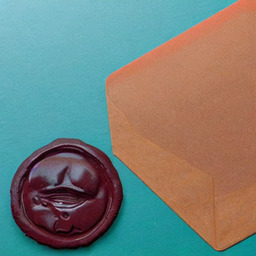}
        \includegraphics[width=0.3\textwidth]{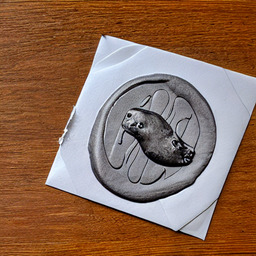}\\
        \includegraphics[width=0.3\textwidth]{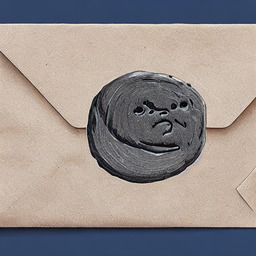}    \includegraphics[width=0.3\textwidth]{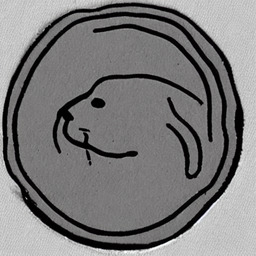}
        \includegraphics[width=0.3\textwidth]{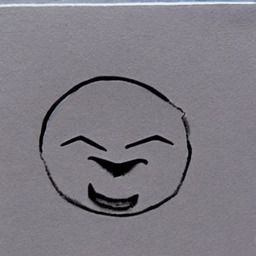}\\
        \includegraphics[width=0.3\textwidth]{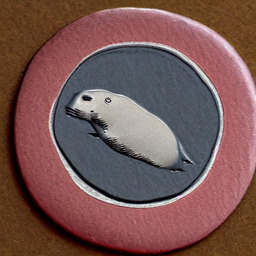}
        \includegraphics[width=0.3\textwidth]{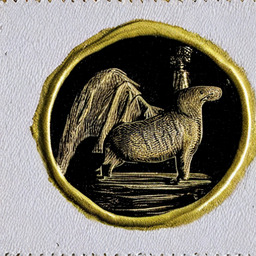}
        \includegraphics[width=0.3\textwidth]{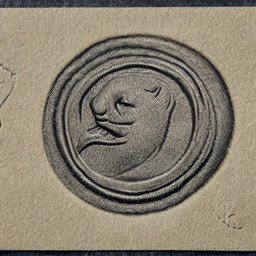}
        \caption{Encoding edited to favour sense of wax or official seal}
        \label{fig:seal_env_1}
    \end{subfigure}
    \begin{subfigure}{0.45\textwidth}
        \centering
        \includegraphics[width=0.3\textwidth]{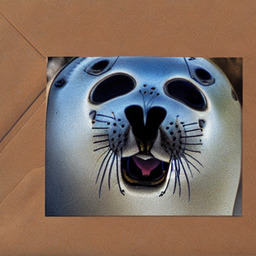}
        \includegraphics[width=0.3\textwidth]{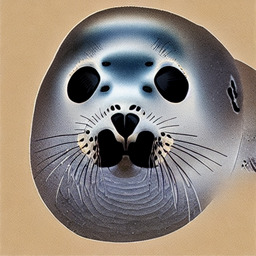}
        \includegraphics[width=0.3\textwidth]{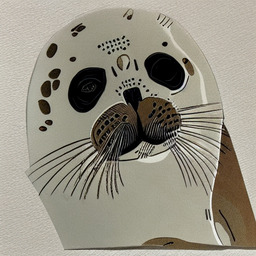}\\
        \includegraphics[width=0.3\textwidth]{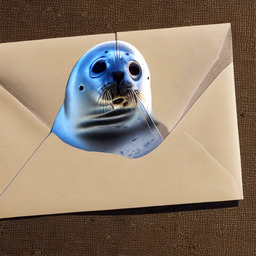}
        \includegraphics[width=0.3\textwidth]{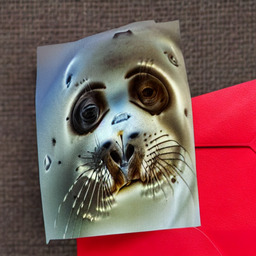}
        \includegraphics[width=0.3\textwidth]{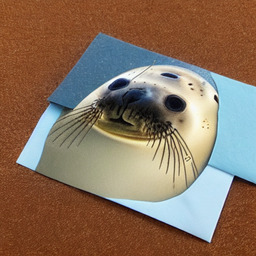}\\
        \includegraphics[width=0.3\textwidth]{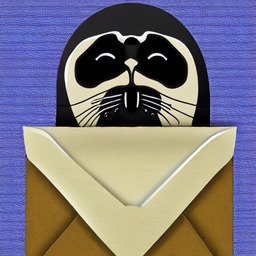}
        \includegraphics[width=0.3\textwidth]{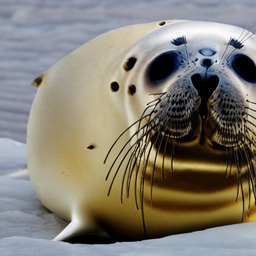}
        \includegraphics[width=0.3\textwidth]{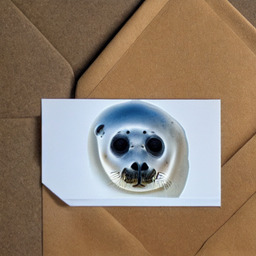}\\
        \includegraphics[width=0.3\textwidth]{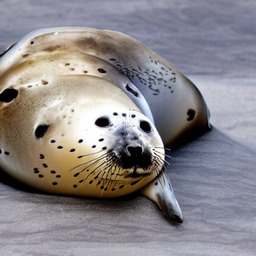}    \includegraphics[width=0.3\textwidth]{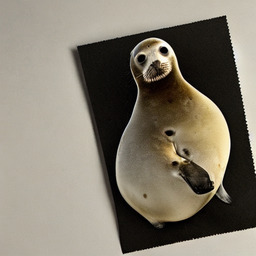}
        \includegraphics[width=0.3\textwidth]{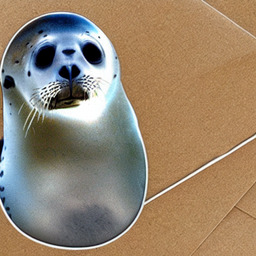}\\
        \includegraphics[width=0.3\textwidth]{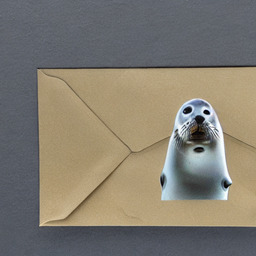}
        \includegraphics[width=0.3\textwidth]{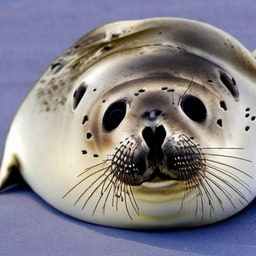}
        \includegraphics[width=0.3\textwidth]{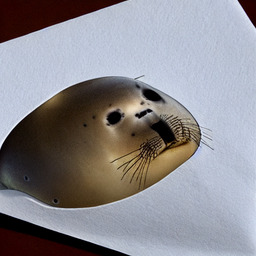}
        \caption{Encoding edited to favour animal sense}
        \label{fig:seal_env_2}
    \end{subfigure}
    \caption{Prompt: \prompt{a seal on an envelope}}
    \label{fig:seal_envelope}
\end{figure}
\begin{figure}
    \centering
    \begin{subfigure}{\textwidth}
        \centering
        \includegraphics[width=0.15\textwidth]{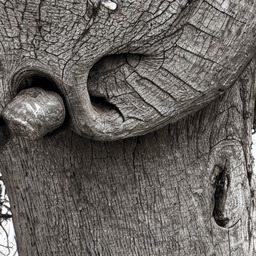}
        \includegraphics[width=0.15\textwidth]{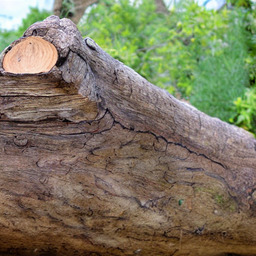}
        \includegraphics[width=0.15\textwidth]{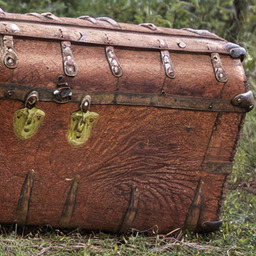}
        \includegraphics[width=0.15\textwidth]{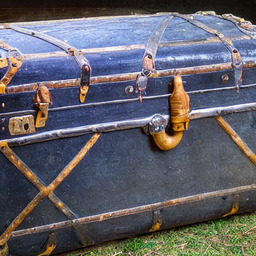}
        \includegraphics[width=0.15\textwidth]{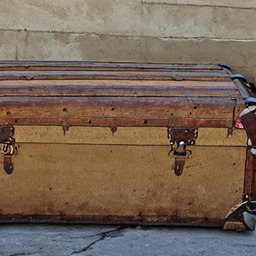}\\
        \includegraphics[width=0.15\textwidth]{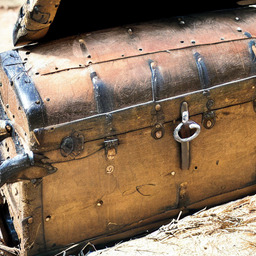}
        \includegraphics[width=0.15\textwidth]{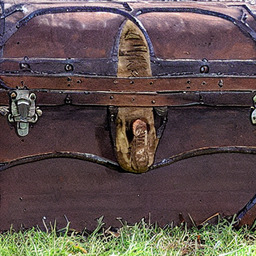}
        \includegraphics[width=0.15\textwidth]{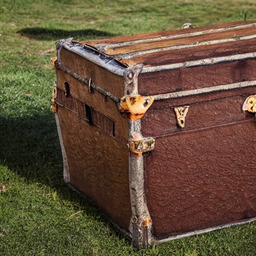}
        \includegraphics[width=0.15\textwidth]{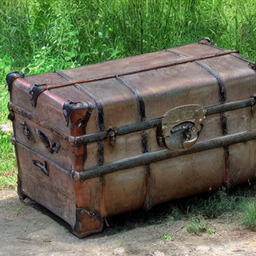}
        \includegraphics[width=0.15\textwidth]{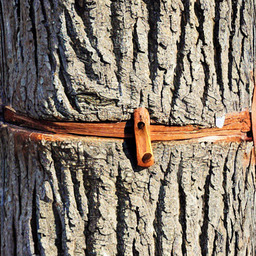}\\        \includegraphics[width=0.15\textwidth]{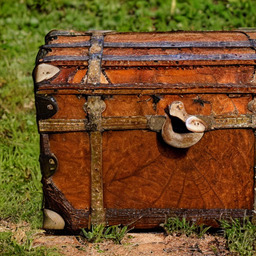}
        \includegraphics[width=0.15\textwidth]{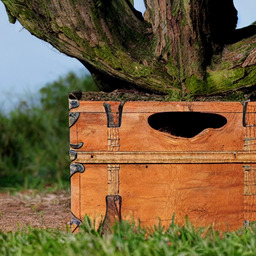}
        \includegraphics[width=0.15\textwidth]{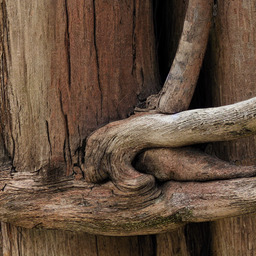}
        \includegraphics[width=0.15\textwidth]{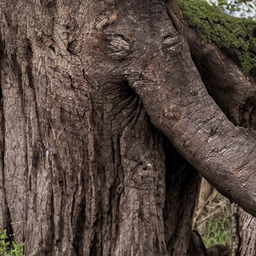}
        \includegraphics[width=0.15\textwidth]{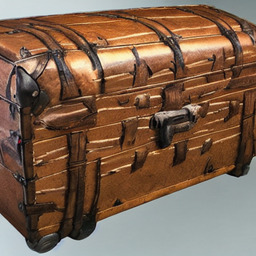}
        \caption{Unedited prompt encoding}
        \label{fig:trunk_amb}
    \end{subfigure}
    \begin{subfigure}{0.45\textwidth}
        \centering
        \includegraphics[width=0.3\textwidth]{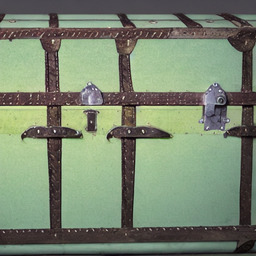}
        \includegraphics[width=0.3\textwidth]{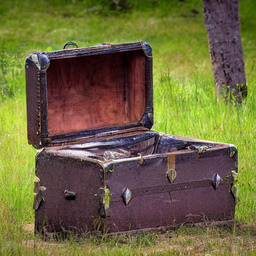}
        \includegraphics[width=0.3\textwidth]{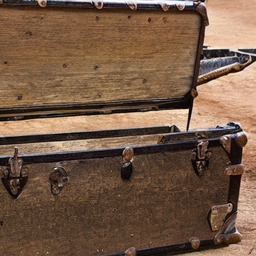}\\
        \includegraphics[width=0.3\textwidth]{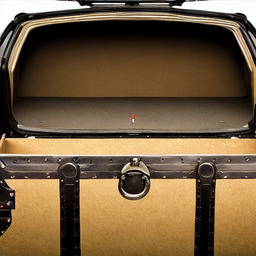}
        \includegraphics[width=0.3\textwidth]{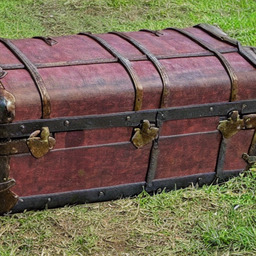}
        \includegraphics[width=0.3\textwidth]{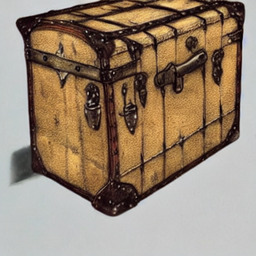}\\
        \includegraphics[width=0.3\textwidth]{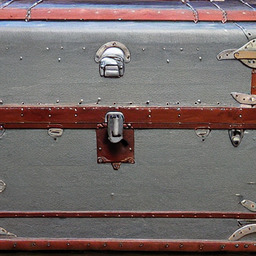}
        \includegraphics[width=0.3\textwidth]{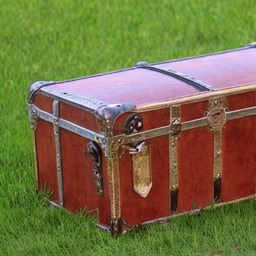}
        \includegraphics[width=0.3\textwidth]{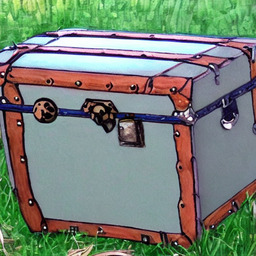}\\
        \includegraphics[width=0.3\textwidth]{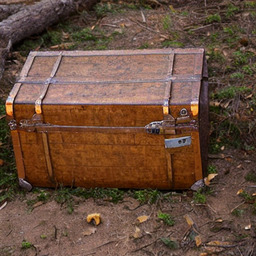}    \includegraphics[width=0.3\textwidth]{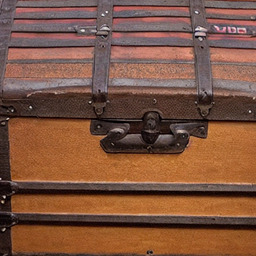}
        \includegraphics[width=0.3\textwidth]{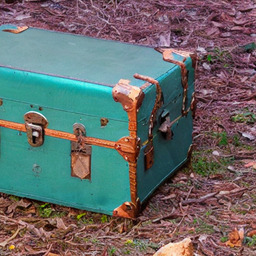}\\
        \includegraphics[width=0.3\textwidth]{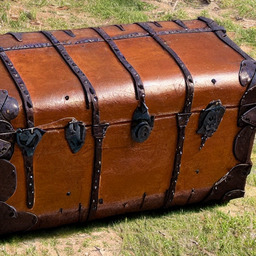}
        \includegraphics[width=0.3\textwidth]{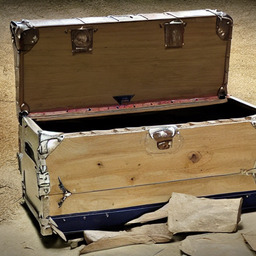}
        \includegraphics[width=0.3\textwidth]{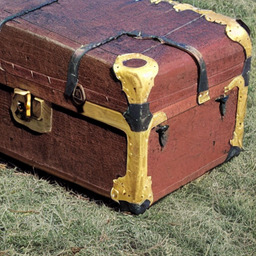}
        \caption{Encoding edited to favour luggage-related sense}
        \label{fig:trunk_1}
    \end{subfigure}
    \begin{subfigure}{0.45\textwidth}
        \centering
        \includegraphics[width=0.3\textwidth]{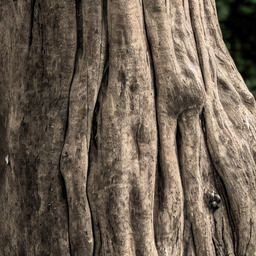}
        \includegraphics[width=0.3\textwidth]{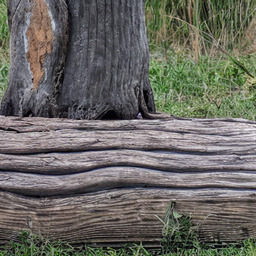}
        \includegraphics[width=0.3\textwidth]{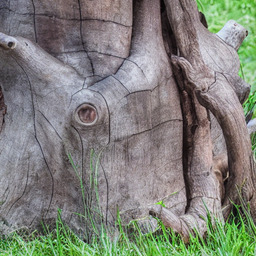}\\
        \includegraphics[width=0.3\textwidth]{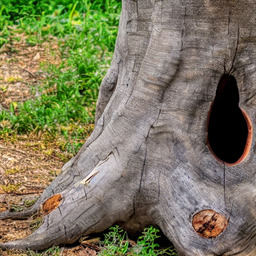}
        \includegraphics[width=0.3\textwidth]{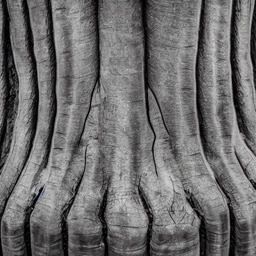}
        \includegraphics[width=0.3\textwidth]{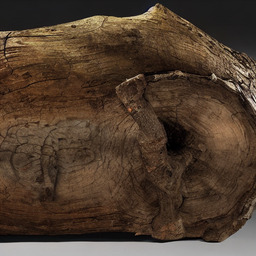}\\
        \includegraphics[width=0.3\textwidth]{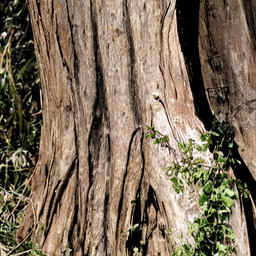}
        \includegraphics[width=0.3\textwidth]{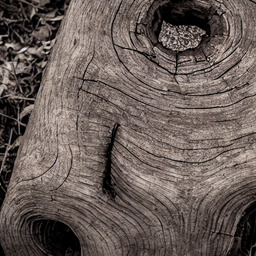}
        \includegraphics[width=0.3\textwidth]{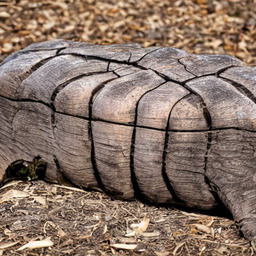}\\
        \includegraphics[width=0.3\textwidth]{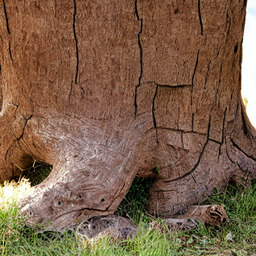}    \includegraphics[width=0.3\textwidth]{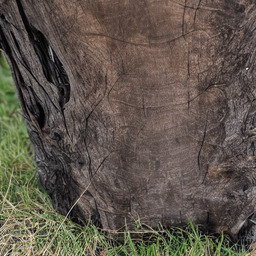}
        \includegraphics[width=0.3\textwidth]{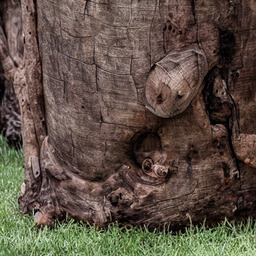}\\
        \includegraphics[width=0.3\textwidth]{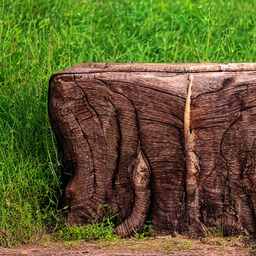}
        \includegraphics[width=0.3\textwidth]{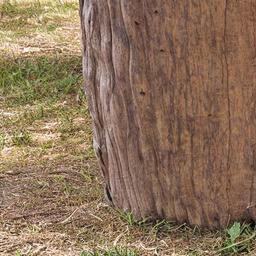}
        \includegraphics[width=0.3\textwidth]{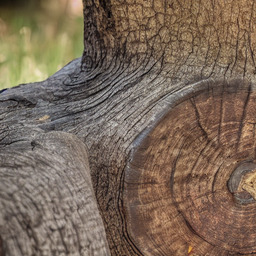}
        \caption{Encoding edited to favour tree-related sense}
        \label{fig:trunk_2}
    \end{subfigure}
    \caption{Prompt: \prompt{a trunk}}
    \label{fig:atrunk}
\end{figure}

\newpage
\section{Statistical Comparisons}

In this section we provide detailed information about the content of the images generated in our experiments, and provide details on statistical significance testing.

\subsection{Summing Encodings}\label{subsec:sum_stats}

The information in this section concerns the experiments described in \cref{sec:sumenc}.

For each pair of prompts, $\vb{s}_1$ and $\vb{s}_2$, 30 images were generated from each prompt, and 30 were generated from their weighted sum.

We were interested in the frequency with which both $\vb{s}_1$ and $\vb{s}_2$ were realised in an image generated by the weighted sum of their encodings.
To ensure that this was not a coincidence, statistical significance was established using an unpaired permutation test where the statistic concerned was the difference between the number of images in which both prompts were realised in the images generated from the weighted sum and the images generated from each prompt alone.
Some prompt pairs that we initially tried were discarded because they co-occurred too frequently in images generated by one of the prompts (for example, $\vb{s}_1=$\prompt{a dog in a public square} and $\vb{s}_2=$\prompt{a tree in a public square} were discarded because trees appear frequently in images generated by $\vb{s}_1$ alone, and thus a large number of samples would have been required to establish a statistically significant difference).

All weighted sums in this section were found to realise both $\vb{s}_1$ and $\vb{s}_2$ more frequently than either prompt alone to a statistically significant extent ($p<0.05$).

\begin{table}[H]
    \centering
    \begin{tabular}{ccccc}\toprule
    & \multicolumn{4}{c}{Number containing:}\\
        Input & Dog & Lake & Both & Neither \\\midrule
        $\clip(\text{\prompt{dog}})$ & 30 & 0 & 0 & 0\\
        $\clip(\text{\prompt{lake}})$  & 0 & 30 & 0 & 0\\
        $0.5\cdot(\clip(\text{\prompt{dog}})) + 0.5\cdot(\clip(\text{\prompt{lake}}))$& 3 & 11 & \textbf{16} & 0\\\bottomrule
    \end{tabular}
    \caption{Frequency of realisation of each concept for $\vb{s}_1=$\prompt{dog}, $\vb{s}_2=$\prompt{lake}}
    \label{tab:doglakesum}
\end{table}

\begin{table}[H]
    \centering
    \begin{tabular}{ccccc}\toprule
    & \multicolumn{4}{c}{Number containing:}\\
        Input & Cat & Tree & Both & Neither \\\midrule
        $\clip(\text{\prompt{cat}})$ & 30 & 0 & 0 & 0\\
        $\clip(\text{\prompt{tree}})$  & 0 & 30 & 0 & 0\\
        $0.5\cdot(\clip(\text{\prompt{cat}})) + 0.5\cdot(\clip(\text{\prompt{tree}}))$& 6 & 2 & \textbf{22} & 0\\\bottomrule
    \end{tabular}
    \caption{Frequency of realisation of each concept for $\vb{s}_1=$\prompt{cat}, $\vb{s}_2=$\prompt{tree}}
    \label{tab:cattreesum}
\end{table}

\begin{table}[H]
    \centering
    \begin{tabular}{ccccc}\toprule
    & \multicolumn{4}{c}{Number containing:}\\
        Input & Bear & Hat & Both & Neither \\\midrule
        $\clip(\text{\prompt{bear}})$ & 30 & 0 & 0 & 0\\
        $\clip(\text{\prompt{hat}})$  & 0 & 30 & 0 & 0\\
        $0.5\cdot(\clip(\text{\prompt{bear}})) + 0.5\cdot(\clip(\text{\prompt{hat}}))$& 20 & 3 & \textbf{5} & 2\\\bottomrule
    \end{tabular}
    \caption{Frequency of realisation of each concept for $\vb{s}_1=$\prompt{bear}, $\vb{s}_2=$\prompt{hat}}
    \label{tab:bearhatsum}
\end{table}

\begin{table}[H]
    \centering
    \begin{tabular}{ccccc}\toprule
    & \multicolumn{4}{c}{Number containing:}\\
        Input & Bear & Waterfall & Both & Neither \\\midrule
        $\clip(\text{\prompt{bear}})$ & 30 & 0 & 0 & 0\\
        $\clip(\text{\prompt{waterfall}})$  & 0 & 30 & 0 & 0\\
        $0.5\cdot(\clip(\text{\prompt{bear}})) + 0.5\cdot(\clip(\text{\prompt{waterfall}}))$& 10 & 12 & \textbf{6} & 2 \\\bottomrule
    \end{tabular}
    \caption{Frequency of realisation of each concept for $\vb{s}_1=$\prompt{bear}, $\vb{s}_2=$\prompt{waterfall}}
    \label{tab:bearwaterfallsum}
\end{table}

\begin{table}[H]
    \centering
    \begin{tabular}{ccccc}\toprule
    & \multicolumn{4}{c}{Number containing:}\\
        Input & Elephant & Snow & Both & Neither \\\midrule
        $\clip(\text{\prompt{elephant}})$ & 30 & 0 & 0 & 0\\
        $\clip(\text{\prompt{snow}})$  & 0 & 30 & 0 & 0\\
        $0.5\cdot(\clip(\text{\prompt{elephant}})) + 0.5\cdot(\clip(\text{\prompt{snow}}))$& 7 & 3 & \textbf{20} & 0 \\\bottomrule
    \end{tabular}
    \caption{Frequency of realisation of each concept for $\vb{s}_1=$\prompt{elephant}, $\vb{s}_2=$\prompt{snow}}
    \label{tab:elephantsnowsum}
\end{table}

\begin{table}[H]
    \centering
    \begin{tabular}{ccccc}\toprule
    & \multicolumn{4}{c}{Number containing:}\\
        Input & Giraffe & Beach & Both & Neither \\\midrule
        $\clip(\text{\prompt{giraffe}})$ & 30 & 0 & 0 & 0\\
        $\clip(\text{\prompt{beach}})$  & 0 & 30 & 0 & 0\\
        $0.5\cdot(\clip(\text{\prompt{giraffe}})) + 0.5\cdot(\clip(\text{\prompt{beach}}))$& 3 & 17 & \textbf{9} & 1 \\\bottomrule
    \end{tabular}
    \caption{Frequency of realisation of each concept for $\vb{s}_1=$\prompt{giraffe}, $\vb{s}_2=$\prompt{beach}}
    \label{tab:giraffebeach}
\end{table}

\begin{table}[H]
    \centering
    \begin{tabular}{ccccc}\toprule
    & \multicolumn{4}{c}{Number containing:}\\
        Input & Goat & Crown & Both & Neither \\\midrule
        $\clip(\text{\prompt{goat}})$ & 30 & 0 & 0 & 0\\
        $\clip(\text{\prompt{crown}})$  & 0 & 30 & 0 & 0\\
        $0.5\cdot(\clip(\text{\prompt{goat}})) + 0.5\cdot(\clip(\text{\prompt{crown}}))$& 11 & 4 & \textbf{11} & 4 \\\bottomrule
    \end{tabular}
    \caption{Frequency of realisation of each concept for $\vb{s}_1=$\prompt{goat}, $\vb{s}_2=$\prompt{crown}}
    \label{tab:goatcrown}
\end{table}

\begin{table}[H]
    \centering
    \begin{tabular}{ccccc}\toprule
    & \multicolumn{4}{c}{Number containing:}\\
        Input & Snow & Church & Both & Neither \\\midrule
        $\clip(\text{\prompt{snow}})$ & 30 & 0 & 0 & 0\\
        $\clip(\text{\prompt{church}})$  & 0 & 30 & 0 & 0\\
        $0.5\cdot(\clip(\text{\prompt{snow}})) + 0.5\cdot(\clip(\text{\prompt{church}}))$& 14 & 1 & \textbf{15} & 0 \\\bottomrule
    \end{tabular}
    \caption{Frequency of realisation of each concept for $\vb{s}_1=$\prompt{snow}, $\vb{s}_2=$\prompt{church}}
    \label{tab:snowchurch}
\end{table}

\begin{table}[H]
    \centering
    \begin{tabular}{ccccc}\toprule
    & \multicolumn{4}{c}{Number containing:}\\
        Input & Tiger & Glasses & Both & Neither \\\midrule
        $\clip(\text{\prompt{tiger}})$ & 30 & 0 & 0 & 0\\
        $\clip(\text{\prompt{glasses}})$  & 0 & 30 & 0 & 0\\
        $0.5\cdot(\clip(\text{\prompt{tiger}})) + 0.5\cdot(\clip(\text{\prompt{glasses}}))$& 21 & 1 & \textbf{8} & 0 \\\bottomrule
    \end{tabular}
    \caption{Frequency of realisation of each concept for $\vb{s}_1=$\prompt{tiger}, $\vb{s}_2=$\prompt{glasses}}
    \label{tab:tigerglasses}
\end{table}

\subsection{Sense Editing}\label{subsec:meaning_stats}

The information in this section concerns the experiments described in \cref{sec:superpos}.

For each prompt, $\vb{s}$, 30 images were generated from the unedited prompt encoding and 30 each were generated from the version edited to favour each sense.

We were interested in the frequency with which the intended sense was the only sense realised in an image generated by the one of the edited encodings.
The value for the targeted sense of an edited encoding is bolded if it was found to be a statistically significant increase of realisation of that sense when compared to the unedited encoding.
Statistical significance was established using an unpaired permutation test where the statistic concerned was the difference between the number of images in which the intended sense was realised, between images generated by the edited encoding and the unedited encoding.
The significance level used in these experiments was $p<0.05$.

\Cref{tab:bass_edit}, \cref{tab:bass_man_edit} and \cref{tab:bass_wall_edit} concern prompts including the word \prompt{bass}, where \colone{Sense 1} is a type of musical instrument (for counting purposes we included all kinds of instruments that can be described as a \prompt{bass}) and \coltwo{Sense 2} is a type of fish.

\begin{table}[H]
    \centering
    \begin{tabular}{ccccc}\toprule
        & \colone{Sense 1}& \coltwo{Sense 2} & Both & Neither \\\midrule
        \colamb{Unedited} & 16 & 14 & 0 & 0\\
        Edited (\colone{$\rightarrow$ Sense 1}) & \textbf{29} & 1 & 0 & 0\\
        Edited (\coltwo{$\rightarrow$ Sense 2}) & 3 & \textbf{27} & 0 & 0\\\bottomrule
    \end{tabular}
    \caption{Frequency of realisation of each concept for $\vb{s}=$\prompt{a bass} from the unedited prompt encoding, and encodings edited to favour each sense.}
    \label{tab:bass_edit}
\end{table}

\begin{table}[H]
    \centering
    \begin{tabular}{ccccc}\toprule
        & \colone{Sense 1}& \coltwo{Sense 2} & Both & Neither \\\midrule
        \colamb{Unedited} & 10 & 19 & 1 & 0 \\
        Edited (\colone{$\rightarrow$ Sense 1}) & \textbf{24} & 2 & 0 & 4\\
        Edited (\coltwo{$\rightarrow$ Sense 2}) & 0 & \textbf{30} & 0 & 0 \\\bottomrule
    \end{tabular}
    \caption{Frequency of realisation of each concept for $\vb{s}=$\prompt{a man holds a bass} from the unedited prompt encoding, and encodings edited to favour each sense.}
    \label{tab:bass_man_edit}
\end{table}

\begin{table}[H]
    \centering
    \begin{tabular}{ccccc}\toprule
        & \colone{Sense 1}& \coltwo{Sense 2} & Both & Neither \\\midrule
        \colamb{Unedited} & 23 & 5 & 0 & 2 \\
        Edited (\colone{$\rightarrow$ Sense 1}) & \textbf{30} &0 & 0 & 0\\
        Edited (\coltwo{$\rightarrow$ Sense 2}) & 4 & \textbf{25} & 0 & 1 \\\bottomrule
    \end{tabular}
    \caption{Frequency of realisation of each concept for $\vb{s}=$\prompt{a bass is displayed on a wall} from the unedited prompt encoding, and encodings edited to favour each sense.}
    \label{tab:bass_wall_edit}
\end{table}

\Cref{tab:bat_edit}, \cref{tab:bat_fly_edit}, \cref{tab:bat_boy_edit} and \cref{tab:bat_grass_edit} concern prompts that include the ambiguous word \prompt{bat}, 
for which \colone{Sense 1} is a winged nocturnal mammal, and \coltwo{Sense 2} is a piece of equipment used to play sports such as baseball.

\begin{table}[H]
    \centering
    \begin{tabular}{ccccc}\toprule
        & \colone{Sense 1}& \coltwo{Sense 2} & Both & Neither \\\midrule
        \colamb{Unedited} & 30 & 0 & 0 & 0\\
        Edited (\colone{$\rightarrow$ Sense 1}) & 30 & 0 & 0 & 0\\
        Edited (\coltwo{$\rightarrow$ Sense 2}) & 2 & \textbf{16} & 8 & 4\\\bottomrule
    \end{tabular}
    \caption{Frequency of realisation of each concept for $\vb{s}=$\prompt{a bat} from the unedited prompt encoding, and encodings edited to favour each sense.}
    \label{tab:bat_edit}
\end{table}

\begin{table}[H]
    \centering
    \begin{tabular}{ccccc}\toprule
        & \colone{Sense 1}& \coltwo{Sense 2} & Both & Neither \\\midrule
        \colamb{Unedited} & 22 & 4 & 2 & 2\\
        Edited (\colone{$\rightarrow$ Sense 1}) & 27 & 0 & 1 & 2\\
        Edited (\coltwo{$\rightarrow$ Sense 2}) & 1 & \textbf{24} & 3 & 2\\\bottomrule
    \end{tabular}
    \caption{Frequency of realisation of each concept for $\vb{s}=$\prompt{a bat and a baseball fly through the air} from the unedited prompt encoding, and encodings edited to favour each sense.}
    \label{tab:bat_fly_edit}
\end{table}

\begin{table}[H]
    \centering
    \begin{tabular}{ccccc}\toprule
        & \colone{Sense 1}& \coltwo{Sense 2} & Both & Neither \\\midrule
        \colamb{Unedited} & 4 & 17 & 4 & 5\\
        Edited (\colone{$\rightarrow$ Sense 1}) & \textbf{18} & 3 & 5 & 4\\
        Edited (\coltwo{$\rightarrow$ Sense 2}) & 0 & \textbf{27} & 0 & 3\\\bottomrule
    \end{tabular}
    \caption{Frequency of realisation of each concept for $\vb{s}=$\prompt{a boy holds a black bat} from the unedited prompt encoding, and encodings edited to favour each sense.}
    \label{tab:bat_boy_edit}
\end{table}

\begin{table}[H]
    \centering
    \begin{tabular}{ccccc}\toprule
        & \colone{Sense 1}& \coltwo{Sense 2} & Both & Neither \\\midrule
        \colamb{Unedited} & 30 & 0 & 0 & 0\\
        Edited (\colone{$\rightarrow$ Sense 1}) & 30 & 0 & 0 & 0\\
        Edited (\coltwo{$\rightarrow$ Sense 2}) & 2 & \textbf{18} & 3 & 7\\\bottomrule
    \end{tabular}
    \caption{Frequency of realisation of each concept for $\vb{s}=$\prompt{a bat laying on the grass} from the unedited prompt encoding, and encodings edited to favour each sense.}
    \label{tab:bat_grass_edit}
\end{table}

\Cref{tab:crane_edit}, \cref{tab:crane_ocean_edit} and \cref{tab:crane_nature_edit} all include the ambiguous word \prompt{crane}, for which \colone{Sense 1} is a long-legged bird, and \coltwo{Sense 2} is a piece of machinery often used in construction.

\begin{table}[H]
    \centering
    \begin{tabular}{ccccc}\toprule
        & \colone{Sense 1}& \coltwo{Sense 2} & Both & Neither \\\midrule
        \colamb{Unedited} & 24 & 4 & 2 & 0\\
        Edited (\colone{$\rightarrow$ Sense 1}) & 29 & 0 & 0 & 1\\
        Edited (\coltwo{$\rightarrow$ Sense 2}) & 10 & \textbf{18} & 2 & 0\\\bottomrule
    \end{tabular}
    \caption{Frequency of realisation of each concept for $\vb{s}=$\prompt{a crane} from the unedited prompt encoding, and encodings edited to favour each sense.}
    \label{tab:crane_edit}
\end{table}

\begin{table}[H]
    \centering
    \begin{tabular}{ccccc}\toprule
        & \colone{Sense 1}& \coltwo{Sense 2} & Both & Neither \\\midrule
        \colamb{Unedited} & 19 & 9 & 1 & 1\\
        Edited (\colone{$\rightarrow$ Sense 1}) & \textbf{27} & 1 & 0 & 2\\
        Edited (\coltwo{$\rightarrow$ Sense 2}) & 1 & \textbf{28} & 1 & 0\\\bottomrule
    \end{tabular}
    \caption{Frequency of realisation of each concept for $\vb{s}=$\prompt{a crane by the ocean} from the unedited prompt encoding, and encodings edited to favour each sense.}
    \label{tab:crane_ocean_edit}
\end{table}

\begin{table}[H]
    \centering
    \begin{tabular}{ccccc}\toprule
        & \colone{Sense 1}& \coltwo{Sense 2} & Both & Neither \\\midrule
        \colamb{Unedited} & 9 & 16 & 5 & 0\\
        Edited (\colone{$\rightarrow$ Sense 1}) & \textbf{24} & 4 & 1 & 1\\
        Edited (\coltwo{$\rightarrow$ Sense 2}) & 1 & \textbf{29} & 0 & 0\\\bottomrule
    \end{tabular}
    \caption{Frequency of realisation of each concept for $\vb{s}=$\prompt{a crane surrounded by nature} from the unedited prompt encoding, and encodings edited to favour each sense.}
    \label{tab:crane_nature_edit}
\end{table}

\Cref{tab:glasses_edit} gives results for the prompt \prompt{glasses on a table}, where \colone{Sense 1} of \prompt{glasses} refers to glasses used for drinking, and \coltwo{Sense 2} refers to eyeglasses.

\begin{table}[H]
    \centering
    \begin{tabular}{ccccc}\toprule
        & \colone{Sense 1}& \coltwo{Sense 2} & Both & Neither \\\midrule
        \colamb{Unedited} & 8 & 21 & 1 & 0\\
        Edited (\colone{$\rightarrow$ Sense 1}) & \textbf{24} & 3 & 0 & 3\\
        Edited (\coltwo{$\rightarrow$ Sense 2}) & 0 & \textbf{29} & 0 & 1\\\bottomrule
    \end{tabular}
    \caption{Frequency of realisation of each concept for $\vb{s}=$\prompt{glasses on a table} from the unedited prompt encoding, and encodings edited to favour each sense.}
    \label{tab:glasses_edit}
\end{table}

\Cref{tab:seal_edit} and \cref{tab:seal_envelope_edit} give results for prompts containing the word \prompt{seal}, where \colone{Sense 1} refers to a wax seal or an official seal on a letter or document, and \coltwo{Sense 2} refers to a marine mammal.

\begin{table}[H]
    \centering
    \begin{tabular}{ccccc}\toprule
        & \colone{Sense 1}& \coltwo{Sense 2} & Both & Neither \\\midrule
        \colamb{Unedited} & 2 & 22 & 6 & 0\\
        Edited (\colone{$\rightarrow$ Sense 1}) & \textbf{26} & 0 & 4 & 0\\
        Edited (\coltwo{$\rightarrow$ Sense 2}) & 0 & \textbf{30} & 0 & 0\\\bottomrule
    \end{tabular}
    \caption{Frequency of realisation of each concept for $\vb{s}=$\prompt{a seal} from the unedited prompt encoding, and encodings edited to favour each sense.}
    \label{tab:seal_edit}
\end{table}

\begin{table}[H]
    \centering
    \begin{tabular}{ccccc}\toprule
        & \colone{Sense 1}& \coltwo{Sense 2} & Both & Neither \\\midrule
        \colamb{Unedited} & 0 & 25 & 2 & 3\\
        Edited (\colone{$\rightarrow$ Sense 1}) & \textbf{21} & 1 & 7 & 1\\
        Edited (\coltwo{$\rightarrow$ Sense 2}) & 0 & \textbf{30} & 0 & 0\\\bottomrule
    \end{tabular}
    \caption{Frequency of realisation of each concept for $\vb{s}=$\prompt{a seal on an envelope} from the unedited prompt encoding, and encodings edited to favour each sense.}
    \label{tab:seal_envelope_edit}
\end{table}

\Cref{tab:trunk_edit} provides results for the prompt \prompt{a trunk} where \colone{Sense 1} of \prompt{trunk} refers to a box used as luggage (for counting purposes, the trunk of a car was also included in this sense) and \coltwo{Sense 2} refers to the trunk of a tree.

\begin{table}[H]
    \centering
    \begin{tabular}{ccccc}\toprule
        & \colone{Sense 1}& \coltwo{Sense 2} & Both & Neither \\\midrule
        \colamb{Unedited} & 15 & 13 & 2 & 0\\
        Edited (\colone{$\rightarrow$ Sense 1}) & \textbf{30} & 0 & 0 & 0\\
        Edited (\coltwo{$\rightarrow$ Sense 2}) & 0 & \textbf{28} & 2 & 0\\\bottomrule
    \end{tabular}
    \caption{Frequency of realisation of each concept for $\vb{s}=$\prompt{a trunk} from the unedited prompt encoding, and encodings edited to favour each sense.}
    \label{tab:trunk_edit}
\end{table}

\section{Sentences Used to Find Meaning Directions}

Here we provide the sentences $\samb{}$, $\sone{}$ and $\stwo{}$ that were used to calculate difference subspaces and meaning directions for each of the words we used.

\begin{table}[H]
    \setlength{\extrarowheight}{2.5pt} 
    \centering
    \begin{tabularx}{\textwidth}{LLL}\toprule
    $\samb{}$  & $\sone{}$ &$\stwo{}$\\\midrule
a bass & a double bass & a sea bass\\ 
there is a bass & there is a double bass & there is a sea bass\\ 
the person saw a bass & the musician played a double bass & the fisherman caught a sea bass\\ 
a person mentions a bass & a musician plays a double bass & an angler holds a sea bass\\ 
a nearby location has a bass & a jazz band has a double bass & a local aquarium has a sea bass\\ 
\bottomrule
    \end{tabularx}
    \caption{Sentences used to find meaning directions for \prompt{bass}}
    \label{tab:bass_sents}
\end{table}

\begin{table}[H]
    \setlength{\extrarowheight}{2.5pt} 
    \centering
    \begin{tabularx}{\textwidth}{LLL}\toprule
    $\samb{}$  & $\sone{}$ &$\stwo{}$\\\midrule
a bat & a fruit bat & a baseball bat\\ 
there is a bat & there is a fruit bat & there is a baseball bat\\ 
i do things with the bat & i feed insects to the fruit bat & i play baseball with the baseball bat\\ 
the person saw a bat & the boy saw a fruit bat & the boy bought a baseball bat\\ 
a person mentions a bat & a wildlife expert feeds a fruit bat & a baseball player swings a baseball bat\\ 
a bat is laying on the floor & a fruit bat is hanging from the tree & a baseball bat is laying on the base\\ 
a bat in a box & a fruit bat in a cave & a baseball bat in a store\\ 
a nearby location has a bat & a local zoo keeps an fruit bat & a sports store sells a baseball bat\\ 
\bottomrule
    \end{tabularx}
    \caption{Sentences used to find meaning directions for \prompt{bat}}
    \label{tab:bat_sents}
\end{table}

\begin{table}[H]
    \setlength{\extrarowheight}{2.5pt} 
    \centering
    \begin{tabularx}{\textwidth}{LLL}\toprule
    $\samb{}$  & $\sone{}$ &$\stwo{}$\\\midrule
a crane & a sandhill crane & a tower crane\\ 
there is a crane & there is a sandhill crane & there is a tower crane\\ 
there is a crane on the other side & there is a sandhill crane on the nature reserve & there is a tower crane on the building site\\ 
a crane is tall & a sandhill crane hunts fish & a tower crane lifts loads\\ 
a boy sees a crane & a boy feeds a sandhill crane & a man operates a tower crane\\ 
a crane beside a tree & a sandhill crane beside a nest & a tower crane beside a bulldozer\\ 
a crane is casting a shadow & a sandhill crane is eating some fish & a tower crane is lifting a container\\ 
a crane by the ocean & a sandhill crane in a nest & a tower crane in a quarry\\ 
\bottomrule
    \end{tabularx}
    \caption{Sentences used to find meaning directions for \prompt{crane}}
    \label{tab:crane_sents}
\end{table}

\begin{table}[H]
    \setlength{\extrarowheight}{2.5pt} 
    \centering
    \begin{tabularx}{\textwidth}{LLL}\toprule
    $\samb{}$  & $\sone{}$ &$\stwo{}$\\\midrule
glasses & wine glasses & reading glasses\\ 
there are glasses & there are wine glasses & there are reading glasses\\ 
the person saw glasses & the waiter filled wine glasses & the scientist wore reading glasses\\ 
a person holds glasses & a waiter fills wine glasses & a scientist wears reading glasses\\ 
glasses are being used & wine glasses are being cleaned & reading glasses are being cleaned\\ 
\bottomrule
    \end{tabularx}
    \caption{Sentences used to find meaning directions for \prompt{glasses}}
    \label{tab:glasses_sents}
\end{table}

\begin{table}[H]
    \setlength{\extrarowheight}{2.5pt} 
    \centering
    \begin{tabularx}{\textwidth}{LLL}\toprule
    $\samb{}$  & $\sone{}$ &$\stwo{}$\\\midrule
a seal & a wax seal & a harp seal\\ 
there is a seal & there is a wax seal & there is a harp seal\\ 
the person saw a seal & the postmaster stamped a wax seal & the zookeeper fed a harp seal\\ 
a person mentions a seal & a butler opens a wax seal & a boy pets a harp seal\\ 
a seal in a frame & a wax seal on an envelope & a harp seal in the ocean\\ 
a nearby location has a seal & a fancy letter has a wax seal & a large zoo has a harp seal\\ 
\bottomrule
    \end{tabularx}
    \caption{Sentences used to find meaning directions for \prompt{seal}}
    \label{tab:seal_sents}
\end{table}

\begin{table}[H]
    \setlength{\extrarowheight}{2.5pt} 
    \centering
    \begin{tabularx}{\textwidth}{LLL}\toprule
    $\samb{}$  & $\sone{}$ &$\stwo{}$\\\midrule
a trunk & a storage trunk & a tree trunk\\ 
there is a trunk & there is a storage trunk & there is a tree trunk\\ 
the person saw a trunk & the traveller carried a storage trunk & the lumberjack sawed a tree trunk\\ 
a person mentions a trunk & a passenger carries a storage trunk & a carpenter uses a tree trunk\\ 
a trunk is being used & a storage trunk is being packed & a tree trunk is being felled\\ 
\bottomrule
    \end{tabularx}
    \caption{Sentences used to find meaning directions for \prompt{trunk}}
    \label{tab:trunk_sents}
\end{table}
\end{document}